%% file: main.tex
\documentclass[10pt,twocolumn,letterpaper]{article}

\usepackage[pagenumbers]{wacv} %

\usepackage[accsupp]{axessibility}

\input{preamble}

\definecolor{wacvblue}{rgb}{0.21,0.49,0.74}
\usepackage[pagebackref,breaklinks,colorlinks,allcolors=wacvblue]{hyperref}

\hypersetup{linktoc=all} %

\def\confName{WACV}
\def\confYear{2027}

\title{Entropy-Regularized Integrated Gradients for Stable Feature Attribution}

\author{Rukmangadh Sai Myana\\
Florida International University\\
{\tt\small rmyan001@fiu.edu}
\and
Sumit Kumar Jha\\
University of Florida\\
{\tt\small sumit.jha@ufl.edu}
\and
Yanzhao Wu\\
Florida International University\\
{\tt\small yawu@fiu.edu}
}

\begin{document}
\maketitle

\addtocontents{toc}{\protect\setcounter{tocdepth}{0}}
\input{sec/0_abstract}    
\input{sec/1_intro}
\input{sec/2_related}
\input{sec/3_prelim}
\input{sec/4_method}
\input{sec/5_experiments}
\input{sec/6_conclusion}

{
    \small
    \bibliographystyle{ieeenat_fullname}
    \bibliography{main}
}

\allowdisplaybreaks
\onecolumn
\appendix
\input{sec/7_appendix}

\end{document}

%% file: preamble.tex
\usepackage{bm}
\usepackage{algorithm}
\usepackage{algpseudocode}
\usepackage{dsfont}
\usepackage{tabularx}
\usepackage{wrapfig}  %
\usepackage{bigints}  %
\usepackage{mathtools}  %
\usepackage[normalem]{ulem}  %
\usepackage{multirow}  %
\usepackage{makecell}  %
\usepackage{dashbox}  %
\usepackage{float}  %
\usepackage{amsthm}  %
\usepackage{needspace}
\usepackage{textgreek}  %
\usepackage{arydshln}  %

\newtheorem{theorem}{Theorem}

\newtheorem{proposition}[theorem]{Proposition}

\newtheorem{assumption}{Assumption}

\usepackage{tikz}
\usetikzlibrary{arrows,arrows.meta,decorations.markings, decorations.pathreplacing,calc,positioning,shapes.geometric}

\DeclareMathOperator*{\argmax}{arg\,max}

\DeclareMathOperator{\Softmax}{Softmax}

\def\approxprop{%
  \def\p{%
    \setbox0=\vbox{\hbox{$\propto$}}%
    \ht0=0.6ex \box0 }%
  \def\s{%
    \vbox{\hbox{$\sim$}}%
  }%
  \mathrel{\raisebox{0.7ex}{%
      \mbox{$\underset{\s}{\p}$}%
    }}%
}

%% file: sec/0_abstract.tex
\begin{abstract}
Path-based attribution methods like Integrated Gradients (IG) are widely used to interpret neural networks, yet imperceptible input perturbations can drastically alter their attribution maps while leaving model predictions unchanged. We introduce Entropy-Regularized Integrated Gradients (ERIG), a path-based attribution method that intrinsically attenuates gradient irregularities along the integration path using maximum-entropy softmax weighting with temperature parameter $\tau$. We quantify the effect of the entropy-regularized weighting on attribution stability by using the Lipschitz properties of the model and its gradients. We decompose the attribution change under an input perturbation into a model-response term, whose bound is $\tau$-independent, and a weight-variation term, whose bound decreases at rate $\mathcal{O}(1/(2\tau))$. Their ratio $R(\tau)$ characterizes the trade-off between gradient information and attribution stability. We further add a new dimension to attribution evaluation by adapting classical noise estimators into two metrics, LINV-C and LINV-P, that quantify attribution noise. Across eleven CNN and Transformer backbones on ImageNet and Oxford-IIIT Pet datasets, our experiments show that, relative to the next-best method, ERIG remains comparable on ranking-based fidelity metrics while improving localization by ${\sim}24\%$ and attribution stability by ${\sim}47\times$ on average, reaching up to four orders of magnitude improvement over IG --- all at negligible additional computational cost. \url{https://github.com/braceletboy/ERIG}.
\end{abstract}

%% file: sec/1_intro.tex
\section{Introduction}
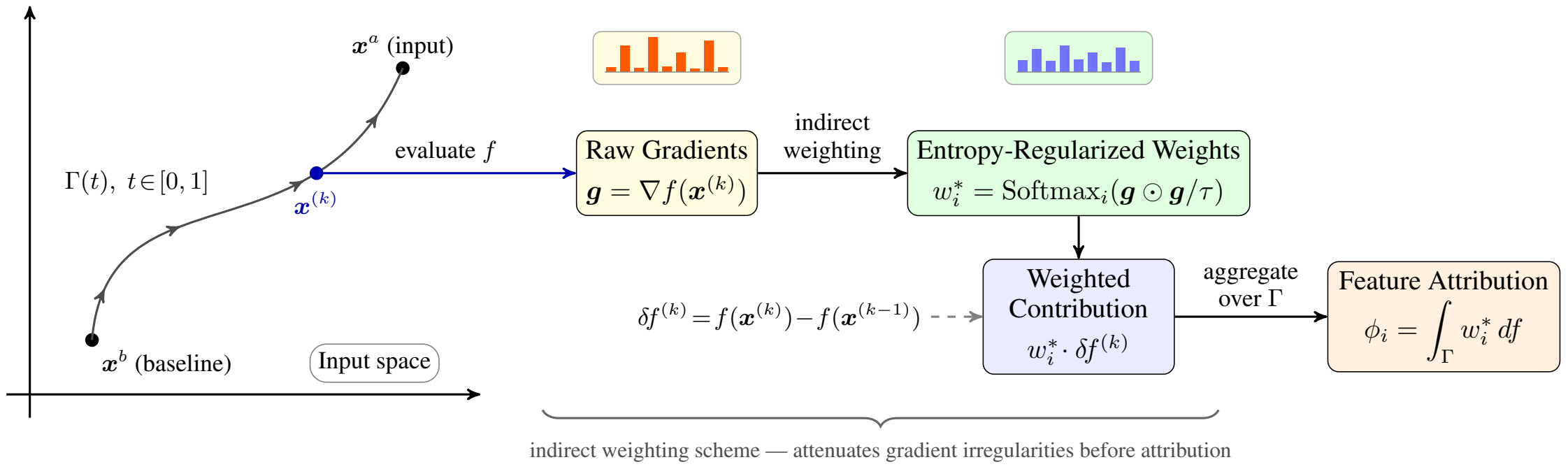
\begin{figure*}[t]
\centering
\resizebox{0.85\textwidth}{!}{
    \begin{tikzpicture}[
        >=stealth',
        every node/.style={font=\normalsize},
        block/.style={
            draw, rounded corners=4pt, fill=blue!8,
            minimum height=2.6em, minimum width=7em,
            align=center, font=\normalsize
        },
        pipeblock/.style={
            draw, rounded corners=4pt,
            minimum height=2.6em, minimum width=5.5em,
            align=center, font=\normalsize
        },
        annot/.style={font=\small, align=center},
    ]

    \draw[->, thick] (-0.3,0) -- (5.8,0);
    \draw[->, thick] (0,-0.3) -- (0,5.0);

    \node[draw=black!50, rounded corners=6pt,
          inner sep=3pt, font=\small, anchor=south west]
        at (3.6, 0.15) {Input space};

    \coordinate (xb) at (0.8, 0.7);
    \coordinate (xa) at (4.8, 4.2);

    \fill[black] (xb) circle (2.5pt) node[below right, font=\small]{$\bm{x}^b$ (baseline)};
    \fill[black] (xa) circle (2.5pt) node[above, font=\small]{$\bm{x}^a$ (input)};

    \draw[thick, color=black!70,
          postaction={decorate, decoration={
              markings,
              mark=at position 0.12 with {\arrow{>}},
              mark=at position 0.35 with {\arrow{>}},
              mark=at position 0.65 with {\arrow{>}},
              mark=at position 0.88 with {\arrow{>}},
          }}]
        (xb) .. controls (1.0,3.0) and (3.7,1.6) .. (xa);

    \path (xb) .. controls (1.0,3.0) and (3.7,1.6) .. (xa)
          coordinate[pos=0.74] (mid);
    \fill[blue!70!black] (mid) circle (2.5pt);
    \node[font=\small, blue!70!black, anchor=north, yshift=-3pt]
        at (mid) {$\bm{x}^{(k)}$};

    \node[annot, anchor=south, yshift=2pt] at ($(xb)!0.22!(xa)+(-.3,0.9)$)
        {$\Gamma(t),\; t\!\in\![0,1]$};

    \draw[rounded corners=3pt, draw=black!35, fill=yellow!15, line width=0.4pt] (7.25,3.99) rectangle (9.15,4.67);
    \fill[orange!70!red] (7.42,4.15) rectangle (7.54,4.21);
    \fill[orange!70!red] (7.60,4.15) rectangle (7.72,4.49);
    \fill[orange!70!red] (7.78,4.15) rectangle (7.90,4.20);
    \fill[orange!70!red] (7.96,4.15) rectangle (8.08,4.60);
    \fill[orange!70!red] (8.14,4.15) rectangle (8.26,4.22);
    \fill[orange!70!red] (8.32,4.15) rectangle (8.44,4.40);
    \fill[orange!70!red] (8.50,4.15) rectangle (8.62,4.19);
    \fill[orange!70!red] (8.68,4.15) rectangle (8.80,4.55);
    \fill[orange!70!red] (8.86,4.15) rectangle (8.98,4.21);
    \draw[black!45, line width=0.4pt] (7.40,4.15) -- (9.00,4.15);
    \draw[rounded corners=3pt, draw=black!35, fill=green!12, line width=0.4pt] (12.55,3.99) rectangle (14.45,4.67);
    \fill[blue!55] (12.72,4.15) rectangle (12.84,4.30);
    \fill[blue!55] (12.90,4.15) rectangle (13.02,4.44);
    \fill[blue!55] (13.08,4.15) rectangle (13.20,4.29);
    \fill[blue!55] (13.26,4.15) rectangle (13.38,4.49);
    \fill[blue!55] (13.44,4.15) rectangle (13.56,4.31);
    \fill[blue!55] (13.62,4.15) rectangle (13.74,4.40);
    \fill[blue!55] (13.80,4.15) rectangle (13.92,4.27);
    \fill[blue!55] (13.98,4.15) rectangle (14.10,4.46);
    \fill[blue!55] (14.16,4.15) rectangle (14.28,4.29);
    \draw[black!45, line width=0.4pt] (12.70,4.15) -- (14.30,4.15);

    \coordinate (x_82) at (8.2, 0);
    \coordinate (x_135) at (13.5, 0);

    \node[pipeblock, fill=yellow!15] (grad) at (mid -| x_82)
        {Raw Gradients\\[3pt] $\bm{g} = \nabla f(\bm{x}^{(k)})$};

    \draw[->, thick, blue!70!black]
        (mid) -- (grad.west)
        node[midway, above, annot, black]{evaluate $f$};

    \node[pipeblock, fill=green!12, minimum width=8em] (smax) at (mid -| x_135)
        {Entropy-Regularized Weights\\[3pt] $w_i^* = \Softmax_{i}(\bm{g} \odot \bm{g}/\tau)$};

    \draw[->, thick] (grad.east) -- (smax.west)
        node[midway, above, annot]{indirect\\[-1pt] weighting};

    \node[block, fill=blue!8] (agg) at (13.5, 1.0)
        {Weighted\\[-1pt] Contribution\\[3pt] $w_i^{*}\!\cdot\delta\!f^{(k)}$};

    \draw[->, thick] (smax.south) -- (agg.north);

    \node[annot, anchor=east] at (11.6, 1.0) {$\delta\!f^{(k)}\!=\!f(\bm{x}^{(k)})\!-\!f(\bm{x}^{(k-1)})$};
    \draw[->, dashed, thick, black!50] (11.6, 1.0) -- (agg.west);

    \node[block, fill=orange!12, minimum width=8.5em] (fa) at (18.2, 1.0)
        {Feature Attribution\\[3pt] $\phi_i = \displaystyle\int_\Gamma w_i^{*}\,df$};

    \draw[->, thick] (agg.east) -- (fa.west)
        node[midway, above, annot]{aggregate\\[-1pt] over $\Gamma$};

    \draw[decorate, decoration={brace, amplitude=6pt, mirror}, thick, black!60]
        (6.6, -0.2) -- (15.3, -0.2)
        node[midway, below=8pt, font=\footnotesize, black!70, align=center]
        {indirect weighting scheme --- attenuates gradient irregularities before attribution};

    \end{tikzpicture}
}
\caption{\textbf{Entropy-Regularized Integrated Gradients (ERIG).} An integration path $\Gamma(t)$ connects the baseline $\bm{x}^b$ to the input $\bm{x}^a$. At each path point $\bm{x}^{(k)}$, the indirect weighting scheme maps the raw gradient $\bm{g} = \nabla f$ to importance weights $w_i^{*}$ through a softmax over the squared gradients $\bm{g} \odot \bm{g}$ with temperature $\tau$. Intuitively, the weights stay close to uniform unless the local sensitivity warrants concentration, and $\tau$ controls how readily they concentrate. Summing the weighted contributions $w_i^{*}\cdot\delta\!f^{(k)}$ over the path yields the final attribution $\phi_i$.}
\label{fig: isg}
\end{figure*}
Deep learning has advanced a myriad of fields such as vision~\cite{wang2024yolov10, woo2023convnext}, language~\cite{he2021debertav3}, science~\cite{shi2025deep, krishna2024generalized, shi2026weather}, and archaeology~\cite{sakai2024ai}. Yet these models remain \emph{black boxes}, offering little insight into their predictions. High-stakes applications demand transparent explanations and raise concerns about model biases~\cite{chauhan2022role, park2022fairness} and fairness~\cite{wang2024individual}.

Explainable Artificial Intelligence (XAI) has emerged to address this lack of transparency~\cite{adadi2018peeking, gunning2019darpa}, yet existing methods carry critical weaknesses~\cite{10.1007/978-3-032-02489-3_15}; Ghorbani et al.~\cite{ghorbani2019interpretation} demonstrate that imperceptible input perturbations can drastically alter gradient-based explanations while leaving predictions unchanged. The root cause is the irregular input-space gradient field $\nabla f$ of neural networks. Standard deep networks exhibit excess input-space curvature~\cite{srinivas2022efficiently, dombrowski2019explanations, moosavi2019robustness}, so minuscule perturbations produce large gradient shifts~\cite{szegedy2014intriguing, goodfellow2015explaining}, and the gradient field carries high-frequency components misaligned with semantic structure~\cite{tsipras2019robustness, yin2019fourier, etmann2019connection}. This irregularity is baked in by training. Standard objectives do not regularize input gradients; gradients \emph{shatter} toward white noise as depth grows~\cite{balduzzi2017shattered}, sparse activation patterns concentrate gradient information in a few noise-sensitive directions~\cite{hanin2019deep}, and while batch normalization smooths the optimization landscape~\cite{santurkar2018does}, it does not constrain input gradients explicitly. In this context, path integration averages out some of these fluctuations, but the effect is incomplete, and attribution maps still inherit substantial noise~\cite{ghorbani2019interpretation, kindermans2019unreliability, adebayo2018sanity, alvarez2018robustness, yeh2019fidelity, bansal2020sam}.

For model predictions, diverse ensembles can improve accuracy~\cite{wu2021boosting} and robustness~\cite{wu2023exploring, wei2020crosslayer}. At the explanation level, SmoothGrad~\cite{smilkov2017smoothgrad} addresses these issues post hoc by averaging gradients over local input perturbations to improve stability and reduce noise. This motivates explicitly regularizing gradient noise for reliable explanations.

We design \textbf{Entropy-Regularized Integrated Gradients (ERIG)}, a path-based feature attribution method that addresses the question: \emph{How can we weight gradient contributions along the integration path so that attribution noise is suppressed while genuine model sensitivity is preserved?} Intuitively, ERIG answers this by not passing raw gradients directly into the attribution. Instead, it converts them into weights that stay close to uniform unless the local sensitivity warrants concentration. Here, a single temperature $\tau$ controls how readily the weights concentrate. We provide a summary of ERIG in~\Cref{fig: isg}. Unlike IG~\cite{sundararajan2017axiomatic} and IDGI~\cite{yang2023idgi}, which couple weights directly to raw gradients, ERIG uses a maximum-entropy softmax over squared gradients as an indirect sensitivity weighting. A formal Lipschitz analysis then reveals that $\tau$ selectively controls how gradient perturbations propagate through to the weights. The associated ratio $R(\tau)$, defined in \Cref{sec: lipschitz-decomp}, quantifies this selectiveness. Our contributions are threefold:
\begin{itemize}
    \item We define an indirect weighting scheme and leverage a maximum-entropy principle to derive a closed-form softmax expression with a temperature parameter $\tau$ that controls the method's sensitivity to gradient irregularities.
    \item We prove a decomposition theorem establishing that $\tau$ attenuates the weight-variation component of attribution perturbation at rate $\mathcal{O}(1/(2\tau))$, while the model-response component admits a $\tau$-independent upper bound.
    \item We adapt established noise estimators~\cite{immerkaer1996fast, chen2015efficient} to the attribution setting, yielding two attribution noise metrics, LINV-C and LINV-P. This adds a new noise dimension to traditional quantitative attribution evaluations.
\end{itemize}

%% file: sec/2_related.tex
\section{\texorpdfstring{Related Work}{Related Work}}

Since its introduction, Integrated Gradients (IG)~\cite{sundararajan2017axiomatic} has inspired a broad family of extensions. Some improve its numerical approximation~\cite{swain2024riemann}, while others adapt it to temporal data~\cite{jang2025timing}, NLP~\cite{enguehard2023sequential}, and learned manifolds~\cite{zaher2024manifold}. Further works decompose it at the neuron or layer level~\cite{dhamdhere2018important}, combine it with segmentation~\cite{kapishnikov2019xrai}, or connect it to game theory~\cite{walker2024integrated, chen2022explaining}. Another line of work modifies the baseline or the path. For example, Blur-IG~\cite{xu2020attribution} blurs the input progressively, Guided IG~\cite{kapishnikov2021guided} selects low-gradient path points, and Expected Gradients~\cite{erion2021improving} averages over data-sampled baselines. Iterative counterfactual updates~\cite{zhuo2024ig2} and geometric optimization~\cite{simpson2025tangentially} optimize the baseline, while Shapley-inspired sampling~\cite{tachikawa2018compensated, liu2023new} eliminates baseline dependence. Closest to our work, IDGI~\cite{yang2023idgi} weights gradient contributions by $g_i^2$, demonstrating that squared-gradient weighting suppresses attribution noise relative to IG.

A complementary line of work tailors attribution to specific architectures. AttnLRP~\cite{achtibat2024attnlrp} derives relevance propagation rules for Transformer attention, OMENN~\cite{wrobel2024omenn} represents piecewise-linear networks as a single input-dependent matrix, DAVE~\cite{wrobel2026dave} scores ViT attributions through a distribution-aware gradient decomposition, and LibraGrad~\cite{mehri2025libragrad} balances Transformer gradient flow to improve all gradient-based attributions. Such modified backward computations are known to require careful interpretation~\cite{sixt2020explanations}. ERIG is instead architecture-agnostic; it uses the exact autograd gradients of any differentiable model, so its theoretical guarantees hold for the deployed model as is. LibraGrad in particular composes with ERIG, correcting how gradients are computed while ERIG weights these corrected gradients. We leave this combination to future work.

%% file: sec/3_prelim.tex
\section{Background}\label{sec: background}

\subsection{A General Framework}\label{sec: path-attrib}
Let $f : \mathbb{R}^N \to \mathbb{R}$ be a neural network, and let $\Gamma : [0,1] \to \mathbb{R}^N$ be an integration path from baseline $\bm{x}^b = \Gamma(0)$ to input $\bm{x}^a = \Gamma(1)$. At each path point~$t$, the gradient vector is $\bm{g}(t) = \nabla f(\Gamma(t))$, with components $g_i(t) = \frac{\partial f}{\partial x_i}\big|_{\Gamma(t)}$. For a weighting function $\bm{w} : \mathbb{R}^N \to \Delta^{N-1}$ mapping the gradient vector to the $N$-dimensional probability simplex, the attribution for feature $i$, inspired by path-based methods is,
\begin{equation}\label{eq:path-attrib}
    \phi_i = \int_\Gamma w_i(\bm{g}(t))\, df(t)
    \;\approx\;
    \sum_{k=1}^K w_i\!\bigl(\bm{g}(\bm{x}^{(k)})\bigr)\,\delta\!f^{(k)},
\end{equation}
where $\{\bm{x}^{(k)}\}_{k=0}^K$ discretizes $\Gamma$ and $\delta\!f^{(k)} = f(\bm{x}^{(k)}) - f(\bm{x}^{(k-1)})$. Different choices of $\bm{w}$ yield different methods; Integrated Gradients (IG)~\cite{sundararajan2017axiomatic} approximately corresponds to $w_i \propto g_i$ and Important Direction Gradient Integration (IDGI)~\cite{yang2023idgi} to $w_i \propto g_i^2$. Here one can ask: \emph{what is the optimal choice of $\bm{w}$?}

\subsection{Indirect Gradient Weighting}\label{sec: why-squared}
Both IG and IDGI couple attribution weights directly to the gradient, leaving no controllable mechanism to attenuate irregularities or noise in the gradient field. Guided IG~\cite{kapishnikov2021guided} takes the opposite approach, reducing attribution noise by adaptively steering the integration path away from high-gradient regions; however, this discards potentially informative gradient signal along with the noise. A key observation is that introducing a \emph{weighting function} $\bm{w}(\bm{g})$ that maps the gradient vector to the simplex adds a layer of indirectness between the raw gradient field and the final attribution. This creates a design space that controls how much gradient sensitivity propagates into the attributions. In our specific design, a single temperature $\tau$ governs this trade-off.

%% file: sec/4_method.tex
\section{Entropy-Regularized Integrated Gradients} \label{sec: methodology}

\subsection{Maximum-Entropy Sensitivity Weighting}\label{sec: max-ent}
Given the indirect weighting framework established in \cref{sec: background}, we seek a weighting scheme $\bm{w}$ that, for every point on the integration path, is maximally unbiased (maximizes entropy~\cite{jaynes1957information}) while still concentrating weight on locally sensitive features. Formally, we impose the constraint $\sum_{i=1}^N w_i\, g_i^2 \geq \gamma$, keeping the expected local sensitivity under $\bm{w}$ above a threshold. The choice of $g_i^2$ follows derivative-based global sensitivity measures~\cite{sobol2009derivative}. Locally, a first-order Taylor expansion shows that a small unbiased perturbation of feature $i$ induces output variance proportional to $(\partial f/\partial x_i)^2$ (Appendix~\ref{appx: perturb-gradient-proof}), making $g_i^2$ a natural local sensitivity measure. We then have,

\begin{proposition}[Maximum-entropy Weights]\label{theorem: maxent-weights}
  For the point-wise entropy-maximization problem at any point on $\Gamma(t)$,
  \begin{equation}
      \bm{w}^{*} = \argmax_{\bm{w} \in \Delta^{N-1}}
      \left(-\sum_{i=1}^N w_i \ln w_i\right)
      \ \text{s.t.} \
      \sum_{i=1}^N w_i\, g_i^2 \geq \gamma.
  \end{equation}
  The unique optimum solution is
  \begin{equation}\label{eq:max-ent}
      w_i^{*} = \frac{\exp(g_i^2/\tau)}
      {\sum_{j=1}^N \exp(g_j^2/\tau)}
      = \Softmax_i\!\left(\frac{\nabla f \odot \nabla f}{\tau}\right),
  \end{equation}
  where $\tau > 0$ satisfies
  $\sum_{i=1}^N w_i^{*}\, g_i^2 = \gamma$.
\end{proposition}
\begin{proof}
See Appendix~\ref{appx: maxent-proof}.
\end{proof}

\begin{figure}
    \centering
    \includegraphics[width=0.9\linewidth]{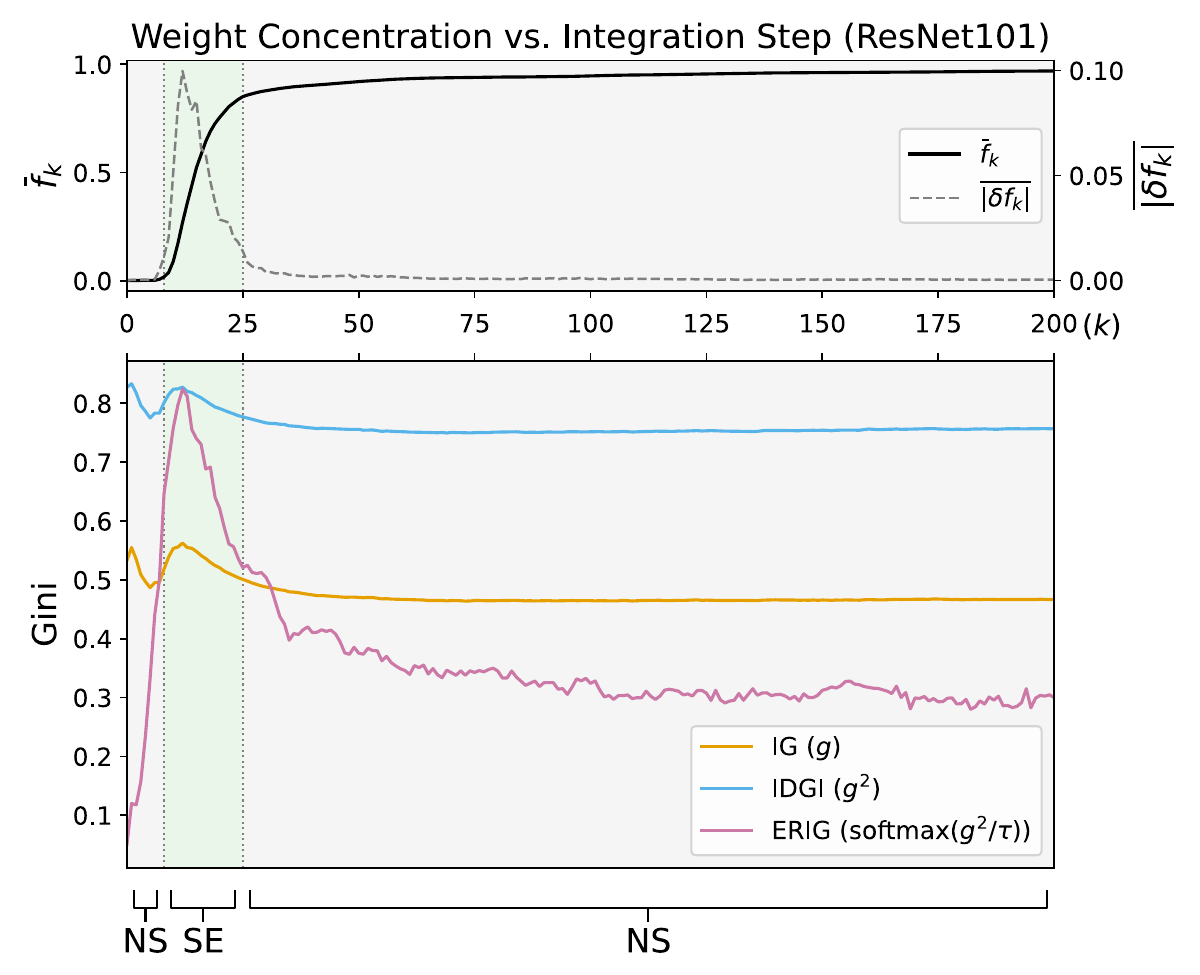}
    \caption{Weight concentration along the integration path (100 ImageNet validation images, $K{=}200$ steps). Top, mean output $\bar f_k$ and increment magnitude $\overline{|\delta f_k|}$; bottom, mean Gini coefficient of the per-step weight $(\bm{w}(\bm{g}))$ maps. Only ERIG ($\tau=3$) suppresses noise over flat and saturated segments (NS) and enhances signal over the informative segment (SE).}
    \label{fig: weight-regions}
    \vspace{-3mm}
\end{figure}

\noindent In \cref{eq:max-ent}, $\odot$ denotes the elementwise (Hadamard) product, and the subscript in $\Softmax_i$ selects the $i$-th component of the softmax output. Then ERIG attribution for feature $i$ is,
\begin{align}\label{eq:erig-formula}
    \mathrm{ERIG}_i(\Gamma;\tau) &= \int_\Gamma \Softmax_i\!\left(\frac{\nabla f \odot \nabla f}{\tau}\right) df.
\end{align}
In practice, $\tau$ controls the concentration of the weights. Smaller $\tau$ focuses attribution on the few features with the largest gradients; larger $\tau$ spreads the attribution more uniformly losing discriminability.\footnote{On the non-negativity of the weights, see Appendix~\ref{appx: serig}.} While Proposition~\ref{theorem: maxent-weights} is a per-path-point optimization, making $\tau$ in principle a per-point parameter $\tau(t)$, we make the simplifying choice of a constant $\tau$; Appendix~\ref{appx: aerig} ablates a path-dependent schedule for comparison.  See Appendix~\ref{appx: erig-algorithm} for ERIG algorithm.
\vspace{-0.5cm}
\paragraph{Intrinsic Properties.}
The max-entropy formulation endows ERIG with attribution stability and improved localization, and \Cref{fig: weight-regions} makes its two mechanisms directly observable by tracking each method's weight concentration along the integration path. The movement of the model output splits the path into an initial flat segment, a region of rapid output change, and a saturation plateau, and ERIG is the only method whose concentration follows this structure. Over the flat and saturated segments, where gradient spikes are noisier, entropy regularization~\cite{peyre2019computational, chaudhari2019entropy} keeps its weights closest to uniform, suppressing the noise (NS) that purely gradient-proportional schemes propagate unchecked. Where the output changes rapidly and the class evidence lies, the sensitivity constraint $\gamma$ concentrates them to IDGI's level, enhancing the signal (SE). The resulting attributions (\cref{fig: illustrative-examples}) confirm this qualitatively, with noise visibly suppressed and discriminative features highlighted. Further, \cref{sec: ablations} validates them quantitatively via ablations. These properties are intrinsic to the method and add no model evaluations beyond the $K$ path points; ensemble methods like SmoothGrad~\cite{smilkov2017smoothgrad} are instead extrinsic, needing $S \times K$ evaluations, where $S$ is the number of noise samples.

\subsection{Decomposition of Attribution Perturbation}\label{sec: lipschitz-decomp}

We now formalize how gradient perturbations propagate into the ERIG attribution. The key insight is that the attribution change under an input perturbation decomposes into two components, only one of whose bounds depends on $\tau$. We adopt two standard assumptions,

\begin{assumption}[Lipschitz model]\label{as:lip-f}
$f$ is $L_f$-Lipschitz: $|f(\bm{x}') - f(\bm{x})| \le L_f \|\bm{x}' - \bm{x}\|_2$ for all $\bm{x}, \bm{x}' \in \mathbb{R}^N$.
\end{assumption}

\begin{assumption}[Lipschitz gradients]\label{as:lip-g}
$\nabla f$ is $L_g$-Lipschitz: $\|\nabla f(\bm{x}') - \nabla f(\bm{x})\|_2 \le L_g \|\bm{x}' - \bm{x}\|_2$ for all $\bm{x}, \bm{x}' \in \mathbb{R}^N$.\footnote{For a discussion on ReLU-based networks, see Appendix~\ref{appx: relu-scope}.}
\end{assumption}

Consider a straight-line path $\Gamma(t) = \bm{x}^b + t(\bm{x}^a - \bm{x}^b)$ for $t \in [0,1]$. Under perturbation $\bm{x}^a \to \tilde{\bm{x}}^a = \bm{x}^a + \bm{\eta}$, the perturbed path is $\tilde{\Gamma}(t) = \Gamma(t) + t\bm{\eta}$, and the attribution change, setting aside $\mathcal{O}(\|\bm{\eta}\|^2_2)$ terms for a small enough $\|\bm{\eta}\|_2 \ll 1$, decomposes via the product rule as
\begin{equation}\label{eq:delta-erig}
  \Delta\mathrm{ERIG}_i
  \approx \bigintss_0^1
    \left[
      \begin{aligned}
        &\underbrace{w_i^{*}(\bm{g}(t);\tau)\;\Delta\dot{f}(t)}_{\text{model-response}} \\
        &\quad +\; \underbrace{\Delta w_i^{*}(\bm{g}(t);\tau)\;\dot{f}(t)}_{\text{weight-variation}}
      \end{aligned}
    \right] dt,
\end{equation}
where $\Delta w_i^{*} = w_i^{*}(\bm{g}(\tilde{\Gamma}(t));\tau) - w_i^{*}(\bm{g}(\Gamma(t));\tau)$, and $\dot{f}(t) = df(\Gamma(t))/dt$ is the function increment rate along the path, with $\Delta\dot{f}(t) = \dot{f}(\tilde{\Gamma}(t)) - \dot{f}(\Gamma(t))$ as its change under the perturbation. Please note that $df = \dot{f}dt$ in \cref{eq:erig-formula}.

The \emph{model-response} term captures the change in function increment rate along the perturbed path. The \emph{weight-variation} term captures how gradient changes induced by the input perturbation propagate through the weighting scheme, redistributing the original function increments across features. Let $\bm{M}, \bm{W} \in \mathbb{R}^N$ denote the model-response and weight-variation vectors with components $M_i = \int_0^1 w_i^{*}\,\Delta\dot{f}\,dt$ and $W_i = \int_0^1 \Delta w_i^{*}\,\dot{f}\,dt$, so that $\Delta\mathrm{ERIG} \approx \bm{M} + \bm{W}$. \Cref{cor: decomp} bounds these components:

\begin{theorem}[Selective Attenuation]\label{cor: decomp}
Let $|\Gamma| = \|\bm{x}^a - \bm{x}^b\|_2$ denote the path length. Under Assumptions~\ref{as:lip-f} and~\ref{as:lip-g},
\begin{equation}\label{eq:decomp-bound}
\begin{split}
    \|\Delta\mathrm{ERIG}\|_2
    &\le \|\bm{M}\|_2 + \|\bm{W}\|_2\\
    &\le
    \underbrace{\left(L_f + \tfrac{1}{2}\,L_g\,|\Gamma|\right)\|\bm{\eta}\|_2}_{\text{bound on }\|\bm{M}\|_2}
    \;\cdot\;\bigl(1 + R(\tau)\bigr),
\end{split}
\end{equation}
where
\begin{equation}\label{eq:wv-ratio}
  R(\tau) = \frac{L_f^2\,L_g\,|\Gamma|}{2\tau\,\bigl(L_f + \tfrac{1}{2}\,L_g\,|\Gamma|\bigr)} + \mathcal{O}(\|\bm{\eta}\|_2).
\end{equation}
Here $L_f^2\,L_g\,|\Gamma|\,\|\bm{\eta}\|_2/(2\tau) + \mathcal{O}(\|\bm{\eta}\|_2^2)$ is the bound on the weight-variation term $\|\bm{W}\|_2$.
\end{theorem}
\begin{proof}
See Appendix~\ref{appx: lipschitz-proof}.
\end{proof}

The model-response bound represents the attribution change a faithful method should reflect, and $R(\tau)$ quantifies the additional perturbation the gradient-dependent weighting introduces relative to it. Both components are informative; the model response tracks the network's sensitivity to the perturbation, while the weight variation carries the gradient information that enables feature-level discrimination. The goal is therefore to choose $\tau$ so that the gradient signal is preserved without overwhelming the model response with gradient irregularities.

The theorem therefore establishes the mechanism by which $\tau$ controls weight-variation relative to the model response and the direction of this control. However, it cannot prescribe an optimal $\tau$ value, because $L_f$ and $L_g$ are generally intractable for deep networks~\cite{virmaux2018lipschitz,fazlyab2019efficient} and the optimal balance between gradient information and stability is context-dependent (\cref{fig: weight-regions}). We therefore determine $\tau$ empirically --- see ablations in \Cref{sec: ablations}. An exception to this are 1-Lipschitz networks~\cite{serrurier2023explainable}, whose architecturally controlled constants would make $R(\tau)$ computable and $\tau$ analytically settable, a direction we leave to future work.

By contrast, for methods such as IG where $w_i \approxprop g_i$, gradient perturbations pass directly into the weights with an $\mathcal{O}(1)$ Lipschitz constant and the corresponding ratio is fixed with no mechanism to control it. Moreover, integrating the weighted contributions over the full path further averages out residual gradient fluctuations, yielding even greater stability in practice than the $\mathcal{O}(1/(2\tau))$ attenuation alone suggests. \Cref{subsec: stability} experimentally verifies this behaviour.

%% file: sec/5_experiments.tex
\section{Experimental Evaluation}

\subsection{Experimental Setup}
\begin{figure}
    \centering
    \includegraphics[width=0.85\linewidth]{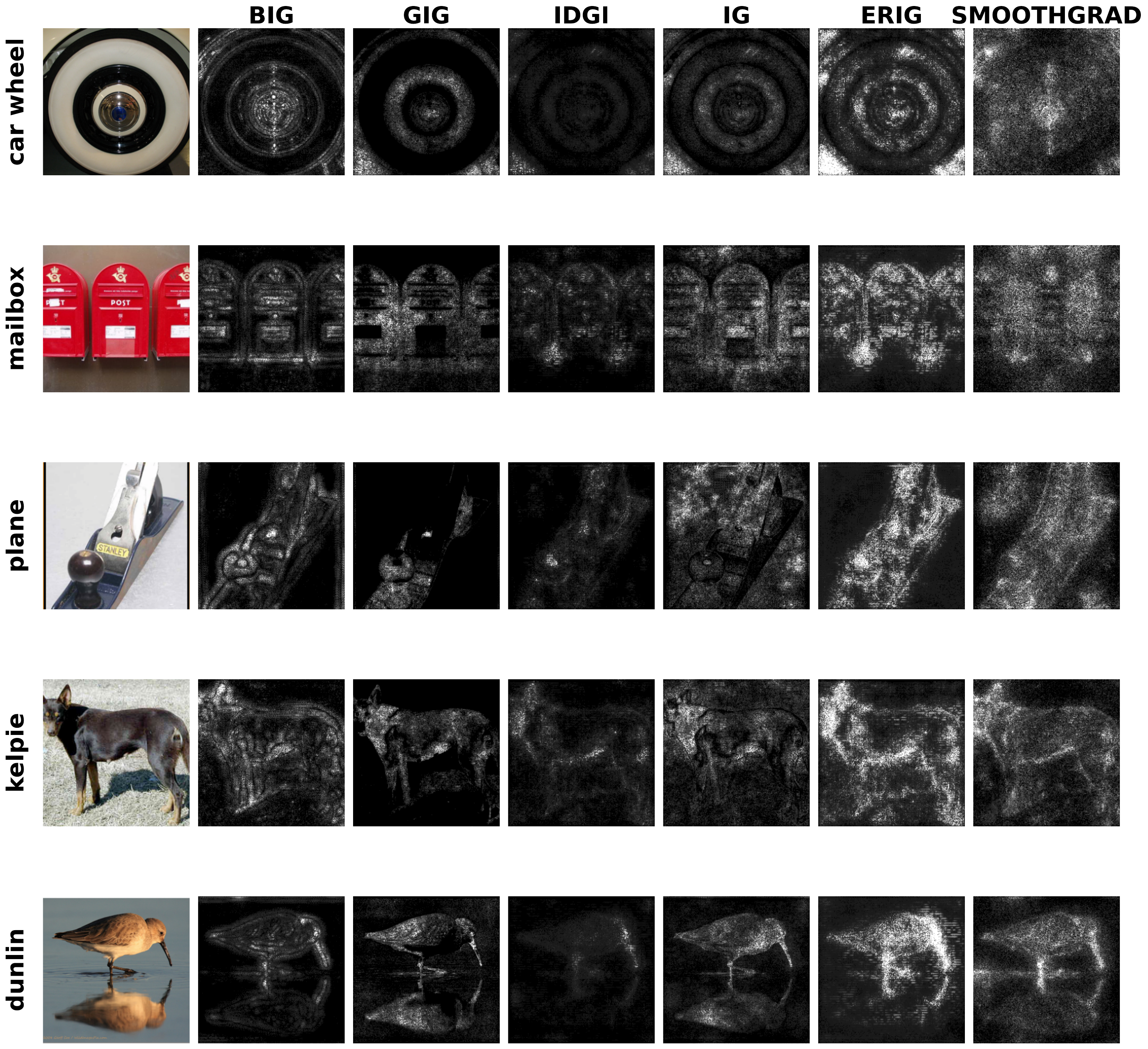}
    \caption{MobileNetV4 Visualizations --- More examples in \ref{appx: viz-examples}}
    \label{fig: illustrative-examples}
    \vspace{-3mm}
\end{figure}
We evaluate the explanation performance of our approach across diverse architectures, namely, EfficientNet~\cite{tan2019efficientnet}, MobileNetV4~\cite{qin2024mobilenetv4}, ResNet101~\cite{he2016deep, He2018BagOT}, EVA-02~\cite{fang2024eva}, DeiT3~\cite{Touvron2022DeiTIR}, and ViT-Base~\cite{wu2020visual}. Following~\cite{yang2023idgi, kapishnikov2019xrai, kapishnikov2021guided, pan2021explaining, sundararajan2017axiomatic}, all experiments are performed on randomly sampled subsets of the ImageNet~\cite{imagenet15russakovsky} and ImageNet-S~\cite{gao2022luss} validation datasets using the \textit{timm}~\cite{rw2019timm} Python library.

We compare ERIG against four established path-based attribution techniques: Integrated Gradients (IG)~\cite{sundararajan2017axiomatic}, Important Direction Gradient Integration (IDGI)~\cite{yang2023idgi}, Blur Integrated Gradients (BIG)~\cite{xu2020attribution}, Guided Integrated Gradients (GIG)~\cite{kapishnikov2021guided}, and additionally SmoothGrad~\cite{smilkov2017smoothgrad}. For all path-based methods, we use 200 integration steps with a black image baseline following standard practice. For ERIG, we sweep the temperature from $10^{-3}$ to $10^{3}$, with full ablations in~\cref{sec: ablations}; while the optimum temperature can vary across architectures, in practice we find that $\tau = 3$ is effective across multiple tested models. For SmoothGrad, we use a 10\% noise level and 50 noise samples, as prescribed.

Computationally, all the above path-based methods scale linearly with integration steps, but ERIG, IG, and IDGI have the same order of per-step runtime, whereas BIG and GIG incur substantially higher per-step overhead due to expensive gaussian blurring and sequential path construction, respectively. Comprehensive statistics in Appendix~\ref{appx: runtime}.

Our evaluation employs a comprehensive suite of metrics assessing four key properties of attribution maps: (1)~\textit{fidelity}; (2)~\textit{localization}; (3)~\textit{attribution stability}; and (4)~\textit{noise characteristics}. Throughout our results, we use arrows to indicate whether higher ($\uparrow$) or lower ($\downarrow$) values are preferable. In all tables, the best-performing method is \underline{\textbf{bold \& underlined}}, while the second-best is \textbf{bolded}.

\subsection{Faithfulness and Fidelity} \label{subsec: fidelity}
\begin{table}
    \centering
    \caption{\textbf{ImageNet Fidelity:} ERIG places first or second on every metric and backbone, with a $+0.44\%$ average lead.}
    \scriptsize
    \begin{tabular}{|c|c|c|c|c|}
        \hline
        \textbf{Model} & \textbf{Method} & \textbf{ISC $(\uparrow)$} & \textbf{AIC $(\uparrow)$} & \textbf{DEL $(\downarrow)$}\\
        \hline
        \multirow{6}{*}{ResNet101} & BIG & $0.3101$ & $0.6348$ & $0.4304$\\
        & GIG & $0.3812$ & $0.5500$ & $0.5765$\\
        & IDGI & $\mathbf{\underline{0.4923}}$ & $0.6571$ & $\mathbf{\underline{0.3211}}$\\
        & IG & $0.3951$ & $\mathbf{\underline{0.6838}}$ & $0.3884$\\
        & ERIG (ours) & $\mathbf{0.4893}$ & $\mathbf{0.6610}$ & $\mathbf{0.3270}$\\
        & SmoothGrad & $0.1833$ & $0.4011$ & $0.5138$\\
        \hline
        \multirow{6}{*}{EfficientNet} & BIG & $0.3720$ & $0.8241$ & $0.4572$\\
        & GIG & $0.3818$ & $0.7061$ & $0.5604$\\
        & IDGI & $\mathbf{0.4380}$ & $\mathbf{\underline{0.8401}}$ & $\mathbf{\underline{0.4310}}$\\
        & IG & $0.3590$ & $0.7882$ & $0.4774$\\
        & ERIG (ours) & $\mathbf{\underline{0.4521}}$ & $\mathbf{0.8323}$ & $\mathbf{0.4562}$\\
        & SmoothGrad & $0.2523$ & $0.6122$ & $0.5440$\\
        \hline
        \multirow{6}{*}{MobileNetV4} & BIG & $0.1792$ & $0.6670$ & $0.4574$\\
        & GIG & $\mathbf{0.3561}$ & $0.5837$ & $0.5630$\\
        & IDGI & $0.2831$ & $\mathbf{0.7223}$ & $\mathbf{0.3643}$\\
        & IG & $0.1688$ & $0.6023$ & $0.4673$\\
        & ERIG (ours) & $\mathbf{\underline{0.4118}}$ & $\mathbf{\underline{0.7234}}$ & $\mathbf{\underline{0.3626}}$\\
        & SmoothGrad & $0.0765$ & $0.4087$ & $0.5598$\\
        \hline
        \multirow{6}{*}{ViT-Base} & BIG & $0.5689$ & $0.8273$ & $0.6543$\\
        & GIG & $0.4932$ & $0.6827$ & $0.7949$\\
        & IDGI & $\mathbf{0.6917}$ & $\mathbf{\underline{0.9057}}$ & $\mathbf{0.4942}$\\
        & IG & $0.5354$ & $0.8154$ & $0.6485$\\
        & ERIG (ours) & $\mathbf{\underline{0.6930}}$ & $\mathbf{0.9013}$ & $\mathbf{\underline{0.4902}}$\\
        & SmoothGrad & $0.3674$ & $0.5796$ & $0.7975$\\
        \hline
        \multirow{6}{*}{EVA-02} & BIG & $0.5372$ & $0.8420$ & $0.7106$\\
        & GIG & $0.5522$ & $0.8051$ & $0.8407$\\
        & IDGI & $\mathbf{0.5876}$ & $\mathbf{\underline{0.8978}}$ & $\mathbf{0.5423}$\\
        & IG & $0.5614$ & $0.8735$ & $0.7255$\\
        & ERIG (ours) & $\mathbf{\underline{0.5913}}$ & $\mathbf{0.8907}$ & $\mathbf{\underline{0.5373}}$\\
        & SmoothGrad & $0.3851$ & $0.6876$ & $0.8279$\\
        \hline
        \multirow{6}{*}{DeiT3} & BIG & $0.5280$ & $0.7411$ & $0.6939$\\
        & GIG & $0.5982$ & $0.7208$ & $0.7844$\\
        & IDGI & $\mathbf{\underline{0.7343}}$ & $\mathbf{\underline{0.9074}}$ & $\mathbf{0.5621}$\\
        & IG & $0.6437$ & $0.8514$ & $0.6794$\\
        & ERIG (ours) & $\mathbf{0.7320}$ & $\mathbf{0.9010}$ & $\mathbf{\underline{0.5561}}$\\
        & SmoothGrad & $0.4292$ & $0.6066$ & $0.7780$\\
        \hline
    \end{tabular}
    \label{tab: auc-results}
\end{table}
We follow prior work~\cite{yang2023idgi} and adopt three metrics: Area Under the Insertion Score Curve (ISC)~\cite{pan2021explaining}, the Accuracy Information Curve (AIC)~\cite{kapishnikov2019xrai,yang2023idgi}, and Iterative Removal of Features (DEL)~\cite{rieger2020irof}. ISC reveals pixels in descending attribution order and records the rise in model confidence. AIC restores pixels from a blurred baseline and tracks accuracy against image information content (MS-SSIM~\cite{wang2003multiscale}). DEL removes pixels in descending attribution order and tracks the decline in confidence. Higher AUC is better for ISC and AIC, lower for DEL. All three depend on the relative ranking of attribution values, so they are insensitive to attribution magnitude and to noise that does not alter the ranking. As shown in \cref{tab: auc-results}, ERIG stays comparable to the strongest baselines, with an average lead of only $+0.44\%$ across all models and metrics. The larger gains on localization, stability, and noise are consistent with this \emph{rank-only limitation}, since those metrics measure properties that ranking cannot capture. Metric distributions in Appendix~\ref{appx: fidelity-distr} further show substantially overlapping inter-quartile ranges between ERIG and the strongest baseline, confirming this behavior.

\subsection{Weakly Supervised Segmentation}  \label{subsec: localization}
To evaluate the spatial correspondence between attributions and object regions, we follow previous literature~\cite{yang2023idgi, hedstrom2023quantus, chattopadhay2018grad} and design a weakly supervised segmentation task with ImageNet-S~\cite{gao2022luss} annotations as reference. We normalize attributions to $[0,1]$, binarize them at thresholds $t \in \{0.05, 0.10, \ldots, 0.95\}$, and compare each binary mask against the reference mask pixel-wise. We report Intersection over Union (IoU), Pixel Accuracy (PA), and Pixel F1 Score (PF1), averaged across all thresholds. As shown in \Cref{tab: localization-results}, ERIG outperforms the second-best method by ${\sim}28\%$ on average across model-metric combinations, showcasing its effective localization capability. Unlike the fidelity metrics, the localization distributions in Appendix~\ref{appx: localization-distr} differentiate the methods clearly, displaying a visible separation between ERIG and its baselines.
\begin{table}
    \centering
    \scriptsize
    \caption{\textbf{ImageNet-S Localization:} ERIG improves localization by ${\sim}28\%$ on average over the next-best method.}
    \begin{tabular}{|c|c|c|c|c|}
        \hline
        \textbf{Model} & \textbf{Method} & \textbf{IoU $(\uparrow)$} & \textbf{PA $(\uparrow)$} & \textbf{PF1 $(\uparrow)$}\\
        \hline
        \multirow{6}{*}{ResNet101} & BIG & $0.0643$ & $0.6386$ & $0.1008$\\
        & GIG & $0.0450$ & $0.6286$ & $0.0718$\\
        & IDGI & $0.0212$ & $0.6319$ & $0.0354$\\
        & IG & $0.0782$ & $0.6401$ & $0.1200$\\
        & ERIG (ours) & $\mathbf{\underline{0.1617}}$ & $\mathbf{\underline{0.6580}}$ & $\mathbf{\underline{0.2419}}$\\
        & SmoothGrad & $\mathbf{0.1286}$ & $\mathbf{0.6498}$ & $\mathbf{0.1824}$\\
        \hline
        \multirow{6}{*}{EfficientNet} & BIG & $0.0629$ & $\mathbf{0.6034}$ & $0.0982$\\
        & GIG & $0.0448$ & $0.5959$ & $0.0713$\\
        & IDGI & $0.0120$ & $0.5937$ & $0.0206$\\
        & IG & $0.0639$ & $0.5955$ & $0.0999$\\
        & ERIG (ours) & $\mathbf{\underline{0.1085}}$ & $0.5982$ & $\mathbf{\underline{0.1714}}$\\
        & SmoothGrad & $\mathbf{0.1012}$ & $\mathbf{\underline{0.6104}}$ & $\mathbf{0.1474}$\\
        \hline
        \multirow{6}{*}{MobileNetV4} & BIG & $0.0681$ & $0.6357$ & $0.1062$\\
        & GIG & $0.0501$ & $0.6269$ & $0.0797$\\
        & IDGI & $0.0305$ & $0.6305$ & $0.0501$\\
        & IG & $0.0834$ & $0.6304$ & $0.1286$\\
        & ERIG (ours) & $\mathbf{\underline{0.1615}}$ & $\mathbf{\underline{0.6521}}$ & $\mathbf{\underline{0.2424}}$\\
        & SmoothGrad & $\mathbf{0.1356}$ & $\mathbf{0.6501}$ & $\mathbf{0.1921}$\\
        \hline
        \multirow{6}{*}{ViT-Base} & BIG & $0.0656$ & $0.6980$ & $0.1023$\\
        & GIG & $0.0371$ & $0.6946$ & $0.0594$\\
        & IDGI & $0.0073$ & $0.6959$ & $0.0127$\\
        & IG & $0.0537$ & $0.6940$ & $0.0843$\\
        & ERIG (ours) & $\mathbf{\underline{0.1203}}$ & $\mathbf{0.7054}$ & $\mathbf{\underline{0.1893}}$\\
        & SmoothGrad & $\mathbf{0.1150}$ & $\mathbf{\underline{0.7113}}$ & $\mathbf{0.1635}$\\
        \hline
        \multirow{6}{*}{EVA-02} & BIG & $0.0540$ & $\mathbf{0.6958}$ & $0.0856$\\
        & GIG & $0.0396$ & $0.6901$ & $0.0635$\\
        & IDGI & $0.0074$ & $0.6947$ & $0.0128$\\
        & IG & $0.0529$ & $0.6846$ & $0.0838$\\
        & ERIG (ours) & $\mathbf{\underline{0.1317}}$ & $\mathbf{\underline{0.7054}}$ & $\mathbf{\underline{0.2031}}$\\
        & SmoothGrad & $\mathbf{0.1038}$ & $0.6952$ & $\mathbf{0.1526}$\\
        \hline
        \multirow{6}{*}{DeiT3} & BIG & $\mathbf{0.0470}$ & $\mathbf{0.6511}$ & $\mathbf{0.0746}$\\
        & GIG & $0.0173$ & $0.6460$ & $0.0279$\\
        & IDGI & $0.0067$ & $0.6476$ & $0.0116$\\
        & IG & $0.0206$ & $0.6467$ & $0.0329$\\
        & ERIG (ours) & $\mathbf{\underline{0.1179}}$ & $\mathbf{\underline{0.6637}}$ & $\mathbf{\underline{0.1819}}$\\
        & SmoothGrad & $0.0296$ & $0.6451$ & $0.0438$\\
        \hline
    \end{tabular}
    \label{tab: localization-results}
\end{table}

\subsection{Attribution Stability} \label{subsec: stability}
In deployment, attribution maps should remain stable under small input perturbations. We evaluate this with three established metrics, Max Sensitivity (MS)~\cite{yeh2019fidelity}, Local Lipschitz Constant (LLC)~\cite{alvarez2018robustness}, and Relative Input Stability (RIS)~\cite{agarwal2022rethinking}. These quantify the worst-case attribution shift within an $\ell_\infty$ ball of radius $r$; lower is more stable.
\begin{align}
\operatorname{MS}(\bm{x},r) &= \max_{\lVert \bm{x}-\bm{x}' \rVert_\infty \le r} 
    \lVert \mathbf{e}_{\bm{x}} - \mathbf{e}_{\bm{x}'} \rVert_2 \label{eq:ms} \\
\operatorname{LLC}(\bm{x},r) &= \max_{\lVert \bm{x}-\bm{x}' \rVert_\infty \le r} 
    \frac{\lVert \mathbf{e}_{\bm{x}} - \mathbf{e}_{\bm{x}'} \rVert_2}
         {\lVert \bm{x} - \bm{x}' \rVert_2} \label{eq:llc} \\
\operatorname{RIS}(\bm{x},r) &= \max_{\lVert \bm{x}-\bm{x}' \rVert_\infty \le r} 
    \frac{\lVert \mathbf{e}_{\bm{x}} - \mathbf{e}_{\bm{x}'} \rVert_2 / \lVert \mathbf{e}_{\bm{x}} \rVert_2}
         {\lVert \bm{x} - \bm{x}' \rVert_2 / \lVert \bm{x} \rVert_2} \label{eq:ris}
\end{align}
 Following their protocols~\cite{yeh2019fidelity, alvarez2018robustness, agarwal2022rethinking}, we approximate each supremum via Monte Carlo sampling of 50 isotropic uniform perturbations per image at $r = 0.1$ over 100 ImageNet validation samples; isotropic perturbations introduce no directional bias and assess local smoothness uniformly across all input directions. Appendix~\ref{appx: astability} complements this protocol with adversarially selected perturbation directions, under which ERIG's stability advantage carries over unchanged. As reported in~\cref{tab: stability-results}, ERIG outperforms the best baseline method by \emph{up to two orders of magnitude}, confirming its superior attribution map stability. Log-scale plots with inter-quartile ranges in the Appendix~\ref{appx: stability-distr} further confirm their significance.
\begin{table}
    \centering
    \scriptsize
    \caption{\textbf{ImageNet Stability:} ERIG improves stability by ${\sim}55\times$ on average over the next-best method.}
    \begin{tabular}{|c|c|c|c|c|}
        \hline
        \textbf{Model} & \textbf{Method} & \textbf{MS $(\downarrow)$} & \textbf{LLC $(\downarrow)$} & \textbf{RIS $(\downarrow)$}\\
        \hline
        \multirow{6}{*}{ResNet101} & BIG & $0.1750$ & $0.0078$ & $\mathbf{12.498}$\\
        & GIG & $0.3723$ & $0.0166$ & $88.373$\\
        & IDGI & $\mathbf{0.0100}$ & $\mathbf{0.0004}$ & $18.653$\\
        & IG & $0.6872$ & $0.0307$ & $30.295$\\
        & ERIG (ours) & $\mathbf{\underline{8.6e\text{\textbf{ --}}05}}$ & $\mathbf{\underline{3.8e\text{\textbf{ --}}06}}$ & $\mathbf{\underline{1.066}}$\\
        & SmoothGrad & $0.1391$ & $0.0062$ & $102.340$\\
        \hline
        \multirow{6}{*}{EfficientNet} & BIG & $0.0316$ & $0.0010$ & $\mathbf{6.318}$\\
        & GIG & $0.2300$ & $0.0072$ & $23.019$\\
        & IDGI & $\mathbf{0.0040}$ & $\mathbf{0.0001}$ & $12.666$\\
        & IG & $0.3463$ & $0.0108$ & $18.834$\\
        & ERIG (ours) & $\mathbf{\underline{6.7e\text{\textbf{ --}}05}}$ & $\mathbf{\underline{2.1e\text{\textbf{ --}}06}}$ & $\mathbf{\underline{1.271}}$\\
        & SmoothGrad & $0.0458$ & $0.0014$ & $16.972$\\
        \hline
        \multirow{6}{*}{MobileNetV4} & BIG & $0.1638$ & $0.0064$ & $\mathbf{9.863}$\\
        & GIG & $0.5514$ & $0.0215$ & $30.213$\\
        & IDGI & $\mathbf{0.0051}$ & $\mathbf{0.0002}$ & $15.979$\\
        & IG & $0.9021$ & $0.0352$ & $23.123$\\
        & ERIG (ours) & $\mathbf{\underline{1.0e\text{\textbf{ --}}04}}$ & $\mathbf{\underline{5.4e\text{\textbf{ --}}06}}$ & $\mathbf{\underline{2.196}}$\\
        & SmoothGrad & $0.1126$ & $0.0044$ & $24.148$\\
        \hline
        \multirow{6}{*}{ViT-Base} & BIG & $0.0652$ & $0.0017$ & $\mathbf{7.142}$\\
        & GIG & $0.2167$ & $0.0056$ & $20.346$\\
        & IDGI & $\mathbf{0.0093}$ & $\mathbf{0.0002}$ & $15.883$\\
        & IG & $0.4678$ & $0.0122$ & $20.596$\\
        & ERIG (ours) & $\mathbf{\underline{3.0e\text{\textbf{ --}}04}}$ & $\mathbf{\underline{9.0e\text{\textbf{ --}}06}}$ & $\mathbf{\underline{4.946}}$\\
        & SmoothGrad & $0.0452$ & $0.0012$ & $12.456$\\
        \hline
        \multirow{6}{*}{EVA-02} & BIG & $0.0483$ & $0.0011$ & $\mathbf{11.652}$\\
        & GIG & $0.1586$ & $0.0035$ & $21.562$\\
        & IDGI & $\mathbf{0.0053}$ & $\mathbf{0.0001}$ & $13.423$\\
        & IG & $0.6562$ & $0.0146$ & $134.160$\\
        & ERIG (ours) & $\mathbf{\underline{6.7e\text{\textbf{ --}}05}}$ & $\mathbf{\underline{1.5e{\textbf{ --}}06}}$ & $\mathbf{\underline{1.601}}$\\
        & SmoothGrad & $0.0438$ & $0.0010$ & $16.770$\\
        \hline
        \multirow{6}{*}{DeiT3} & BIG & $0.0967$ & $0.0025$ & $\mathbf{9.570}$\\
        & GIG & $0.1746$ & $0.0045$ & $49.933$\\
        & IDGI & $\mathbf{0.0078}$ & $\mathbf{0.0002}$ & $13.132$\\
        & IG & $0.6409$ & $0.0167$ & $42.144$\\
        & ERIG (ours) & $\mathbf{\underline{3.9e\text{\textbf{ --}}05}}$ & $\mathbf{\underline{1.0e\text{\textbf{ --}}06}}$ & $\mathbf{\underline{0.756}}$\\
        & SmoothGrad & $0.0380$ & $0.0010$ & $40.339$\\
        \hline
    \end{tabular}
    \label{tab: stability-results}
\end{table}

\subsection{Quantitative Estimation of Attribution Noise}  \label{subsec: noise-suppression}
\begin{table}
    \centering
    \scriptsize
    \caption{\textbf{ImageNet Noise:} ERIG attains the highest LINV on every backbone, $6.7$--$10.5$ LINV above the next best.}
    \begin{tabular}{|c|c|c|c|}
        \hline
        \textbf{Model} & \textbf{Method} & \textbf{LINV-C $(\uparrow)$} & \textbf{LINV-P $(\uparrow)$}\\
        \hline
        \multirow{6}{*}{ResNet101} & BIG & $6.6490$ & $6.6053$\\
        & GIG & $7.9654$ & $7.9093$\\
        & IDGI & $9.4308$ & $9.4017$\\
        & IG & $6.5869$ & $6.4886$\\
        & ERIG (ours) & $\mathbf{\underline{16.8069}}$ & $\mathbf{\underline{16.7781}}$\\
        & SmoothGrad & $\mathbf{10.1355}$ & $\mathbf{9.9179}$\\
        \hline
        \multirow{6}{*}{EfficientNet} & BIG & $7.9165$ & $8.0544$\\
        & GIG & $7.3640$ & $7.2324$\\
        & IDGI & $\mathbf{10.2429}$ & $\mathbf{10.2457}$\\
        & IG & $6.8766$ & $6.8567$\\
        & ERIG (ours) & $\mathbf{\underline{20.7048}}$ & $\mathbf{\underline{20.7121}}$\\
        & SmoothGrad & $8.7287$ & $8.7101$\\
        \hline
        \multirow{6}{*}{MobileNetV4} & BIG & $6.5056$ & $6.4759$\\
        & GIG & $6.5981$ & $6.4472$\\
        & IDGI & $\mathbf{10.0863}$ & $\mathbf{10.0732}$\\
        & IG & $5.9712$ & $5.7927$\\
        & ERIG (ours) & $\mathbf{\underline{18.1303}}$ & $\mathbf{\underline{18.1133}}$\\
        & SmoothGrad & $7.8476$ & $7.6511$\\
        \hline
        \multirow{6}{*}{ViT-Base} & BIG & $8.0362$ & $7.9448$\\
        & GIG & $8.0055$ & $7.9185$\\
        & IDGI & $\mathbf{10.1619}$ & $\mathbf{10.0085}$\\
        & IG & $7.2225$ & $7.0610$\\
        & ERIG (ours) & $\mathbf{\underline{19.3426}}$ & $\mathbf{\underline{19.2005}}$\\
        & SmoothGrad & $9.2599$ & $8.8763$\\
        \hline
        \multirow{6}{*}{EVA-02} & BIG & $8.4674$ & $8.4218$\\
        & GIG & $8.5047$ & $8.3316$\\
        & IDGI & $\mathbf{10.8027}$ & $\mathbf{10.7783}$\\
        & IG & $7.5677$ & $7.5228$\\
        & ERIG (ours) & $\mathbf{\underline{20.1285}}$ & $\mathbf{\underline{20.1150}}$\\
        & SmoothGrad & $9.5953$ & $9.1273$\\
        \hline
        \multirow{6}{*}{DeiT3} & BIG & $7.7467$ & $8.1019$\\
        & GIG & $8.6576$ & $8.7212$\\
        & IDGI & $\mathbf{10.2471}$ & $\mathbf{10.2542}$\\
        & IG & $7.1575$ & $7.5698$\\
        & ERIG (ours) & $\mathbf{\underline{19.5778}}$ & $\mathbf{\underline{19.5845}}$\\
        & SmoothGrad & $9.5920$ & $9.3614$\\
        \hline
    \end{tabular}
    \label{tab: noise-results}
\end{table}
A critical limitation of stability metrics such as MS, LLC, and RIS is their high computational cost, as they require multiple perturbed inputs and the corresponding attribution maps. For path-based methods with $T$ integration steps, evaluating 100 images with 50 perturbations each requires $50 \times 100 \times T$ forward and backward passes, reaching a \textit{million} evaluations for $T=200$. Since attribution noise and instability both arise from irregularities in the gradient field and sampling noise from the input image, attribution noise level serves as an efficient proxy for stability. We hence adapt two established image-noise estimators that directly quantify attribution noise through principled statistical estimation. These metrics also address a limitation of prior work, which largely assessed noise reduction qualitatively through visual inspection or anecdotal examples rather than direct quantitative measurement~\cite{yang2023idgi,kapishnikov2021guided}. These estimators enable objective and reproducible comparisons across attribution methods. However, we would like to categorically mention that they should be \emph{used along with} fidelity and localization metrics, as reduced noise and improved stability alone do not guarantee explanation faithfulness or accurate localization. In ERIG, we balance these multiple competing factors using an optimum temperature ($\tau$) value. For a comparison between LINV and other complexity, compressibility, and sparsity measures, see Appendix~\ref{appx: linv-validation}.
\vspace{-0.3cm}
\paragraph{Convolution-Based Noise Estimation (LINV-C)~\cite{immerkaer1996fast}}
\begin{equation}
\sigma^2
=
\frac{1}{36(H-2)(W-2)}
\sum_{i=2}^{H-1}
\sum_{j=2}^{W-1}
\bigl(A_{ij} \ast S\bigr)^2
\label{eq:linv-c1}
\end{equation}
LINV-C detects unstructured variations characteristic of random perturbations. In \cref{eq:linv-c1}, $A \in \mathbb{R}^{H \times W}$ is the attribution map, $\ast$ denotes convolution, and $S$ is the $3{\times}3$ noise-isolating filter with rows $(1,-2,1)$, $(-2,4,-2)$, $(1,-2,1)$; the resulting variance $\sigma^2$ estimates how unpredictably attribution values fluctuate across the map.

\vspace{-0.3cm}
\paragraph{PCA-Based Noise Estimation (LINV-P)~\cite{chen2015efficient}}
LINV-P exploits the separation between low-rank structure and Gaussian noise in local attribution patches. Each map is partitioned into non-overlapping patches $\mathbf{x}_t \in \mathbb{R}^r$, modeled as $\mathbf{x}_t = \mathbf{A}\mathbf{y}_t + \mathbf{e}_t$ with $\mathbf{e}_t \sim \mathcal{N}_r(0,\sigma^2\mathbf{I})$, where $\mathbf{A} \in \mathbb{R}^{r \times m}$ spans the principal subspace capturing coherent structure. The eigenvalues $\lambda_i$ of the empirical patch covariance $\mathbf{X}_s = (\mathbf{x}_t)_{t=1}^s$ separate into structure and noise components, with the noise eigenvalues asymptotically following $\lambda_i \sim \mathcal{N}(\sigma^2, 2\sigma^4/s)$ for $i \in [m+1, r]$. We iteratively remove dominant eigenvalues until the mean of the remaining ones equals their median, yielding a robust estimate of the noise variance.

Both estimators convert attribution noise into a quantitative, reproducible estimate. For interpretability, we report the \textbf{Log-Inverse Noise Variance} (LINV-X), defined as $-\log_{10}(\sigma^2)$. Here higher values indicate cleaner, more stable attribution maps.

\begin{figure}
    \centering
    \includegraphics[width=0.9\linewidth]{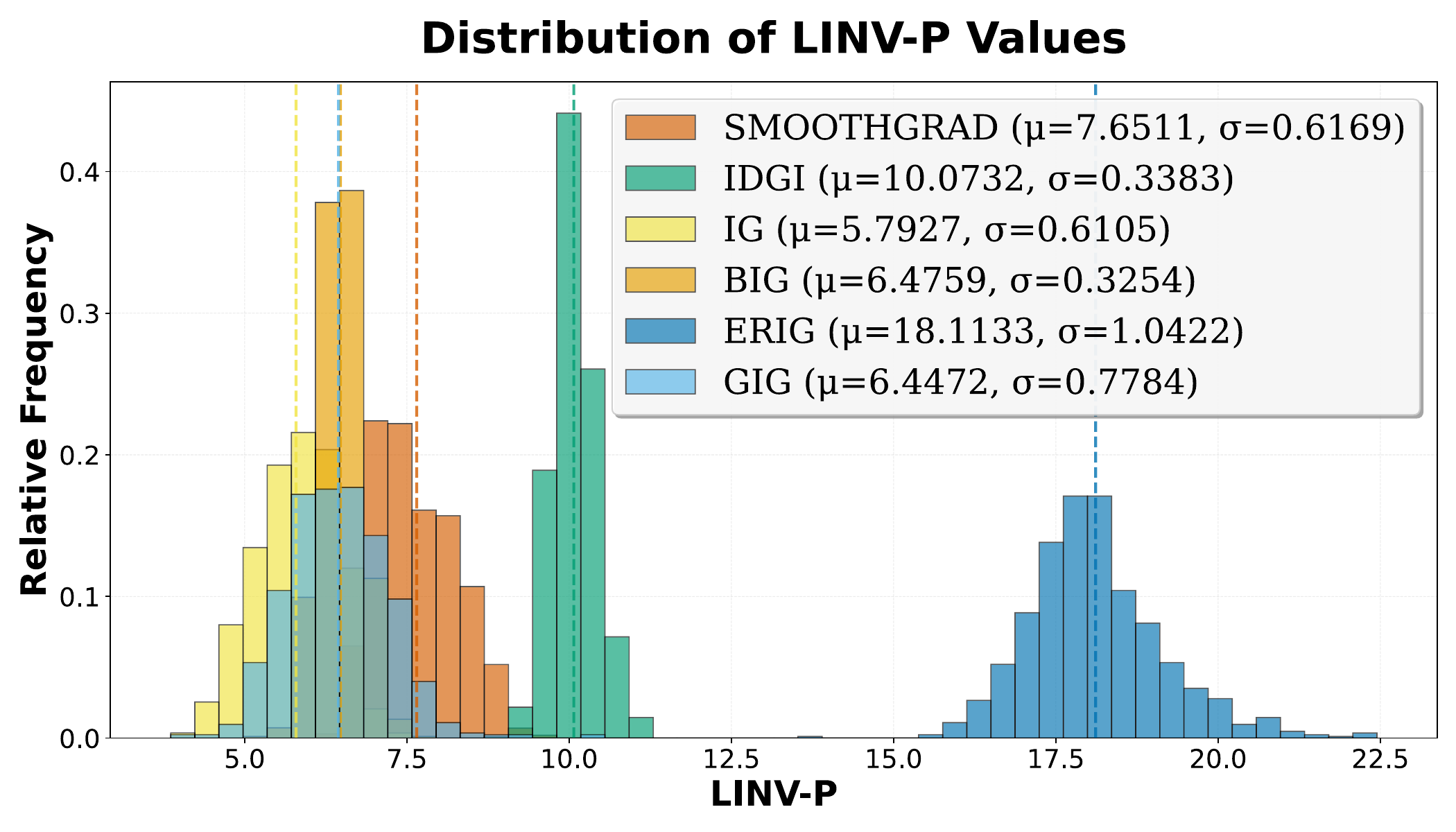}
    \caption{MobileNetV4 Noise Histograms}
    \label{fig: noise-compare}
    \vspace{-2mm}
\end{figure}

\begin{figure*}
    \centering
    \begin{subfigure}[b]{0.44\linewidth}
        \centering
        \includegraphics[width=\linewidth]{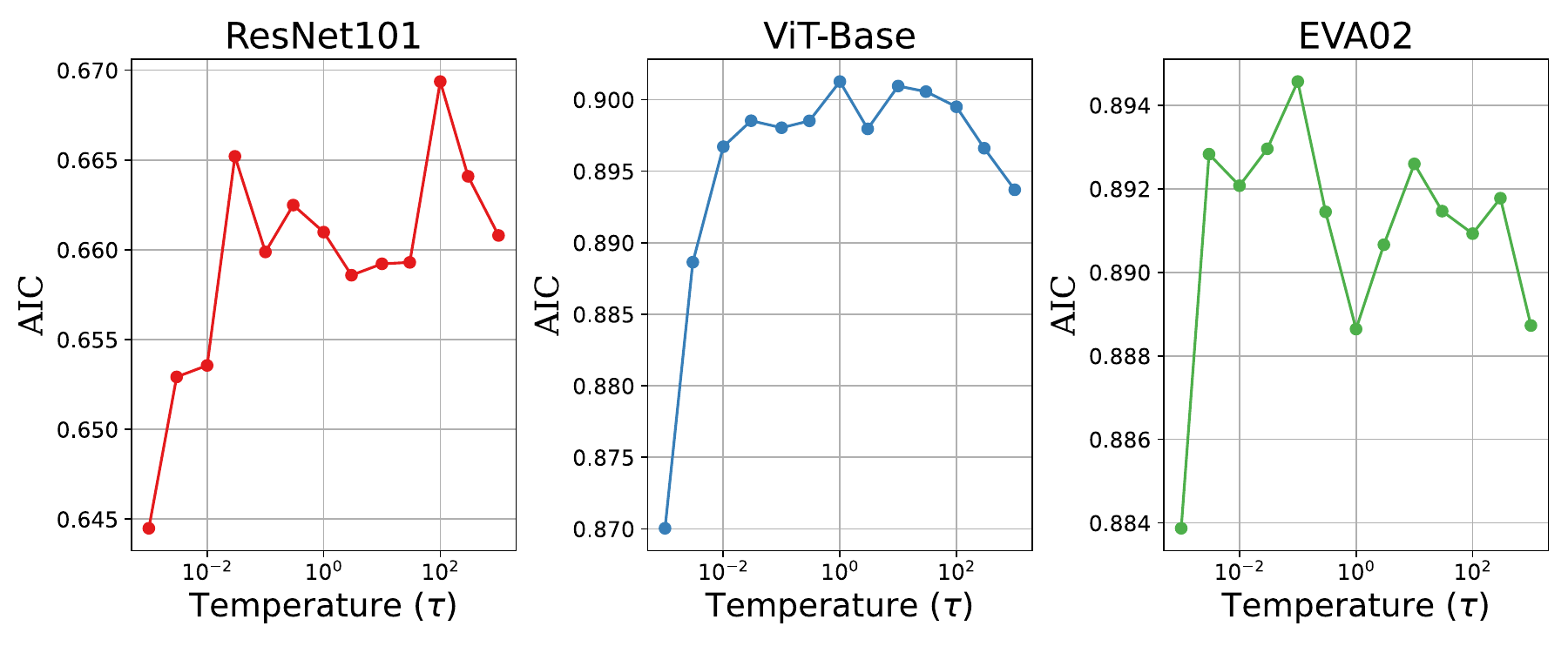}
        \caption{Fidelity (AIC) vs.\ $\tau$}
        \label{fig:sub1}
    \end{subfigure}
    \hspace{0.06\linewidth}
    \begin{subfigure}[b]{0.44\linewidth}
        \centering
        \includegraphics[width=\linewidth]{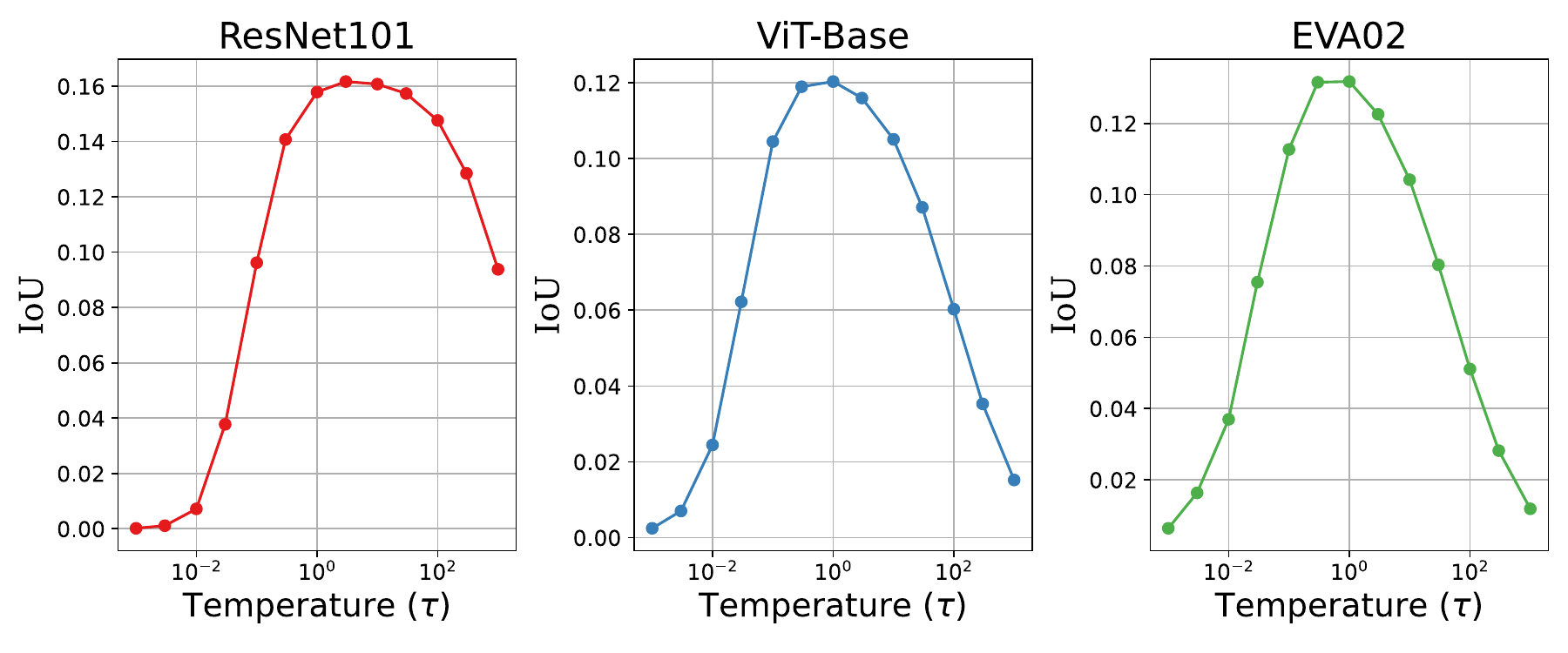}
        \caption{Localization (IoU) vs.\ $\tau$}
        \label{fig:sub2}
    \end{subfigure}
    \medskip
    \begin{subfigure}[b]{0.44\linewidth}
        \centering
        \includegraphics[width=\linewidth]{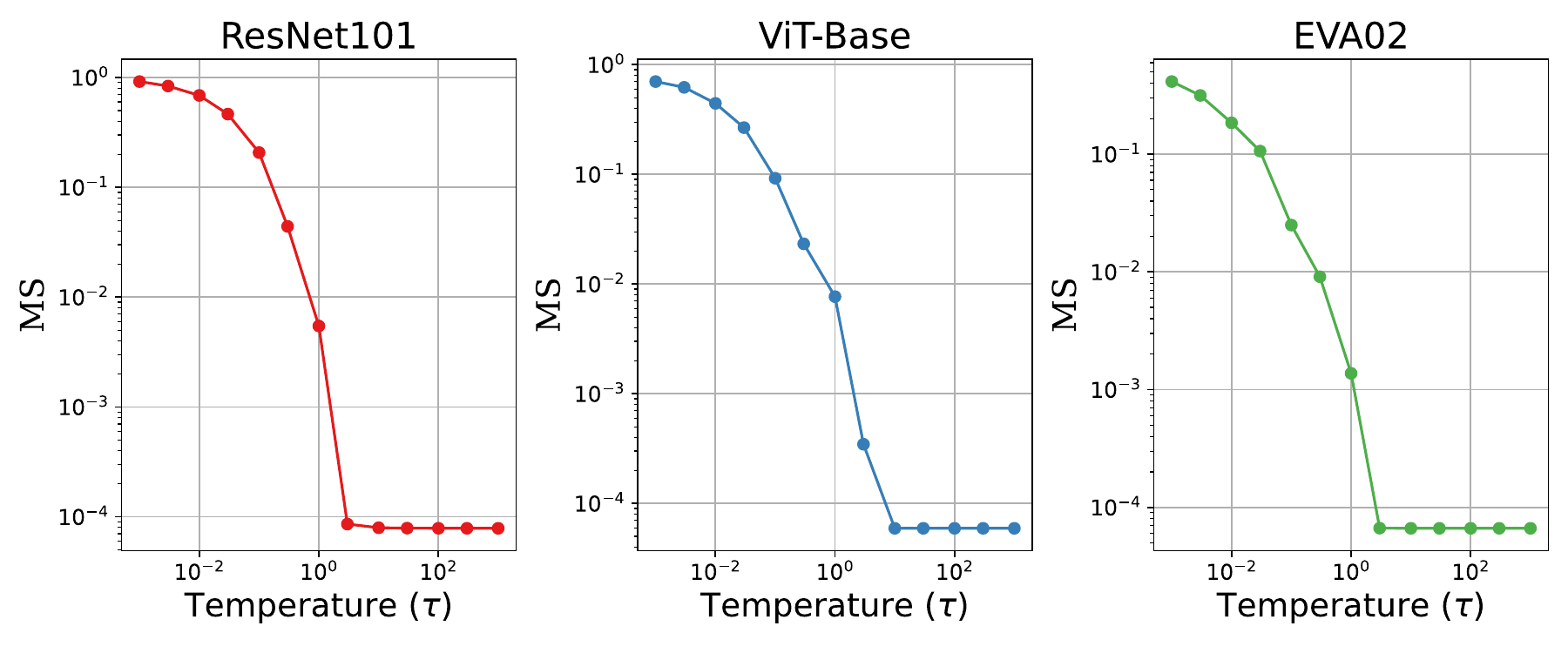}
        \caption{Stability (MS) vs.\ $\tau$}
        \label{fig:sub3}
    \end{subfigure}
    \hspace{0.06\linewidth}
    \begin{subfigure}[b]{0.44\linewidth}
        \centering
        \includegraphics[width=\linewidth]{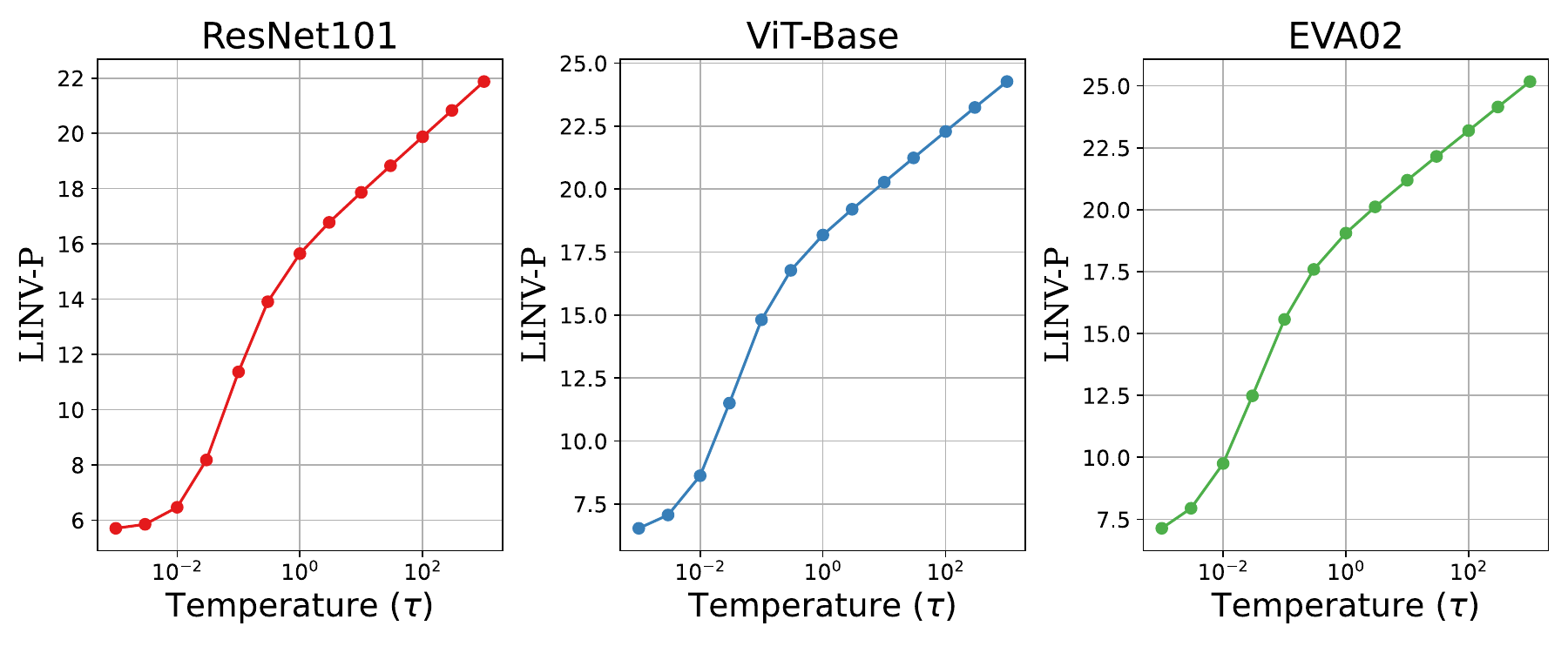}
        \caption{Noise (LINV-P) vs.\ $\tau$}
        \label{fig:sub4}
    \end{subfigure}
    \caption{Ablation of softmax temperature $\tau$ in ERIG across ResNet101, ViT-Base, and EVA02 models. All plots use a log-scale x-axis and a linear-scale y-axis, except stability, which also uses a log-scale y-axis.}
    \label{fig: temperature-ablation}
    \vspace{-5mm}
\end{figure*}
Under these metrics, ERIG achieves attribution noise levels several orders lower than competing methods (\cref{tab: noise-results}), and the MobileNetV4 distributions in \cref{fig: noise-compare} show a clear ordering. GIG~\cite{kapishnikov2021guided}, which steers the path away from high-gradient regions, outperforms vanilla IG but trails SmoothGrad's~\cite{smilkov2017smoothgrad} explicit perturbation averaging. IDGI~\cite{yang2023idgi} in turn surpasses SmoothGrad by weighting contributions with squared gradients. ERIG outperforms all three, so its entropy-regularized weighting attenuates noise more effectively than path-steering, external smoothing, and direct gradient weighting. The same ordering appears in the stability results (\cref{tab: stability-results}), supporting LINV as a meaningful proxy for stability. LINV thus adds a noise dimension absent from previous evaluation protocols; together with the fidelity, localization, and stability metrics, it enables a more complete evaluation of attribution quality.

\subsection{\texorpdfstring{Generalization Beyond ImageNet}{Generalization Beyond ImageNet}} \label{subsec: beyond-imagenet}
To verify that our findings are not specific to the ImageNet family, we
evaluate the full protocol on Oxford-IIIT Pet~\cite{parkhi2012cats}, a
fine-grained dataset with segmentation trimaps, using fine-tuned ResNet-50,
MobileNetV2, ConvNeXt-B, ViT-B, and DINOv2-S checkpoints. We also evaluate on the FunnyBirds
benchmark~\cite{hesse2023funnybirds} with ResNet-50, VGG16, and ViT-B
backbones. The ImageNet conclusions carry over to both settings. On Oxford Pet, ERIG
remains competitive with the strongest baseline on ranking-based fidelity
(mean gap of $-0.40\%$), leads localization by $20.2\%$ on average,
improves stability by ${\sim}38\times$ on average over the next-best method,
and achieves attribution noise levels several orders lower than competing
methods. On FunnyBirds, ERIG leads completeness by $3.4\%$ and contrastivity
by $2.1\%$ on average while staying within $1.5\%$ of the strongest baseline
on correctness. Full analysis in Appendix~\ref{appx: beyond-imagenet}.

\subsection{Ablation Studies} \label{sec: ablations}
\paragraph{Effect of Temperature on Fidelity and Localization.}
As shown in \cref{fig:sub1,fig:sub2}, fidelity and localization peak at an intermediate $\tau$. At low temperatures, the distribution is over-concentrated on a few dominant and possibly noisy gradient components, causing attributions to miss broader informative structure. At high temperatures, the distribution approaches uniformity and loses discriminability between informative and uninformative features. The optimal temperature $(\tau \approx 3)$ balances attribution sharpness against over-concentration, consistent with the balance characterized by $R(\tau)$ in \cref{cor: decomp}.
\vspace{-0.3cm}
\paragraph{Effect of Temperature on Stability and Noise Reduction.}
\Cref{fig:sub3} shows that for $\tau < 3$, attributions exhibit substantially higher
distortion under input perturbations, as the concentrated softmax amplifies spurious
gradient spikes that destabilize attribution maps. Beyond this region, distortion drops
sharply and plateaus as the entropy objective redistributes attribution mass over a broader,
more stable feature set. Concurrently, \Cref{fig:sub4} shows the inverse noise metric increasing
monotonically with $\tau$, where a more uniform softmax prevents any single high-gradient
component from dominating and dilutes the influence of noise spikes. Together, these observations mark $\tau \approx 3$ as a sweet spot, where stability and noise suppression are sufficient while fidelity and localization remain near their peak.
\vspace{-0.3cm}
\paragraph{Complementarity with SmoothGrad} Additionally, we validate our findings by ensembling all methods with SmoothGrad in Appendix~\ref{appx: smoothgrad-ensembles}. There, stability and noise metrics follow the same patterns, confirming that ERIG's intrinsic noise attenuation properties are complementary to external smoothing provided by ensemble methods.
\vspace{-0.3cm}
\paragraph{\texorpdfstring{Baseline Sensitivity.}{Baseline Sensitivity.}} Finally, we also ablate the baseline choice, evaluating ERIG across twelve alternatives on EVA-02 and ResNet101 in Appendix~\ref{appx: baselines}. Constant references track the zero default closely, other families less so; since all methods share one baseline, the relative claims hold.

%% file: sec/6_conclusion.tex
\section{Conclusion}
We present Entropy-Regularized Integrated Gradients (ERIG), a path-based attribution method that converts squared-gradient sensitivities into attribution weights using an entropy-regularized softmax. A temperature parameter $\tau$ balances sensitivity preservation against attenuation of local gradient irregularities. Our Lipschitz analysis decomposes attribution perturbations into a model-response term with a $\tau$-independent bound and a weight-variation term whose bound decays as $\mathcal{O}(1/(2\tau))$. Their ratio, $R(\tau)$, characterizes the trade-off between gradient information and attribution stability. We also adapt established noise estimators into LINV-C and LINV-P. Across eleven models on ImageNet and Oxford-IIIT Pet, ERIG achieves fidelity comparable to the strongest baselines while substantially improving localization, stability, and noise suppression, all at a per-step cost within ${\sim}5\%$ of IDGI and ${\sim}25\%$ of IG. On FunnyBirds, it places first or second in all metrics. Future work will explore further adaptive temperature schedules $\tau(t)$, analytic $\tau$ selection on 1-Lipschitz networks, and composition with gradient-flow corrections such as LibraGrad.

\section*{Acknowledgements}
The authors acknowledge NSF (IIS-2331908 and OAC-2530965), DOE (DE-SC0024576), DARPA (HR00112420004 and HR00112490420), NAIRR Pilot (NAIRR250261), Indiana Jetstream2 GPU (CIS240951 via ACCESS~\citep{hancock2021jetstream2, boerner2023access}), OpenAI, Google Cloud, and NCSA Delta GPU for partially contributing to this research. Any opinions, findings, and conclusions or recommendations expressed in this material are those of the author(s) and do not necessarily reflect the views of funding agencies and companies mentioned above.

%% file: sec/7_appendix.tex
\section{Appendix}

\addtocontents{toc}{\protect\setcounter{tocdepth}{3}}
\begingroup
\makeatletter
\renewcommand*\l@subsection{\@dottedtocline{2}{1.5em}{3.2em}}
\renewcommand*\l@subsubsection{\@dottedtocline{3}{4.7em}{4.4em}}

\def\numberline#1{\hb@xt@\@tempdima{\hfil#1\hskip 1em}}
\@starttoc{toc}
\endgroup

\clearpage

\subsection{Proof of Local Sensitivity} \label{appx: perturb-gradient-proof}

Let $\varepsilon$ be a very small ($\varepsilon \ll 1$) random unbiased $\bigl(\mathbf{E}[\varepsilon] = 0\bigr)$ perturbation of the
$i$-th coordinate of input $\bm{x}$. The corresponding perturbed vector is
$\bm{x}_\varepsilon = \{x_1, x_2,\ldots,x_i + \varepsilon, \ldots, x_N\}$. A Taylor expansion of
$f$ around $\bm{x}$ in the $i$-th coordinate gives
\begin{align}
    f(\bm{x}_\varepsilon) &= f(\bm{x}) + \frac{\partial f}{\partial x_i} \cdot \varepsilon
        + \frac{1}{2} \frac{\partial^2 f}{\partial x_i^2} \cdot \varepsilon^2 + \cdots\\
    &= f(\bm{x}) + \frac{\partial f}{\partial x_i} \cdot \varepsilon + \mathcal{O}(\varepsilon^2)
\end{align}
Taking expectation of $f(\bm{x}_\varepsilon)$ gives us,
\begin{align}
    \mathbf{E}[f(\bm{x}_\varepsilon)]
        &= \mathbf{E}\left[f(\bm{x}) + \frac{\partial f}{\partial x_i} \cdot \varepsilon
            + \mathcal{O}(\varepsilon^2)\right]\\
        &= f(\bm{x}) + \frac{\partial f}{\partial x_i} \cdot \mathbf{E}[\varepsilon]
            + \mathcal{O}(\varepsilon^2)\\
        &= f(\bm{x}) + \mathcal{O}(\varepsilon^2)
\end{align}
Hence, we can compute the variance as,
\begin{align}
    \mathrm{Var}\left[f(\bm{x}_\varepsilon)\right]
        = \mathbf{E}\left[\Bigl(f(\bm{x}_\varepsilon) - \mathbf{E}[f(\bm{x}_\varepsilon)]\Bigr)^{2}\right]
        &= \mathbf{E}\left[\left(\frac{\partial f}{\partial x_i} \cdot \varepsilon
            + \mathcal{O}(\varepsilon^2)\right)^2\right]\\
        &= \left(\frac{\partial f}{\partial x_i}\right)^{2} \cdot \mathbf{E}[\varepsilon^2]
            + \mathcal{O}(\varepsilon^3)\\
        &\approxprop \left(\frac{\partial f}{\partial x_i}\right)^{2}
\end{align}
Here, $\mathbf{E}[\varepsilon^2]$ acts as the proportionality constant in the approximation, while the squared partial derivative, $g_i^2 = \bigl(\partial f / \partial x_i\bigr)^{2}$, measures the local sensitivity of the function $f$ to feature $x_i$. This quantity, integrated over the entire input space, forms the basis of \emph{Derivative-based Global Sensitivity Measures} (DGSMs) \citep{sobol2009derivative, lamboni2013derivative}, a widely used alternative to Sobol' sensitivity indices \citep{sobol2001global}. DGSMs have been applied across a range of domains, including input screening in environmental models \citep{andrea2008global}, sensitivity analysis of ODEs in biology \citep{zi2011sensitivity}, and uncertainty quantification in geophysical simulations \citep{cacuci2003sensitivity}, etc.

\subsection{Maximum Entropy Weight Derivation} \label{appx: maxent-proof}

\subsubsection{Solution via KKT Conditions}

We address the constrained optimization problem via the Karush-Kuhn-Tucker (KKT) conditions. First, we restate the problem in the following standard form:
\begin{gather}
    \min_{\bm{w} \in [0,1]^N} \left(\sum_{i = 1}^N w_i \ln{w_i}\right), \qquad \text{s.t.} \\
    \left(1 - \sum_{i = 1}^N w_i\right) = 0, \quad \left(\gamma - \sum_{i = 1}^N w_i\, g_i^2\right) \leq 0
\end{gather}
Next, we construct the Lagrangian:
\begin{equation}
    \mathcal{L}(\bm{w}, \alpha, \lambda) = \sum_{i = 1}^N w_i \ln{w_i} + \alpha \left(1 - \sum_{i=1}^N w_i\right) + \lambda \left(\gamma - \sum_{i = 1}^N w_i\, g_i^2\right),
    \label{eq:lagrangian}
\end{equation}
where $\alpha \in \mathbb{R}$ is the Lagrange multiplier for the equality constraint (normalization), and $\lambda \geq 0$ is the Lagrange multiplier for the inequality constraint (gradient magnitude threshold). The KKT conditions for optimality are:
\begin{align}
    \text{Stationarity:} \quad & \frac{\partial \mathcal{L}}{\partial w_i} = 1 + \ln{w_i} - \alpha - \lambda g_i^2 = 0, \forall i \in [N], \label{eq:stationarity}\\
    \text{Primal feasibility:} \quad & \sum_{i=1}^N w_i = 1, \quad \sum_{i = 1}^N w_i g_i^2 \geq \gamma, \label{eq:primal}\\
    \text{Dual feasibility:} \quad & \lambda \geq 0, \label{eq:dual}\\
    \text{Complementary slackness:} \quad & \lambda \left(\gamma - \sum_{i = 1}^N w_i g_i^2\right) = 0. \label{eq:slackness}
\end{align}
From the stationarity condition~\eqref{eq:stationarity}, we can solve for $w_i$:
\begin{align}
    \ln{w_i} &= (\alpha - 1) + \lambda g_i^2, \\
    w_i &= \exp(\alpha - 1) \exp(\lambda g_i^2). \label{eq:weight-unnormalized}
\end{align}
Using the normalization constraint $\sum_{i=1}^N w_i = 1$ from the primal feasibility conditions~\eqref{eq:primal}, we eliminate the constant $\exp(\alpha - 1)$:
\begin{align}
    \sum_{i=1}^N w_i &= \exp(\alpha - 1) \sum_{i=1}^N \exp(\lambda g_i^2) = 1, \\
    \exp(\alpha - 1) &= \frac{1}{\sum_{j=1}^N \exp(\lambda g_j^2)}. \label{eq:normalization-constant}
\end{align}
Substituting equation~\eqref{eq:normalization-constant} into equation~\eqref{eq:weight-unnormalized}, we obtain the normalized weights:
\begin{equation}
    w^{*}_i = \frac{\exp(\lambda g_i^2)}{\sum_{j = 1}^N \exp(\lambda g_j^2)} = \Softmax_i(\lambda \bm{g}^2),
    \label{eq:weight-solution}
\end{equation}
where $\Softmax_i(\cdot)$ denotes the $i$-th component of the softmax function.

The Lagrange multiplier $\lambda \geq 0$ controls the trade-off between entropy maximization and gradient emphasis. When $\lambda = 0$, we recover the uniform distribution (maximum entropy with no gradient bias). As $\lambda$ increases, the distribution concentrates more on features with larger gradient magnitudes while assigning lower probability to features with smaller gradients.

\subsubsection{Relationship Between \texorpdfstring{$\lambda$}{\textlambda} and \texorpdfstring{$\gamma$}{\textgamma}}

From the complementary slackness condition~\eqref{eq:slackness}, at the optimal solution either $\lambda = 0$ (in which case the gradient strength constraint will not be active resulting in a uniform distribution) or the constraint holds with equality:
\begin{equation}
    \sum_{i = 1}^N w^{*}_i g_i^2 = \gamma.
    \label{eq:constraint-active}
\end{equation}
Substituting \cref{eq:weight-solution} into \cref{eq:constraint-active}, we obtain the relationship between $\lambda$ and $\gamma$:
\begin{equation}
    \sum_{i = 1}^N \frac{g_i^2 \exp(\lambda g_i^2)}{\sum_{j = 1}^N \exp(\lambda g_j^2)} = \gamma.
    \label{eq:lambda-gamma-relation}
\end{equation}
This equation implicitly defines $\lambda$ as a function of $\gamma$. In principle, for a given threshold $\gamma$, one could solve \cref{eq:lambda-gamma-relation} numerically to find the corresponding $\lambda$.

\subsubsection{Practical Implementation}

While tuning $\gamma$ would require solving the constrained optimization problem for each value, it is more practical to tune $\lambda$ directly as a hyperparameter. We denote this hyperparameter as $\tau = 1/\lambda$ (the temperature parameter in the softmax), which provides direct control over the concentration of the distribution on high-gradient regions without requiring iterative optimization. Hence, the ERIG weighting scheme becomes:
\begin{equation}
    w_i^{*}(\bm{g}(\bm{x}); \tau) = \Softmax_i\left(\frac{\nabla f \odot \nabla f}{\tau}\right) = \frac{\exp(g_i^2/\tau)}{\sum_{j = 1}^N \exp(g_j^2/\tau)}.
    \label{eq:erig-weight}
\end{equation}
\qed

\subsection{More Visualization Examples}  \label{appx: viz-examples}
The attribution values are mapped to a \([0,1]\) range following the methodology of Integrated Gradients~\cite{sundararajan2017axiomatic}. Below, you can find more comparisons of attribution maps for the MobileNetV4 model:
\begin{figure}[H]
    \centering
    \begin{subfigure}[c]{0.46\linewidth}
        \centering
        \includegraphics[width=\linewidth]{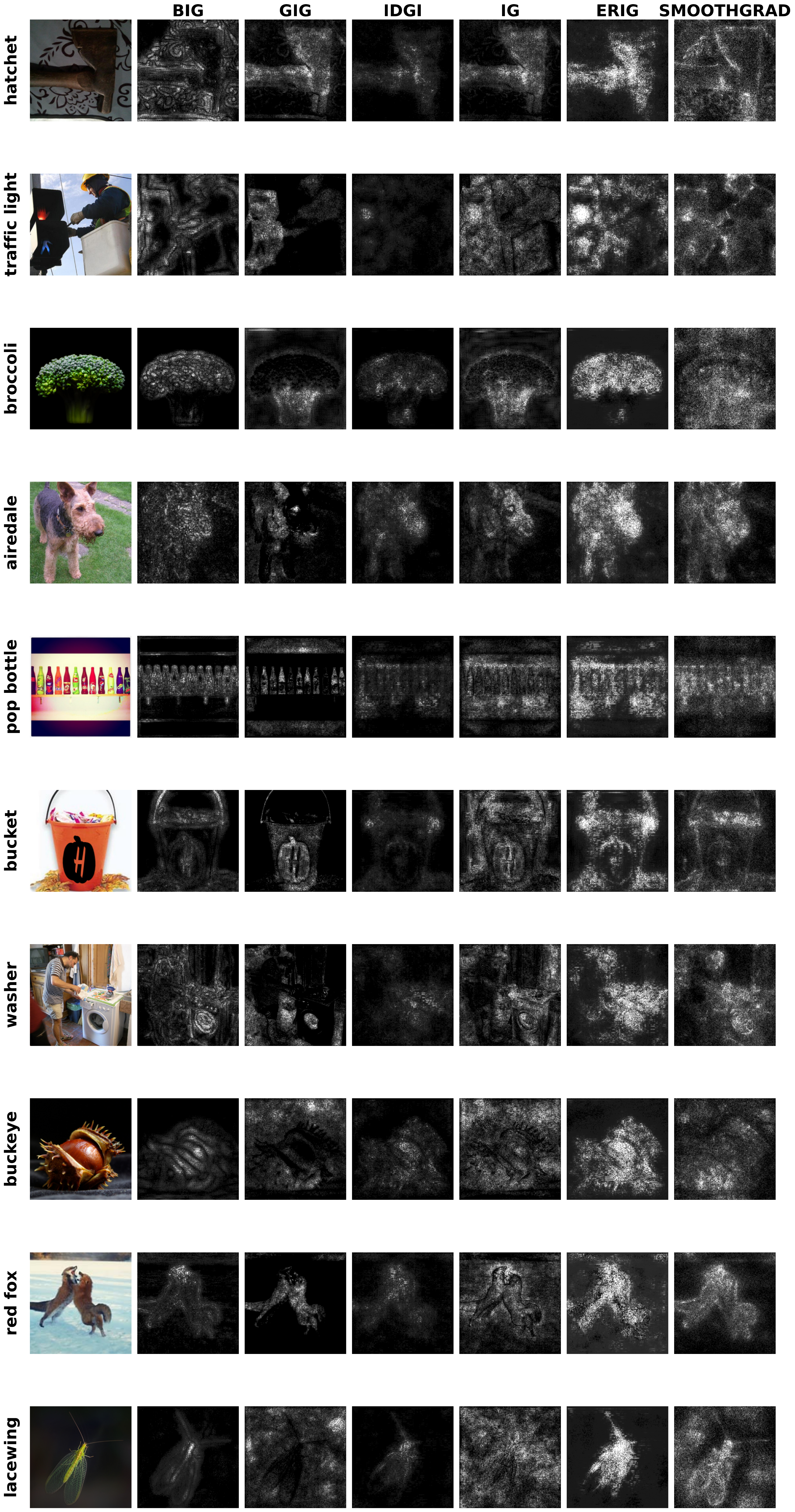}
        \label{fig: more-viz-1}
    \end{subfigure}
    \hfill
    \begin{subfigure}[c]{0.46\linewidth}
        \centering
        \includegraphics[width=\linewidth]{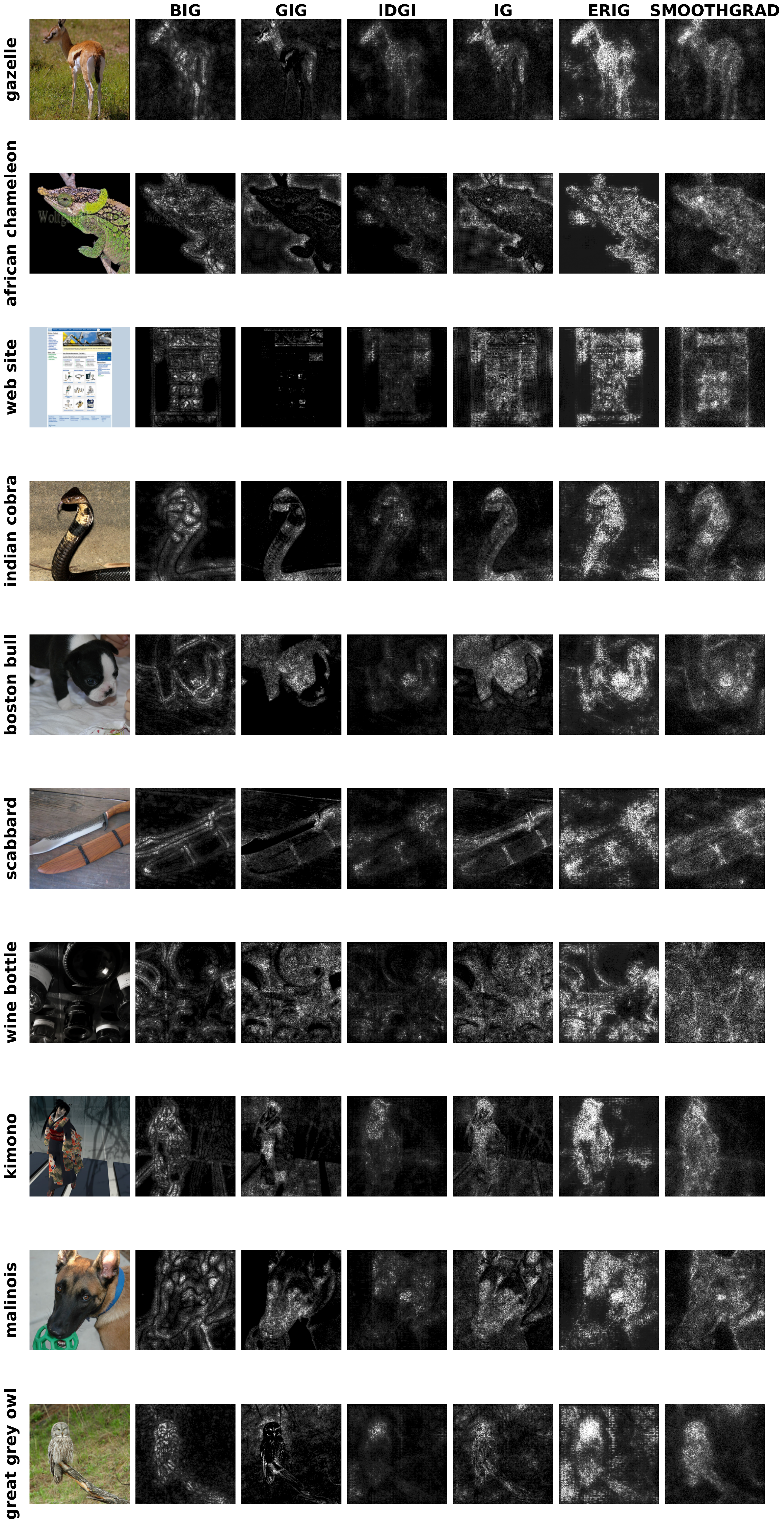}
        \label{fig: more-viz-2}
    \end{subfigure}
    \caption{Additional visualizations of attribution maps for the MobileNetV4 architecture.}
    \label{fig: combined-viz}
\end{figure}

\subsection{Fidelity Distributions} \label{appx: fidelity-distr}
To complement the fidelity results in \cref{tab: auc-results}, we plot the per-sample distributions of each fidelity metric across all six architectures. Each shaded region spans the inter-quartile range (25th to 75th percentile); the horizontal line inside the shaded region represents the mean, and the vertical capped lines indicate the full range across samples. As discussed in \cref{subsec: fidelity}, the substantial overlap in inter-quartile ranges across methods is consistent with the rank-only nature of these metrics, since they depend solely on attribution ranking and methods with markedly different attribution magnitudes and noise characteristics can produce similar fidelity scores.

\begin{figure}[H]
    \centering
    \includegraphics[width=\linewidth]{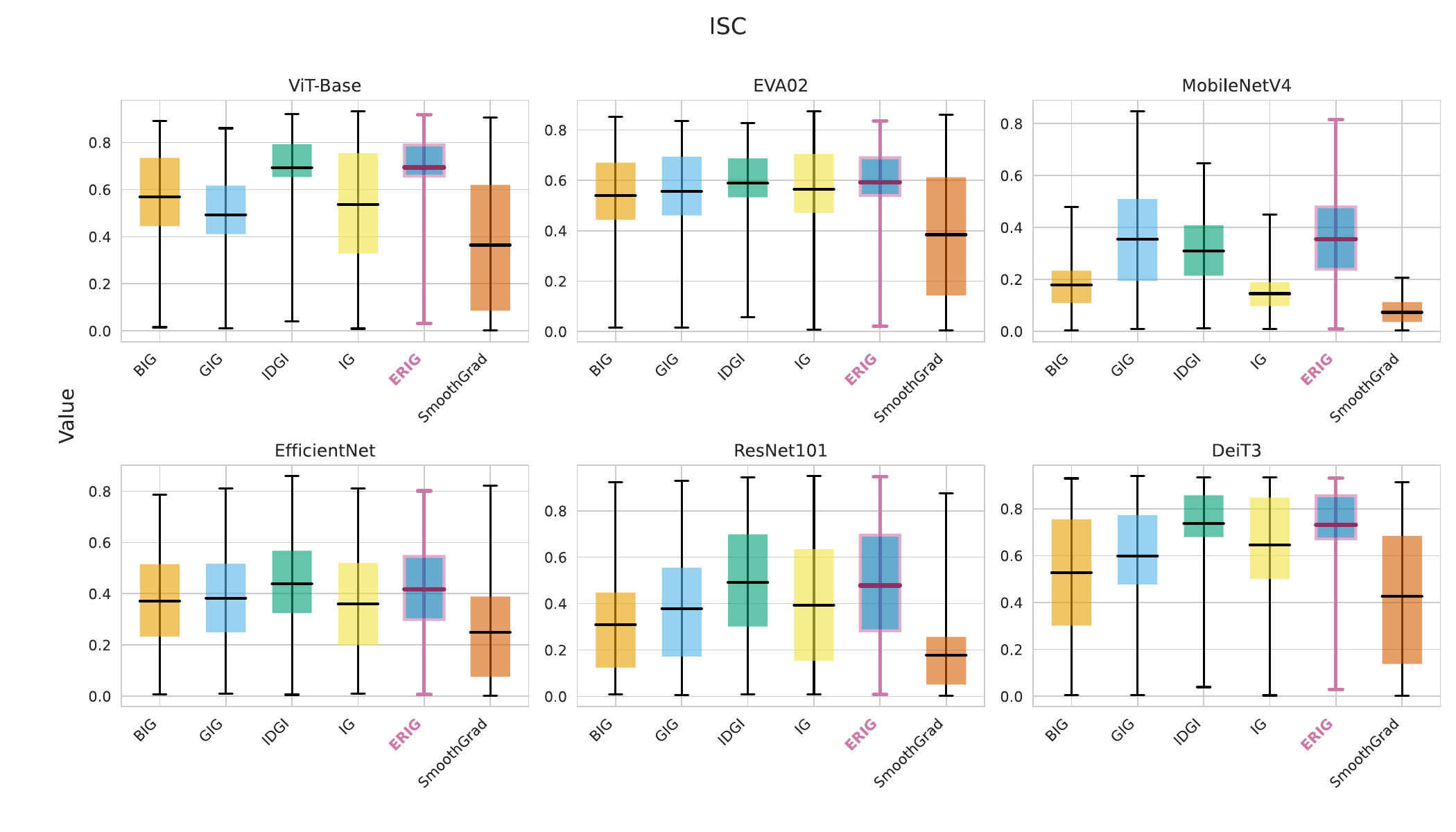}
    \caption{ISC distribution with inter-quartile range and full range across all architectures. Higher is better.}
    \label{fig: distr-isc}
\end{figure}

\begin{figure}[H]
    \centering
    \includegraphics[width=\linewidth]{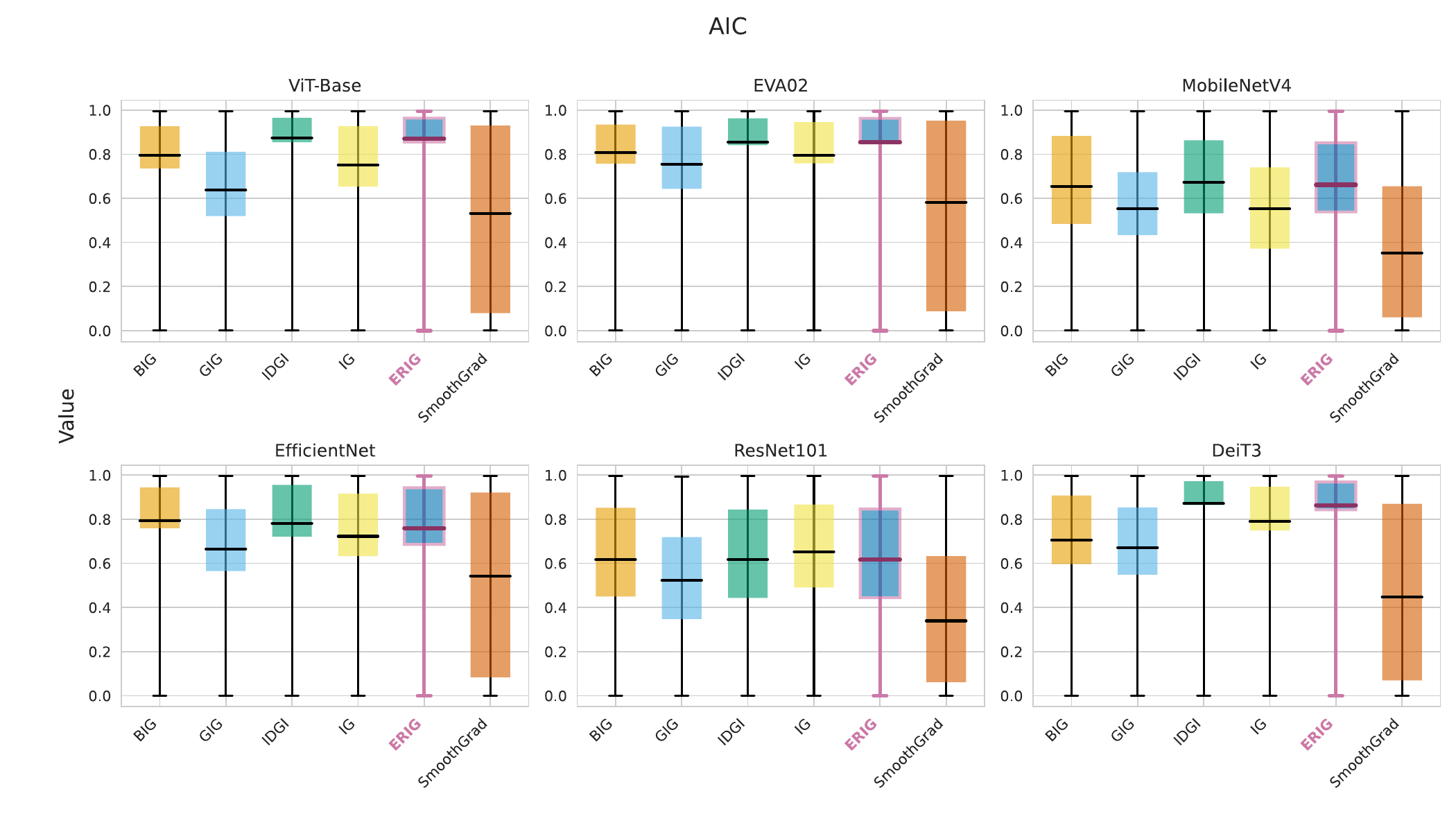}
    \caption{AIC distribution with inter-quartile range and full range across all architectures. Higher is better.}
    \label{fig: distr-aic}
\end{figure}

\begin{figure}[H]
    \centering
    \includegraphics[width=0.85\linewidth]{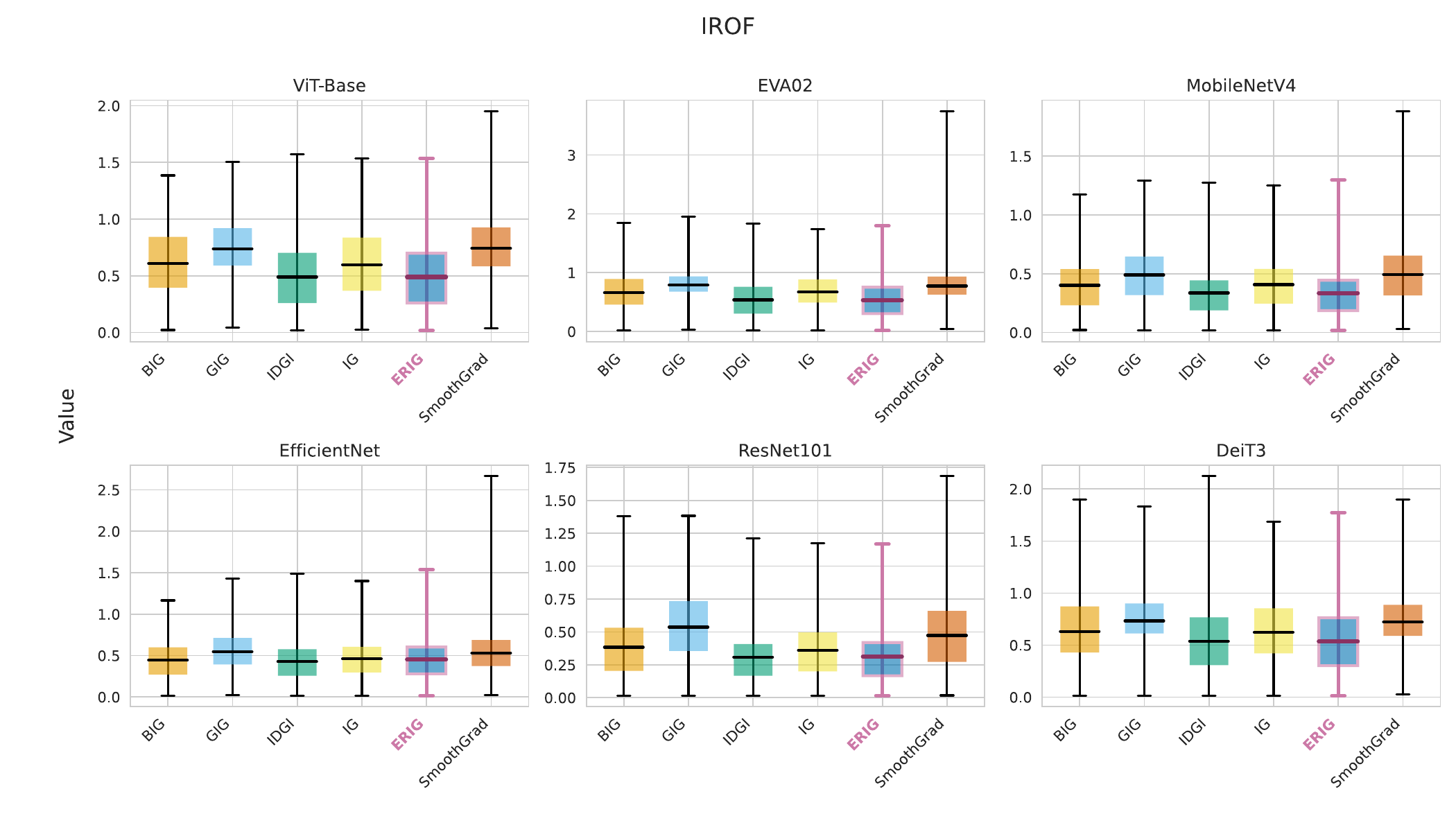}
    \caption{DEL distribution with inter-quartile range and full range across all architectures. Lower is better.}
    \label{fig: distr-del}
\end{figure}

\subsection{Localization Distributions} \label{appx: localization-distr}
Here we plot the per-sample distributions of each localization metric in \cref{tab: localization-results} across all six architectures using the same format as \cref{appx: fidelity-distr}. In contrast to the fidelity distributions, ERIG's inter-quartile ranges are visibly higher than those of all baselines for IoU and Pixel F1, confirming the localization advantages reported in the main text. Pixel Accuracy shows less separation because, unlike IoU and Pixel F1, it does not penalize false positives relative to the object region and is therefore less sensitive to differences in localization quality. This limitation is well known and has been studied by Csurka et al.~\cite{csurka2013good} in 2013 and, more recently, by Wang et al.~\cite{wang2023revisiting} in 2023.
\begin{figure}[H]
    \centering
    \includegraphics[width=0.86\linewidth]{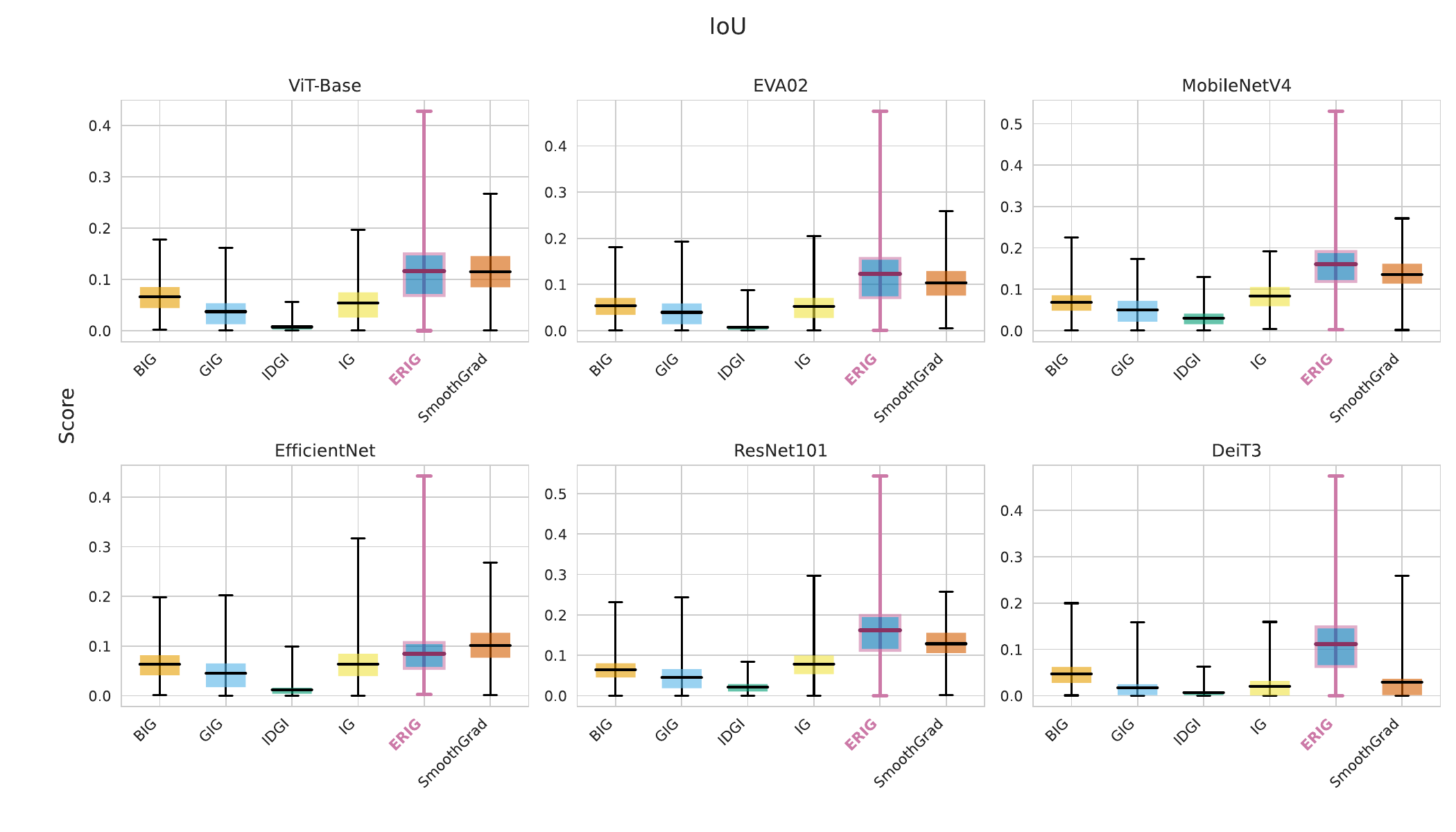}
    \caption{IoU distribution with inter-quartile range and full range across all architectures. Higher is better.}
    \label{fig: distr-iou}
\end{figure}

\begin{figure}[H]
    \centering
    \includegraphics[width=0.86\linewidth]{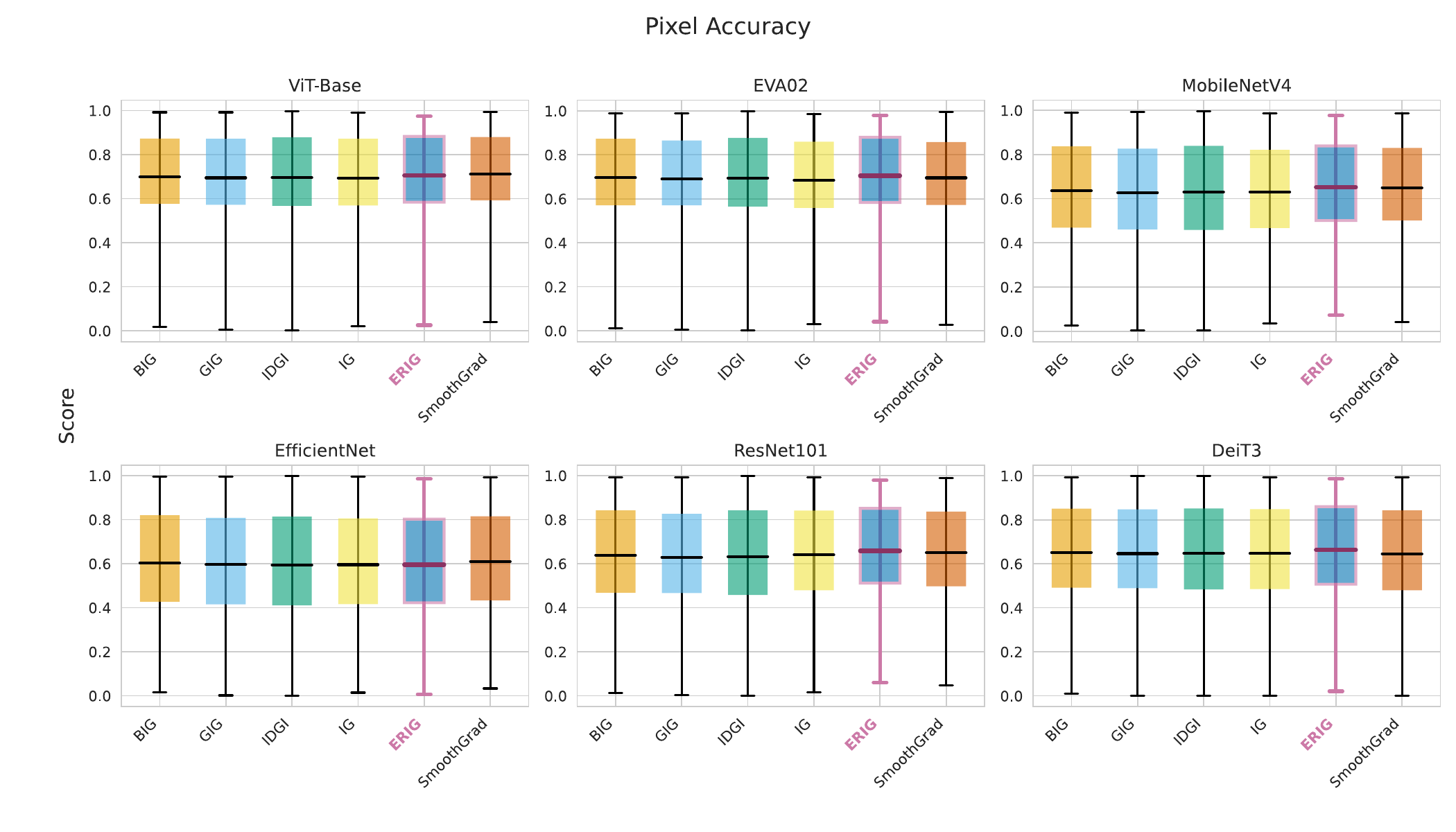}
    \caption{Pixel Accuracy distribution with inter-quartile range and full range across all architectures. Higher is better.}
    \label{fig: distr-pa}
\end{figure}

\begin{figure}[H]
    \centering
    \includegraphics[width=0.86\linewidth]{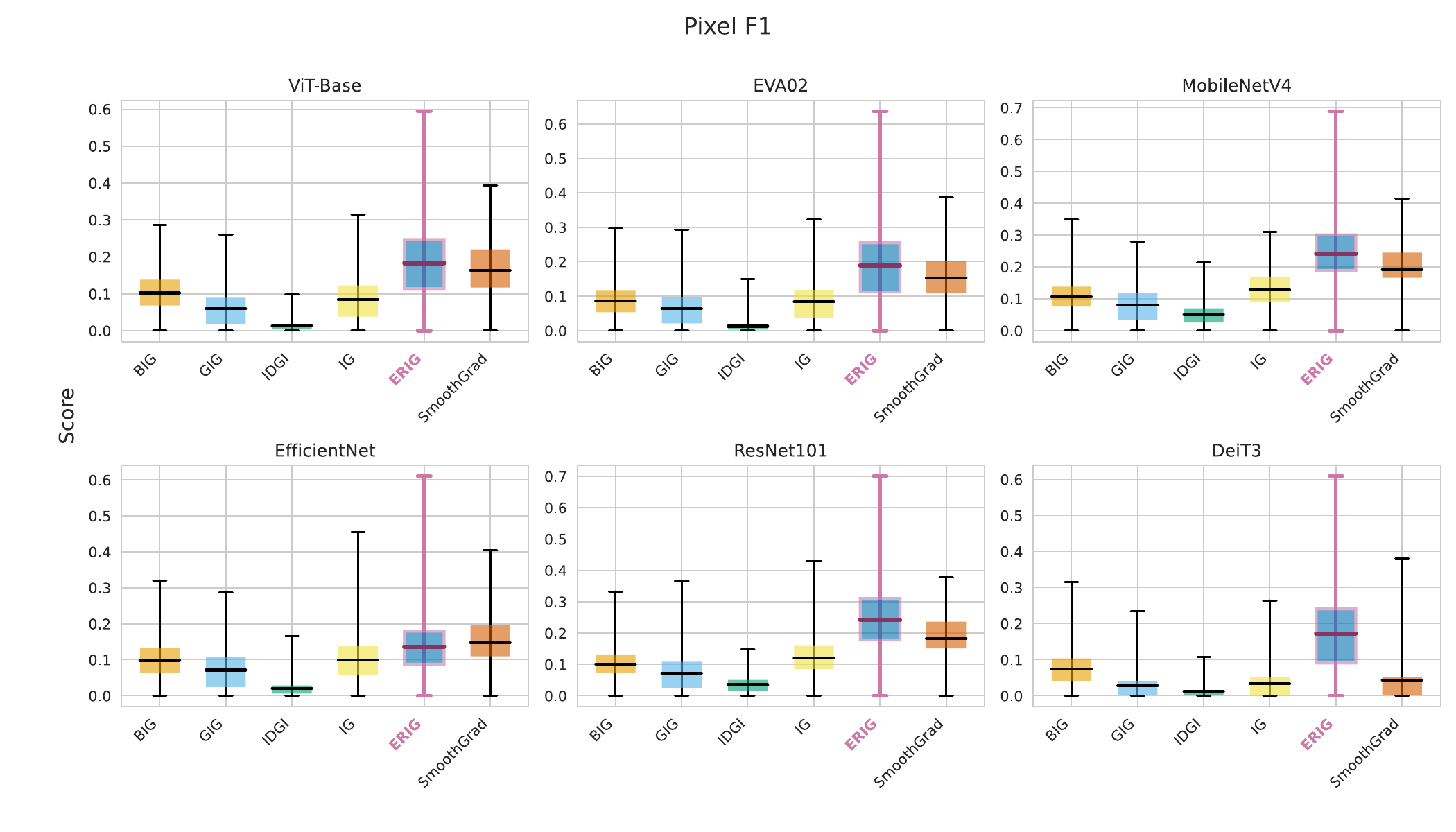}
    \caption{Pixel F1 distribution with inter-quartile range and full range across all architectures. Higher is better.}
    \label{fig: distr-pf1}
\end{figure}

\subsection{Stability Distributions} \label{appx: stability-distr}
To complement the tabulated stability results in \cref{tab: stability-results}, we plot each stability metric across all six architectures on a $\log_{10}$ scale for ease of visualization. Each shaded region spans the inter-quartile range (25th to 75th percentile); the horizontal line inside the shaded region represents the mean, and the vertical capped lines indicate the full range across samples. The ERIG bar is visually highlighted for clarity. Across all metrics and models, ERIG's inter-quartile ranges are entirely non-overlapping with those of every baseline, confirming that the improvements reported in the main text are meaningful.

\begin{figure}[H]
    \centering
    \includegraphics[width=\linewidth]{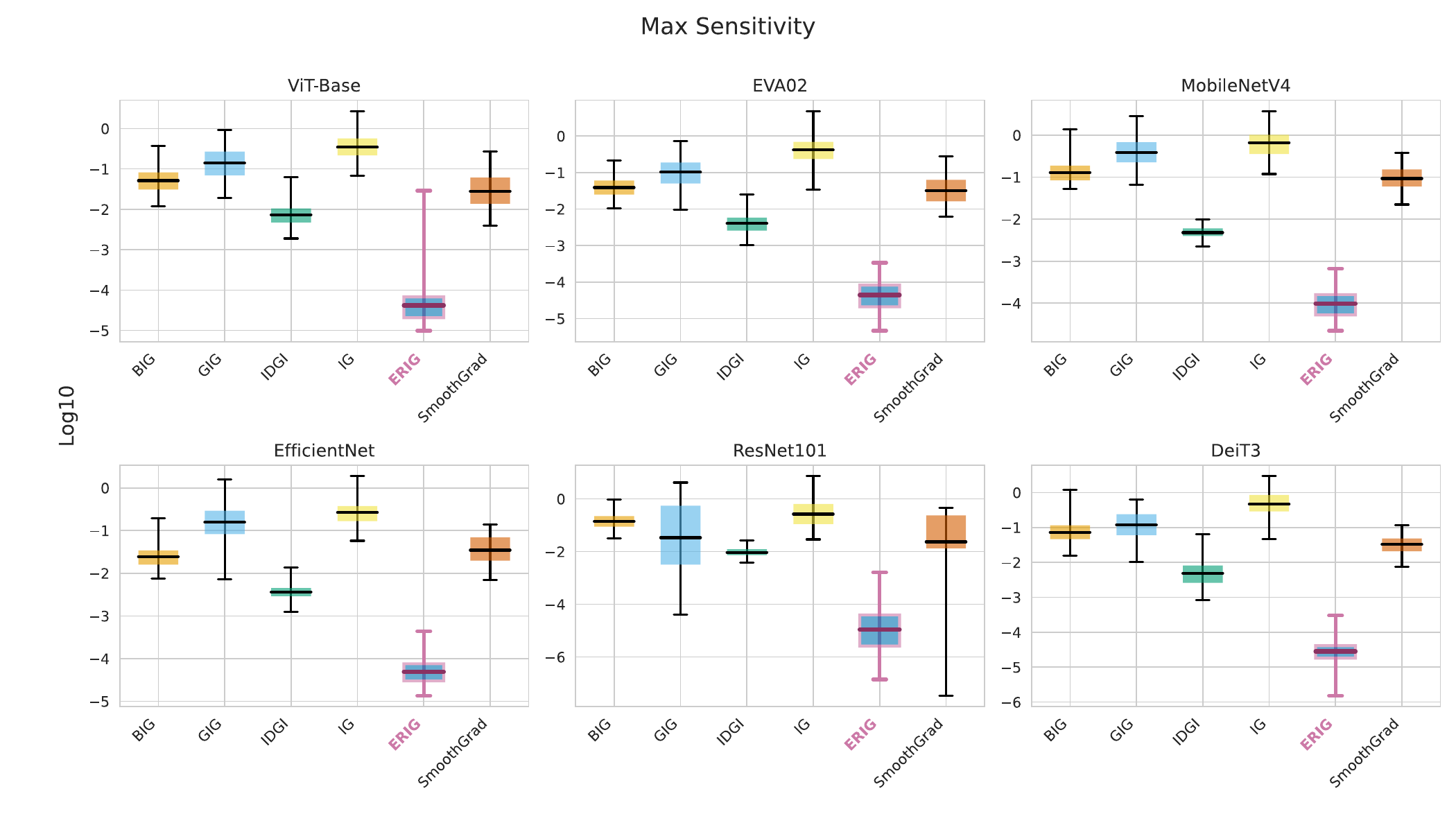}
    \caption{Max Sensitivity (MS) with inter-quartile range and full range across all architectures ($\log_{10}$ scale). Lower is better.}
    \label{fig: ci-ms}
\end{figure}

\begin{figure}[H]
    \centering
    \includegraphics[width=\linewidth]{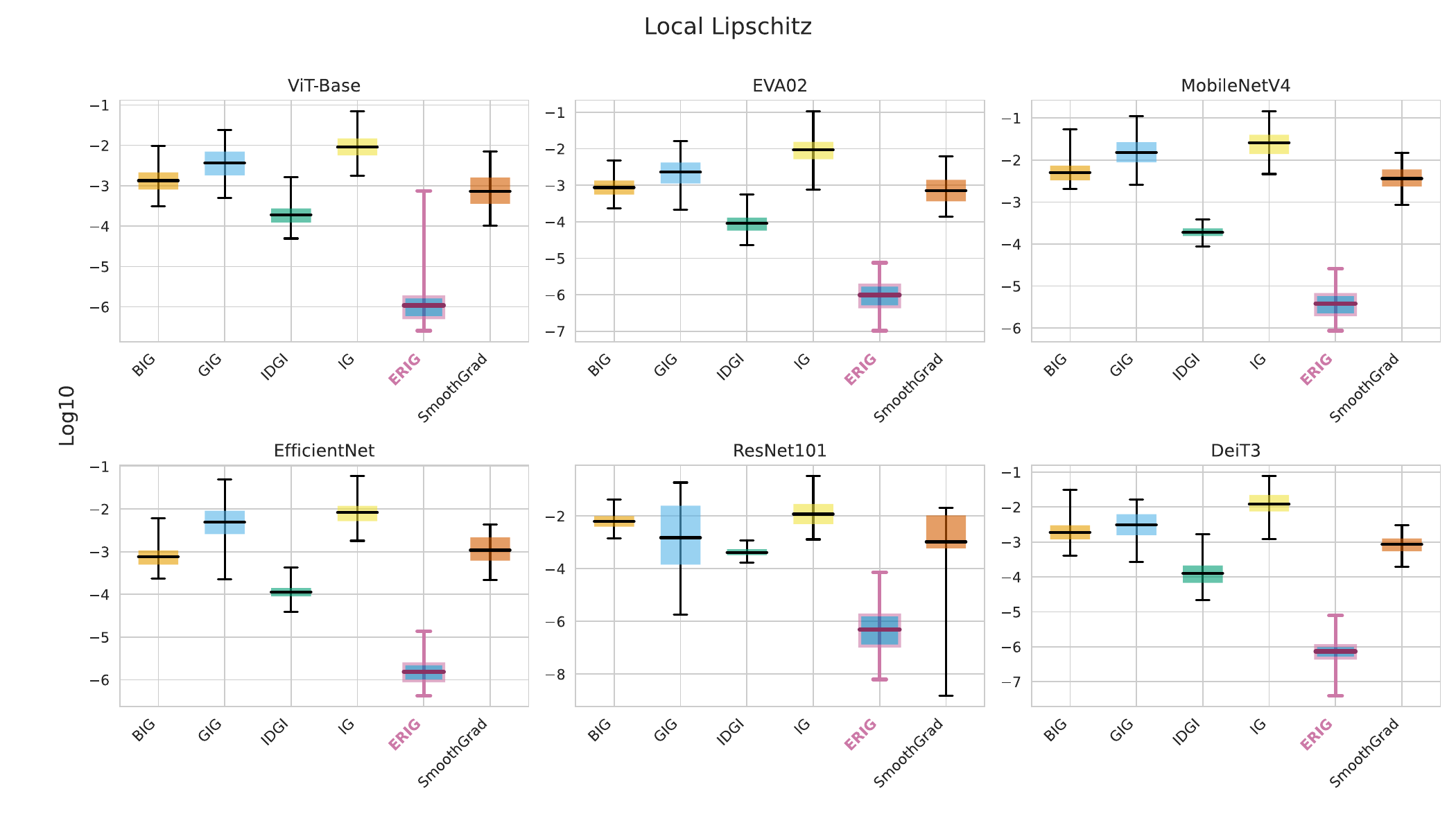}
    \caption{Local Lipschitz Constant (LLC) with inter-quartile range and full range across all architectures ($\log_{10}$ scale). Lower is better.}
    \label{fig: ci-llc}
\end{figure}

\begin{figure}[H]
    \centering
    \includegraphics[width=\linewidth]{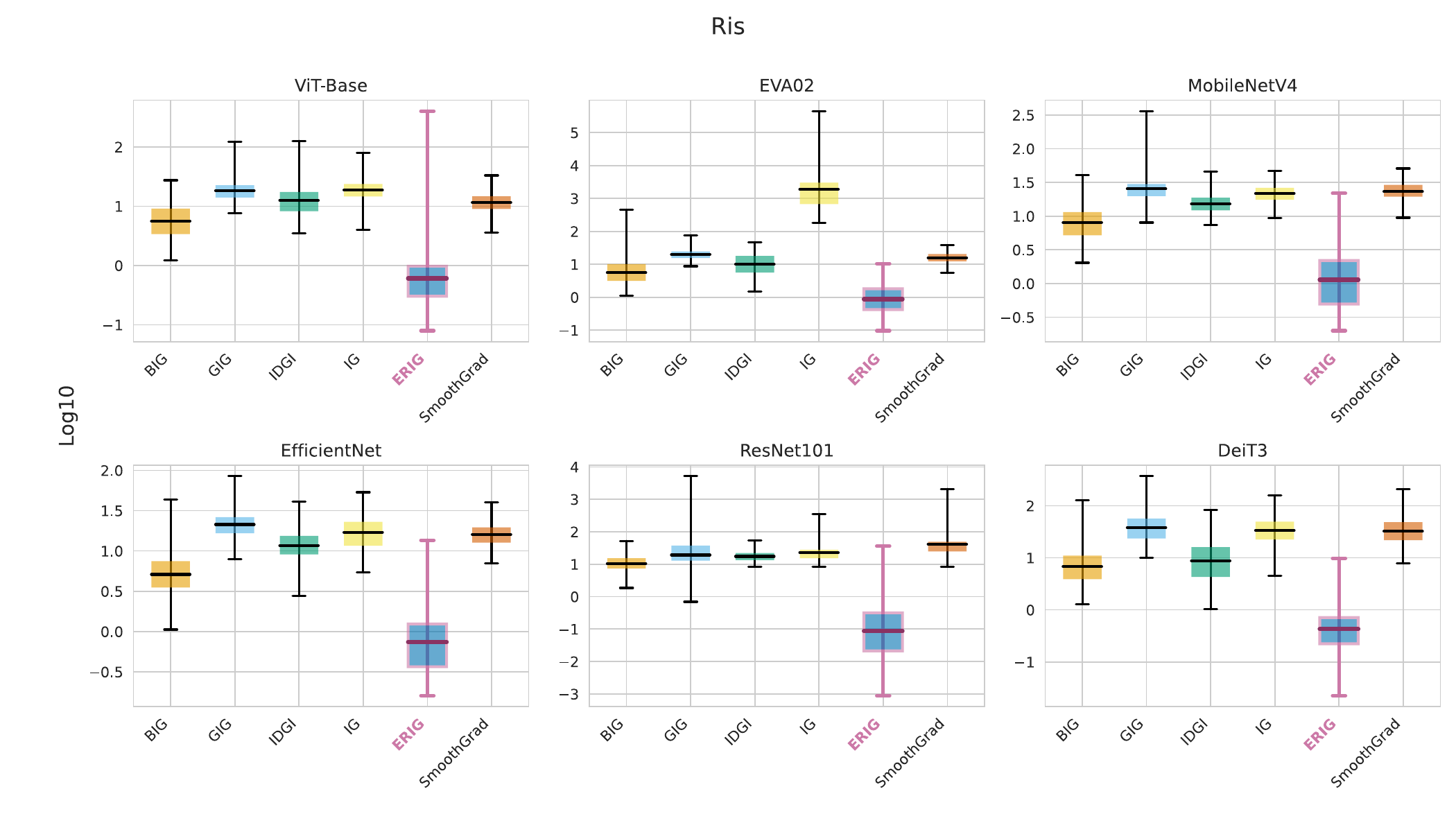}
    \caption{Relative Input Stability (RIS) with inter-quartile range and full range across all architectures ($\log_{10}$ scale). Lower is better.}
    \label{fig: ci-ris}
\end{figure}

\subsection{Proof of Lipschitz Decomposition} \label{appx: lipschitz-proof}
Let $\Gamma(t) = \bm{x}^b + t\bm{d}$ with $\bm{d} = \bm{x}^a - \bm{x}^b$ and $\tilde{\Gamma}(t) = \Gamma(t) + t\bm{\eta}$.  We bound $\|\bm{M}\|_2$ and $\|\bm{W}\|_2$ separately.

\subsubsection{Model-Response Term}
Since $w_i^{*} \in [0,1]$ and $\sum_i w_i^{*} = 1$,
\begin{equation}
  \|\bm{M}\|_2
  = \biggl\|\int_0^1 \bm{w}^{*}(\bm{g}(t);\tau)\;\Delta\dot{f}(t)\,dt\biggr\|_2
  \le \int_0^1 |\Delta\dot{f}(t)|\, dt.
\end{equation}
Define $\phi(t) = f(\tilde{\Gamma}(t)) - f(\Gamma(t))$, so that $\Delta\dot{f}(t) = \phi'(t)$. The derivative of $\phi$ is the difference of the directional derivatives of $f$ along $\tilde{\Gamma}(t)$ and $\Gamma(t)$:
\begin{align}
  \phi'(t) &= \nabla f(\Gamma(t) + t\bm{\eta}) \cdot (\bm{d} + \bm{\eta}) - \nabla f(\Gamma(t)) \cdot \bm{d}\\
  &= \bigl[\nabla f(\Gamma(t) + t\bm{\eta}) - \nabla f(\Gamma(t))\bigr] \cdot \bm{d} \;+\; \nabla f(\Gamma(t) + t\bm{\eta}) \cdot \bm{\eta}.
\end{align}
Using  $\|\Gamma(t) + t\bm{\eta} - \Gamma(t)\|_2 = t\|\bm{\eta}\|_2$, $\|\bm{d}\|_2 = |\Gamma|$ with Assumptions~\ref{as:lip-f} and ~\ref{as:lip-g}, one can write,
\begin{equation}
  |\phi'(t)| \le L_g\, t\, \|\bm{\eta}\|_2\, |\Gamma| + L_f\, \|\bm{\eta}\|_2.
\end{equation}
Integrating over $[0,1]$ yields the model-response bound,
\begin{equation}\label{eq:model-response-bound}
  \|\bm{M}\|_2 \le \int_0^1 \bigl(L_g\, t\, \|\bm{\eta}\|_2\, |\Gamma| + L_f\, \|\bm{\eta}\|_2\bigr)\, dt = \left(L_f + \tfrac{1}{2}\, L_g\, |\Gamma|\right) \|\bm{\eta}\|_2.
\end{equation}

\subsubsection{Weight-Variation Term}
We first bound the sensitivity of the softmax weights to gradient perturbations. Let $\bm{g} \in \mathbb{R}^N$ and $\tilde{\bm{g}} = \bm{g} + \bm{\epsilon}$ denote the gradient vectors at $\Gamma(t)$ and $\tilde{\Gamma}(t)$ respectively, and let $\odot$ denote the Hadamard product. From \cite{nair2025softmax}, the softmax operator is $(1/(2\tau))$-Lipschitz with respect to any $\ell_p$ norm. Therefore,
\begin{align}
\left\|
\Softmax\!\left(\frac{\tilde{\bm{g}}^2}{\tau}\right)
-
\Softmax\!\left(\frac{\bm{g}^2}{\tau}\right)
\right\|_2
&\le
\frac{1}{2\tau}
\left\|
\tilde{\bm{g}}^2 - \bm{g}^2
\right\|_2.
\end{align}
Using $\tilde{\bm{g}} = \bm{g} + \bm{\epsilon}$, we expand elementwise,
\begin{align}
\tilde{\bm{g}}^2 - \bm{g}^2
&=
(\bm{g} + \bm{\epsilon})^2 - \bm{g}^2 \nonumber \\
&=
2\, \bm{\epsilon} \odot \bm{g} + \bm{\epsilon}^2.
\end{align}
Next,
\begin{align}
\left\|
\tilde{\bm{g}}^2 - \bm{g}^2
\right\|_2
&=
\left\|
2\, \bm{\epsilon} \odot \bm{g} + \bm{\epsilon}^2
\right\|_2 \nonumber \\
&\le
2 \|\bm{\epsilon} \odot \bm{g}\|_2 + \|\bm{\epsilon}^2\|_2.
\end{align}
Since each $g_i^2 \le \|\bm{g}\|_2^2$, we have
\begin{align}
\|\bm{\epsilon} \odot \bm{g}\|_2
&=
\sqrt{\sum_{i=1}^N \epsilon_i^2 g_i^2}\\
&\le
\sqrt{\|\bm{g}\|_2^2 \sum_{i=1}^N \epsilon_i^2}
=
\|\bm{\epsilon}\|_2 \, \|\bm{g}\|_2.
\end{align}
Similarly, $\|\bm{\epsilon}^2\|_2 = \sqrt{\sum_i \epsilon_i^4} \le \sum_i \epsilon_i^2 = \|\bm{\epsilon}\|_2^2$. Combining,
\begin{equation}
\left\|
\Softmax\!\left(\frac{\tilde{\bm{g}}^2}{\tau}\right)
-
\Softmax\!\left(\frac{\bm{g}^2}{\tau}\right)
\right\|_2
\le
\frac{2\, \|\bm{\epsilon}\|_2 \, \|\bm{g}\|_2 + \|\bm{\epsilon}\|_2^2}{2\tau}.
\end{equation}
We apply this to path parameter $t$ by defining $\bm{\epsilon}(t) = \nabla f(\Gamma(t) + t\bm{\eta}) - \nabla f(\Gamma(t))$. Then, Assumption~\ref{as:lip-f} gives $\|\bm{g}(t)\|_2 \le L_f$ and Assumption~\ref{as:lip-g} gives $\|\bm{\epsilon}(t)\|_2 \le L_g\,t\,\|\bm{\eta}\|_2$. So, the pointwise weight perturbation satisfies
\begin{equation}\label{eq:per-step-wf}
  \|\Delta \bm{w}^{*}(t)\|_2
  \le
  \frac{L_f\,L_g\,t\,\|\bm{\eta}\|_2}{\tau} + \mathcal{O}(\|\bm{\eta}\|_2^2).
\end{equation}
The function increment rate along the unperturbed path satisfies $|\dot{f}(t)| = |\nabla f(\Gamma(t)) \cdot \bm{d}| \le L_f\,|\Gamma|$ by Assumption~\ref{as:lip-f}. Integrating the product over $[0,1]$ yields the weight-variation bound,
\begin{equation}\label{eq:wf-limit}
  \|\bm{W}\|_2 \le \int_0^1 \|\Delta \bm{w}^{*}(t)\|_2\;|\dot{f}(t)|\,dt \le \frac{L_f^2\,L_g\,|\Gamma|\,\|\bm{\eta}\|_2}{\tau}\int_0^1 t\,dt + \mathcal{O}(\|\bm{\eta}\|_2^2) = \frac{L_f^2\,L_g\,|\Gamma|\,\|\bm{\eta}\|_2}{2\tau} + \mathcal{O}(\|\bm{\eta}\|_2^2).
\end{equation}
Applying the triangle inequality to $\Delta\mathrm{ERIG} = \bm{M} + \bm{W}$,
\begin{align}
  \|\Delta\mathrm{ERIG}\|_2
  = \|\bm{M} + \bm{W}\|_2
  &\le \|\bm{M}\|_2 + \|\bm{W}\|_2 \\
  &\le
  \underbrace{\left(L_f + \tfrac{1}{2}\,L_g\,|\Gamma|\right)\|\bm{\eta}\|_2}_{\text{bound on }\|\bm{M}\|_2\text{, from \eqref{eq:model-response-bound}}}
  \;+\;
  \underbrace{\frac{L_f^2\,L_g\,|\Gamma|\,\|\bm{\eta}\|_2}{2\tau} + \mathcal{O}(\|\bm{\eta}\|_2^2)}_{\text{bound on }\|\bm{W}\|_2\text{, from \eqref{eq:wf-limit}}}.
\end{align}
Factoring out the model-response bound and defining $R(\tau)$ as the ratio of the two bounds,
\begin{equation}
  \|\Delta\mathrm{ERIG}\|_2
  \le
  \left(L_f + \tfrac{1}{2}\,L_g\,|\Gamma|\right)\|\bm{\eta}\|_2
  \;\cdot\;\bigl(1 + R(\tau)\bigr),
  \qquad
  R(\tau) = \frac{L_f^2\,L_g\,|\Gamma|}{2\tau\,\bigl(L_f + \tfrac{1}{2}\,L_g\,|\Gamma|\bigr)} + \mathcal{O}(\|\bm{\eta}\|_2).
\end{equation}
\qed

\subsubsection{\texorpdfstring{Extension to Networks with ReLU Activations}{Extension to Networks with ReLU Activations}} \label{appx: relu-scope}
Assumptions~\ref{as:lip-f} and~\ref{as:lip-g} hold directly for backbones with smooth activations, which in our evaluation are ViT-Base and DeiT3 with GELU, and EVA-02 and EfficientNet with SiLU. Such smooth activations are moreover the design norm in modern architectures, so the assumptions cover most recent models directly. For the remaining backbones the gradient is not continuous everywhere. ResNet101 is piecewise linear, with $\nabla f$ piecewise constant on the network's linear regions~\cite{hanin2019deep}, while MobileNetV4 mixes ReLU stages with smooth attention blocks. In both cases $\nabla f$ exists almost everywhere and Assumption~\ref{as:lip-g} fails only on the kink boundaries.

The standard remedy is Gaussian smoothing, which convolves $f$ with the density $\varphi_\sigma$ of $\mathcal{N}(\bm{0},\sigma^2\bm{I})$. The mollified network $f_\sigma = f \ast \varphi_\sigma$ is infinitely differentiable ($f_\sigma \in C^\infty$) and satisfies Assumptions~\ref{as:lip-f} and~\ref{as:lip-g} with $L_{f,\sigma} \le L_f$ and $L_{g,\sigma} = \mathcal{O}(L_f/\sigma)$, respectively. Moreover, $f_\sigma$ converges uniformly to $f$ as $\sigma \to 0$; these are classical properties of Gaussian smoothing~\cite{nesterov2017random,duchi2012randomized}. \Cref{cor: decomp} therefore applies verbatim to $f_\sigma$ for every $\sigma > 0$, that is, to an arbitrarily close smooth surrogate of the network. The same device has already been seen in the XAI literature, and it underlies the analysis of explanation geometry of ReLU models~\cite{dombrowski2019explanations}. Additionally, $\nabla f_\sigma$ is exactly the gradient map that SmoothGrad~\cite{smilkov2017smoothgrad} averages. Hence, our results extend to ReLU networks through this smoothing technique, with the decomposition stated for $f_\sigma$ and its constants $L_{f,\sigma}$ and $L_{g,\sigma}$. Consistent with this, the stability gains in \cref{tab: stability-results} are as large on the ReLU-based backbones as on the smooth ones, and the same holds for the adversarial stability results in \cref{tab: astability} (Appendix~\ref{appx: astability}). Our empirical results are therefore not tied to the specific activation function used. Finally, as mentioned in \cref{sec: lipschitz-decomp}, exact Lipschitz constants of deep networks are intractable to compute, and we treat $L_f$ and $L_g$ only as analysis quantities rather than computed ones.

\subsection{Per-step Runtime Distributions} \label{appx: runtime}

\begin{figure}[htb]
    \centering
    \begin{minipage}[c]{0.42\linewidth}
        \centering
        \renewcommand{\arraystretch}{1.2}
        \begin{tabular}{|c|c|c|}
            \hline
            \textbf{Method} & \textbf{Mean (ms/step)} & \textbf{Std (ms/step)} \\
            \hline
            BIG    & $9.372$ & $0.843$ \\
            GIG    & $31.682$ & $0.473$ \\
            IDGI   & $1.894$ & $1.083$ \\
            IG     & $1.586$ & $0.671$ \\
            ERIG (ours)   & $1.983$ & $1.075$ \\
            \hline
        \end{tabular}
        \captionof{table}{\textbf{Runtime:} ERIG costs ${\sim}25\%$ more per step than IG, with the softmax adding only ${\sim}5\%$ over IDGI.}
        \label{tab: runtime-stats}
    \end{minipage}%
    \hfill
    \begin{minipage}[c]{0.53\linewidth}
        \centering
        \includegraphics[width=\linewidth]{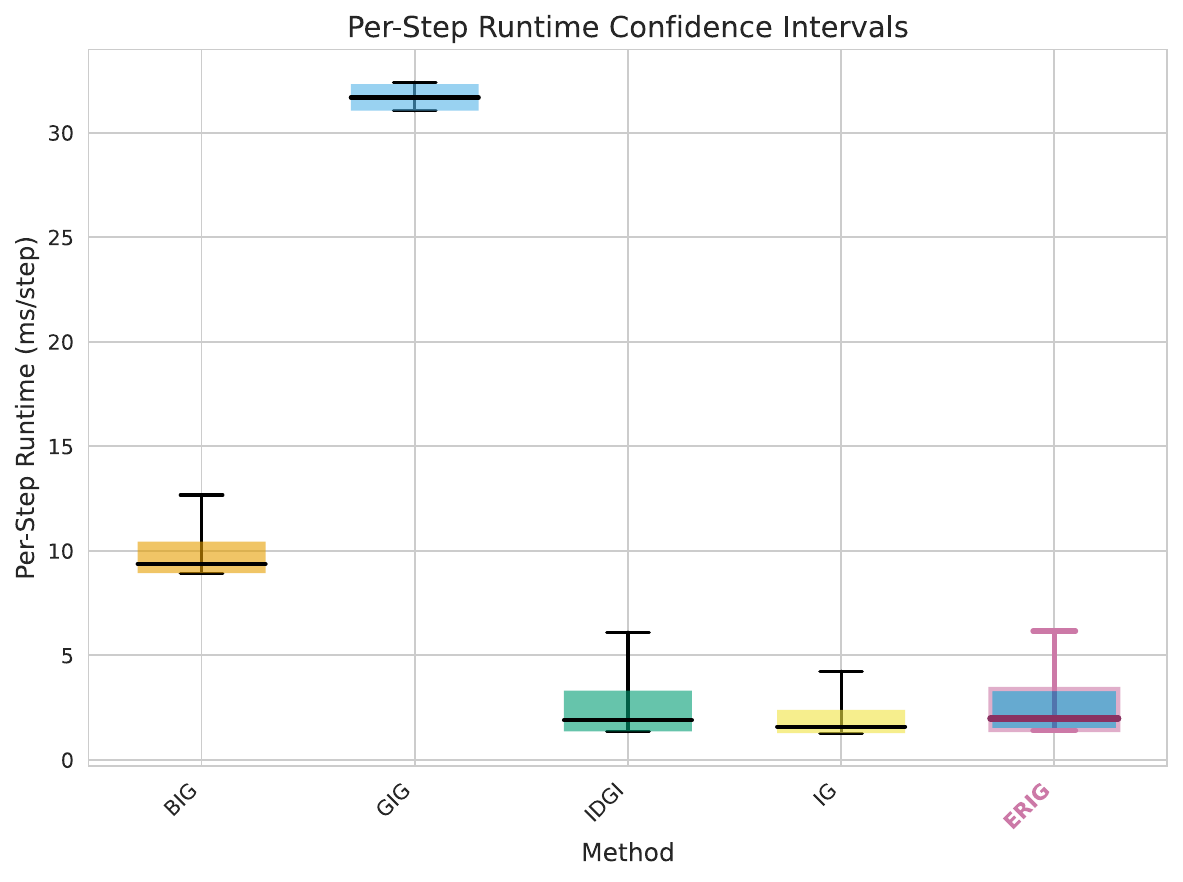}
        \captionof{figure}{Per-step runtime with 5\%--95\% percentile ranges shaded in color. The vertical capped lines indicate the full range across samples. The horizontal lines inside the shaded regions represent the mean values.}
        \label{fig: runtime-ci}
    \end{minipage}
\end{figure}

\noindent All runtime measurements are collected on $1000$ validation samples from the ImageNet dataset and are measured on an NVIDIA A100-SXM4-40GB GPU using PyTorch and the ResNet101 model, with a batch size of $64$. SmoothGrad is excluded here because its ensemble-based design requires $N$ independent attribution passes per example, making it extremely slow and its runtime not directly comparable on a per-step basis with path-based methods. \Cref{tab: runtime-stats} summarizes the per-step statistics for these methods and \cref{fig: runtime-ci} visualizes them with $5\%$--$95\%$ inter-percentile intervals; the shaded regions for IDGI and ERIG roughly align with each other, confirming that their per-step costs are approximately indistinguishable in practice.

\begin{figure}[htbp]
    \centering
    \includegraphics[width=\linewidth]{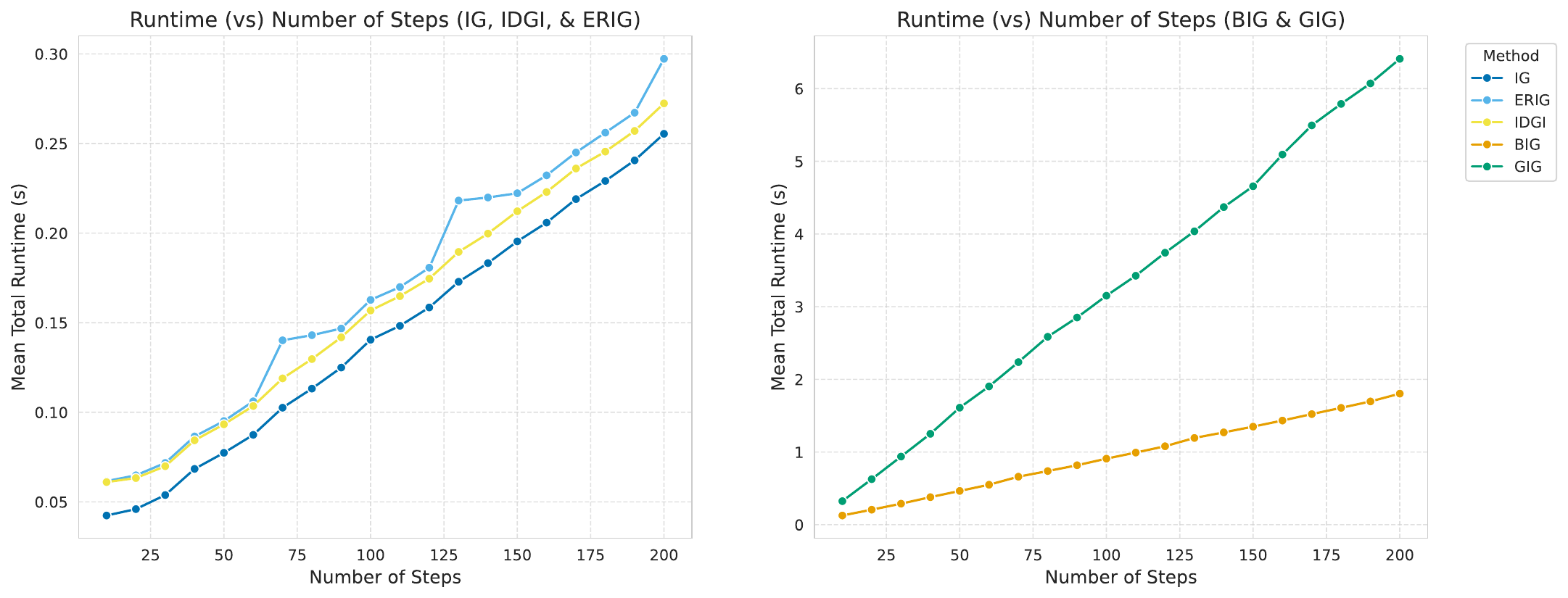}
    \caption{Mean total runtime as a function of the number of integration steps over 1000 examples. The left subplot shows IG, IDGI, and ERIG, which scale linearly and overlap considerably. The right subplot shows BIG and GIG on a separate scale due to their higher costs.}
    \label{fig: runtime-vs-step}
\end{figure}

Particularly, ERIG adds a small overhead relative to IG with its mean per-step runtime ($1.983$\,ms) being ${\sim}0.4$\,ms (${\sim}25\%$) slower than IG ($1.586$\,ms). Additionally, IDGI ($1.894$\,ms) incurs a nearly identical overhead, indicating that the dominant additional cost shared by both methods is the element-wise squaring of the gradient vector, while the softmax normalization that distinguishes ERIG from IDGI adds only ${\sim}0.09$\,ms (${\sim}5\%$) per step. This marginal cost yields substantial gains in attribution stability, localization, and noise characteristics as demonstrated in the main text. All three methods pre-compute the full set of interpolated path points $\{\bm{x}^{(k)}\}_{k=0}^K$ before integration and evaluate them in batched forward-backward passes, yielding linear scaling with the step count (\cref{fig: runtime-vs-step}, left).

BIG is, on average, ${\sim}6\times$ slower per step than IG because each step requires a Gaussian blur of the input, which involves a large spatial convolution in spite of batching. On the other hand, GIG is the slowest method at ${\sim}20\times$ the per-step cost of IG because its path is constructed iteratively with each intermediate point $\bm{x}^{(k)}$ depending on the computation at $\bm{x}^{(k-1)}$. This sequential dependency prevents the path points from being pre-computed and batched together, forcing GIG into $K$ separate forward-backward passes, and making it fundamentally more expensive regardless of hardware (\cref{fig: runtime-vs-step}, right).

\subsection{\texorpdfstring{ERIG Algorithm}{ERIG Algorithm}} \label{appx: erig-algorithm}
The following algorithm provides a practical way of estimating ERIG attributions approximately:
\begin{algorithm}[H]
\caption{Computing ERIG Attributions}\label{alg: erig}
\begin{algorithmic}[1]
\Function{ERIG}{$f, \Gamma, \tau, i$}
    \State $\mathrm{ERIG}_i = 0$
    \For{$\bm{x}^{(k)} \in \Gamma$}
        \State $\delta f \gets f(\bm{x}^{(k)}) - f(\bm{x}^{(k-1)})$
        \State $\mathrm{ERIG}_i \gets \mathrm{ERIG}_i + w_i^{*}(\bm{g}(\bm{x}^{(k)}); \tau) \times \delta f$
    \EndFor
    \State \Return $\mathrm{ERIG}_i$
\EndFunction
\end{algorithmic}
\end{algorithm}

\subsection{SmoothGrad Ensembles} \label{appx: smoothgrad-ensembles}

\begin{table}[htb]
\centering
\begin{minipage}[t]{0.48\textwidth}
    \centering
    \caption{\textbf{SmoothGrad Stability:} The ERIG ensemble improves stability by ${\sim}67\times$ on average over the next-best ensemble.}
    \footnotesize
    \setlength{\tabcolsep}{4pt}
    \begin{tabular}{|c|c|c|c|c|}
        \hline
        \textbf{Model} & \textbf{Method} & \textbf{MS $(\downarrow)$} & \textbf{LLC $(\downarrow)$} & \textbf{RIS $(\downarrow)$}\\
        \hline
        \multirow{5}{*}{ResNet101} & SG + BIG & $0.3116$ & $0.0139$ & $134.050$\\
        & SG + GIG & $0.2648$ & $0.0118$ & $\mathbf{28.502}$\\
        & SG + IDGI & $\mathbf{0.0004}$ & $\mathbf{1.8e\text{\textbf{ --}}05}$ & $138.650$\\
        & SG + IG & $0.2801$ & $0.0125$ & $141.100$\\
        & SG + ERIG (ours) & $\mathbf{\underline{0.0003}}$ & $\mathbf{\underline{1.5e\text{\textbf{ --}}05}}$ & $\mathbf{\underline{4.029}}$\\
        \hline
        \multirow{5}{*}{EfficientNet} & SG + BIG & $0.0393$ & $0.0012$ & $\mathbf{12.868}$\\
        & SG + GIG & $0.1012$ & $0.0032$ & $27.078$\\
        & SG + IDGI & $\mathbf{0.0026}$ & $\mathbf{2.1e\text{\textbf{ --}}05}$ & $14.696$\\
        & SG + IG & $0.2179$ & $0.0068$ & $22.584$\\
        & SG + ERIG (ours) & $\mathbf{\underline{8.2e\text{\textbf{ --}}05}}$ & $\mathbf{\underline{2.5e\text{\textbf{ --}}06}}$ & $\mathbf{\underline{2.003}}$\\
        \hline
        \multirow{5}{*}{MobileNetV4} & SG + BIG & $0.1978$ & $0.0077$ & $16.359$\\
        & SG + GIG & $0.1812$ & $0.0071$ & $31.392$\\
        & SG + IDGI & $\mathbf{0.0025}$ & $\mathbf{9.6e\text{\textbf{ --}}05}$ & $\mathbf{15.627}$\\
        & SG + IG & $0.4141$ & $0.0162$ & $25.626$\\
        & SG + ERIG (ours) & $\mathbf{\underline{2.0e\text{\textbf{ --}}04}}$ & $\mathbf{\underline{6.5e\text{\textbf{ --}}06}}$ & $\mathbf{\underline{3.081}}$\\
        \hline
        \multirow{5}{*}{ViT-Base} & SG + BIG & $0.0658$ & $0.0017$ & $\mathbf{\underline{10.109}}$\\
        & SG + GIG & $0.0715$ & $0.0019$ & $21.334$\\
        & SG + IDGI & $\mathbf{0.0042}$ & $\mathbf{0.0001}$ & $\mathbf{15.440}$\\
        & SG + IG & $0.2662$ & $0.0069$ & $17.865$\\
        & SG + ERIG (ours) & $\mathbf{\underline{0.0023}}$ & $\mathbf{\underline{5.9e\text{\textbf{ --}}05}}$ & $37.984$\\
        \hline
        \multirow{5}{*}{EVA-02} & SG + BIG & $0.0491$ & $0.0011$ & $\mathbf{8.306}$\\
        & SG + GIG & $0.0476$ & $0.0011$ & $20.135$\\
        & SG + IDGI & $\mathbf{0.0044}$ & $\mathbf{9.8e\text{\textbf{ --}}05}$ & $25.772$\\
        & SG + IG & $0.2183$ & $0.0049$ & $21.286$\\
        & SG + ERIG (ours) & $\mathbf{\underline{4.1e\text{\textbf{ --}}05}}$ & $\mathbf{\underline{9.2e{\textbf{ --}}07}}$ & $\mathbf{\underline{1.134}}$\\
        \hline
        \multirow{5}{*}{DeiT3} & SG + BIG & $0.1153$ & $0.0030$ & $\mathbf{10.650}$\\
        & SG + GIG & $0.0561$ & $0.0015$ & $33.567$\\
        & SG + IDGI & $\mathbf{0.0148}$ & $\mathbf{0.0004}$ & $26.624$\\
        & SG + IG & $0.1813$ & $0.0047$ & $30.512$\\
        & SG + ERIG (ours) & $\mathbf{\underline{3.4e\text{\textbf{ --}}05}}$ & $\mathbf{\underline{8.9e\text{\textbf{ --}}07}}$ & $\mathbf{\underline{0.730}}$\\
        \hline
    \end{tabular}
    \label{tab: smoothgrad-stability-results}
\end{minipage}
\hfill
\begin{minipage}[t]{0.48\textwidth}
    \centering
    \caption{\textbf{SmoothGrad Noise:} The ERIG ensemble attains the highest LINV in every cell, exceeding second by $4.8$ to $11.8$ units.}
    \footnotesize
    \begin{tabular}{|c|c|c|c|}
        \hline
        \textbf{Model} & \textbf{Method} & \textbf{LINV-C $(\uparrow)$} & \textbf{LINV-P $(\uparrow)$}\\
        \hline
        \multirow{5}{*}{ResNet101} & SG + BIG & $8.0804$ & $7.9269$\\
        & SG + GIG & $8.2262$ & $8.2521$\\
        & SG + IDGI & $\mathbf{13.9169}$ & $\mathbf{13.8936}$\\
        & SG + IG & $8.2707$ & $8.1122$\\
        & SG + ERIG (ours) & $\mathbf{\underline{18.8182}}$ & $\mathbf{\underline{18.6851}}$\\
        \hline
        \multirow{5}{*}{EfficientNet} & SG + BIG & $8.2824$ & $8.4709$\\
        & SG + GIG & $8.2517$ & $8.1945$\\
        & SG + IDGI & $\mathbf{11.0754}$ & $\mathbf{11.0537}$\\
        & SG + IG & $7.4786$ & $7.4437$\\
        & SG + ERIG (ours) & $\mathbf{\underline{22.4027}}$ & $\mathbf{\underline{22.3890}}$\\
        \hline
        \multirow{5}{*}{MobileNetV4} & SG + BIG & $6.6509$ & $6.6404$\\
        & SG + GIG & $7.5227$ & $7.3665$\\
        & SG + IDGI & $\mathbf{11.0293}$ & $\mathbf{10.9747}$\\
        & SG + IG & $6.7344$ & $6.5129$\\
        & SG + ERIG (ours) & $\mathbf{\underline{20.2329}}$ & $\mathbf{\underline{20.1801}}$\\
        \hline
        \multirow{5}{*}{ViT-Base} & SG + BIG & $8.4789$ & $8.4251$\\
        & SG + GIG & $8.9195$ & $8.6938$\\
        & SG + IDGI & $\mathbf{11.0744}$ & $\mathbf{10.9126}$\\
        & SG + IG & $7.7517$ & $7.4866$\\
        & SG + ERIG (ours) & $\mathbf{\underline{21.6590}}$ & $\mathbf{\underline{21.4986}}$\\
        \hline
        \multirow{5}{*}{EVA-02} & SG + BIG & $8.7818$ & $8.7877$\\
        & SG + GIG & $9.3968$ & $9.1246$\\
        & SG + IDGI & $\mathbf{11.4790}$ & $\mathbf{11.4074}$\\
        & SG + IG & $8.2156$ & $8.0836$\\
        & SG + ERIG (ours) & $\mathbf{\underline{22.7220}}$ & $\mathbf{\underline{22.6599}}$\\
        \hline
        \multirow{5}{*}{DeiT3} & SG + BIG & $7.8633$ & $8.2038$\\
        & SG + GIG & $9.2874$ & $9.2294$\\
        & SG + IDGI & $\mathbf{10.4138}$ & $\mathbf{10.3527}$\\
        & SG + IG & $8.2140$ & $8.2911$\\
        & SG + ERIG (ours) & $\mathbf{\underline{22.1643}}$ & $\mathbf{\underline{22.1306}}$\\
        \hline
    \end{tabular}
    \label{tab: smoothgrad-noise-results}
\end{minipage}
\end{table}

To evaluate whether the intrinsic properties of ERIG complement external stochastic smoothing techniques, we construct ensembles of each attribution method with SmoothGrad. To ensure a principled comparison with the vanilla methods discussed in the main text, we enforce a fixed computational budget of 200 total forward and backward passes. Specifically, we utilize $N=10$ noise samples with $S=20$ integration steps per sample ($10 \times 20 = 200$). While Smilkov et al.~\cite{smilkov2017smoothgrad} suggest diminishing returns beyond approximately 50 noise samples for standalone SmoothGrad, path-based methods in an ensemble setting require a sufficiently high step count ($\geq 20$) to maintain a reliable Riemann sum approximation. This configuration represents a practical equilibrium between smoothing variance reduction and integration accuracy under strict resource constraints.

We maintain a fixed temperature of $\tau = 3$ for ERIG throughout these evaluations, consistent with the stability and noise experiments in the main text. We acknowledge that since SmoothGrad introduces external Gaussian noise which interacts with ERIG's temperature-controlled weighting, the optimal temperature ($\tau$) for ensembles may deviate from the vanilla setting. A joint optimization over the hyper-parameter space of $(\tau, \text{number of samples}, \text{integration steps})$ is deferred to future work. Nonetheless, even with this untuned $\tau$, the ERIG ensemble achieves superior stability and noise scores across nearly all architectures (Tables~\ref{tab: smoothgrad-stability-results}--\ref{tab: smoothgrad-noise-results}). These results demonstrate that ERIG's intrinsic stability and noise reduction mechanisms are orthogonal to external smoothing, confirming its effectiveness across diverse computational regimes.

\begin{figure*}[htb]
    \centering
    \begin{subfigure}[b]{0.48\linewidth}
        \centering
        \includegraphics[width=\linewidth]{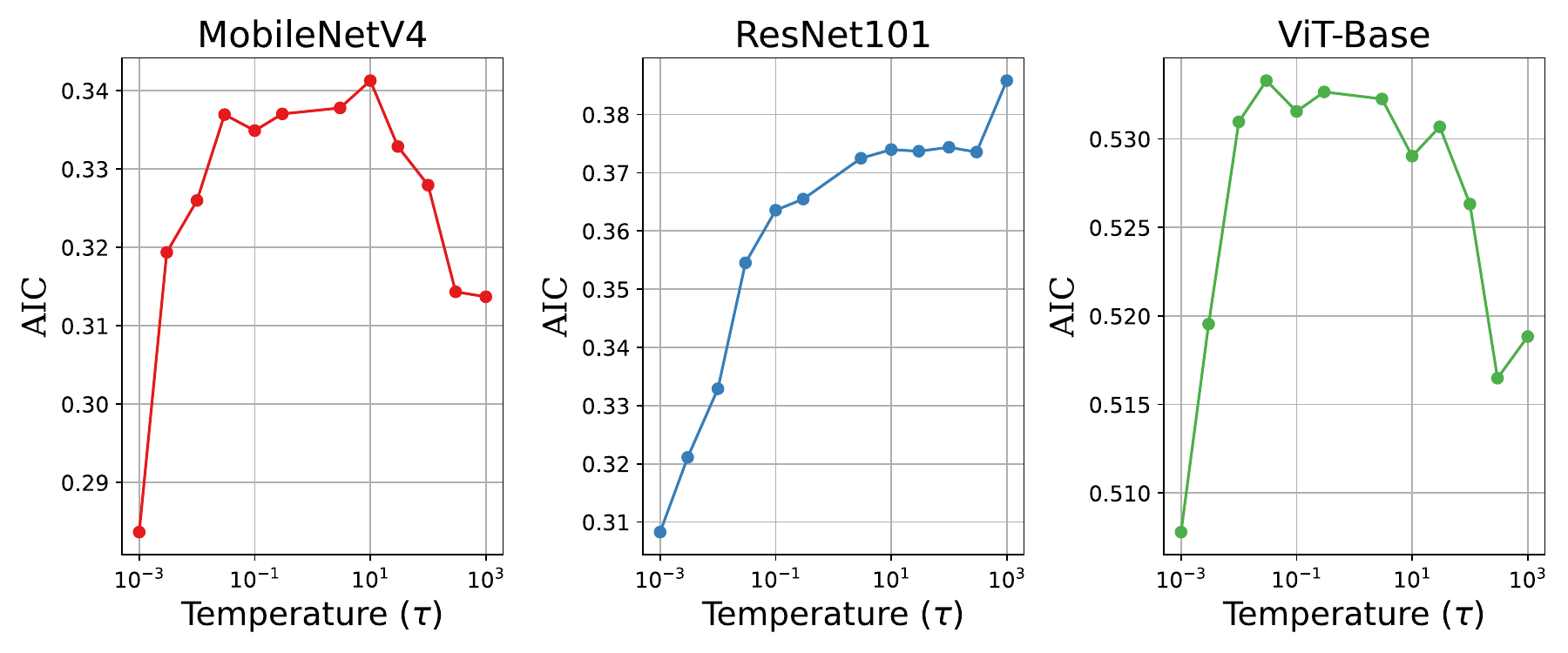}
        \caption{Fidelity (AIC) vs.\ $\tau$}
        \label{fig:sg-sub1}
    \end{subfigure}
    \hfill
    \begin{subfigure}[b]{0.48\linewidth}
        \centering
        \includegraphics[width=\linewidth]{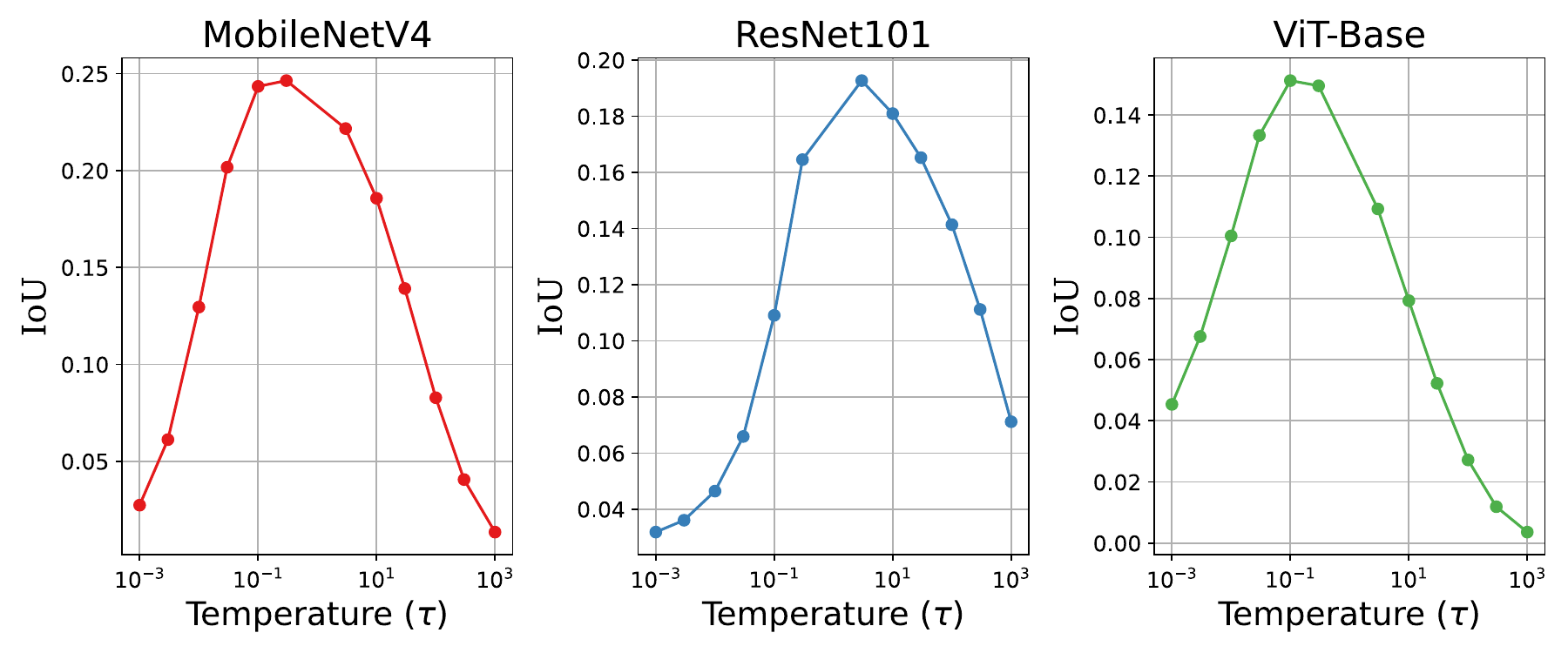}
        \caption{Localization (IoU) vs.\ $\tau$}
        \label{fig:sg-sub2}
    \end{subfigure}
    \medskip
    \begin{subfigure}[b]{0.48\linewidth}
        \centering
        \includegraphics[width=\linewidth]{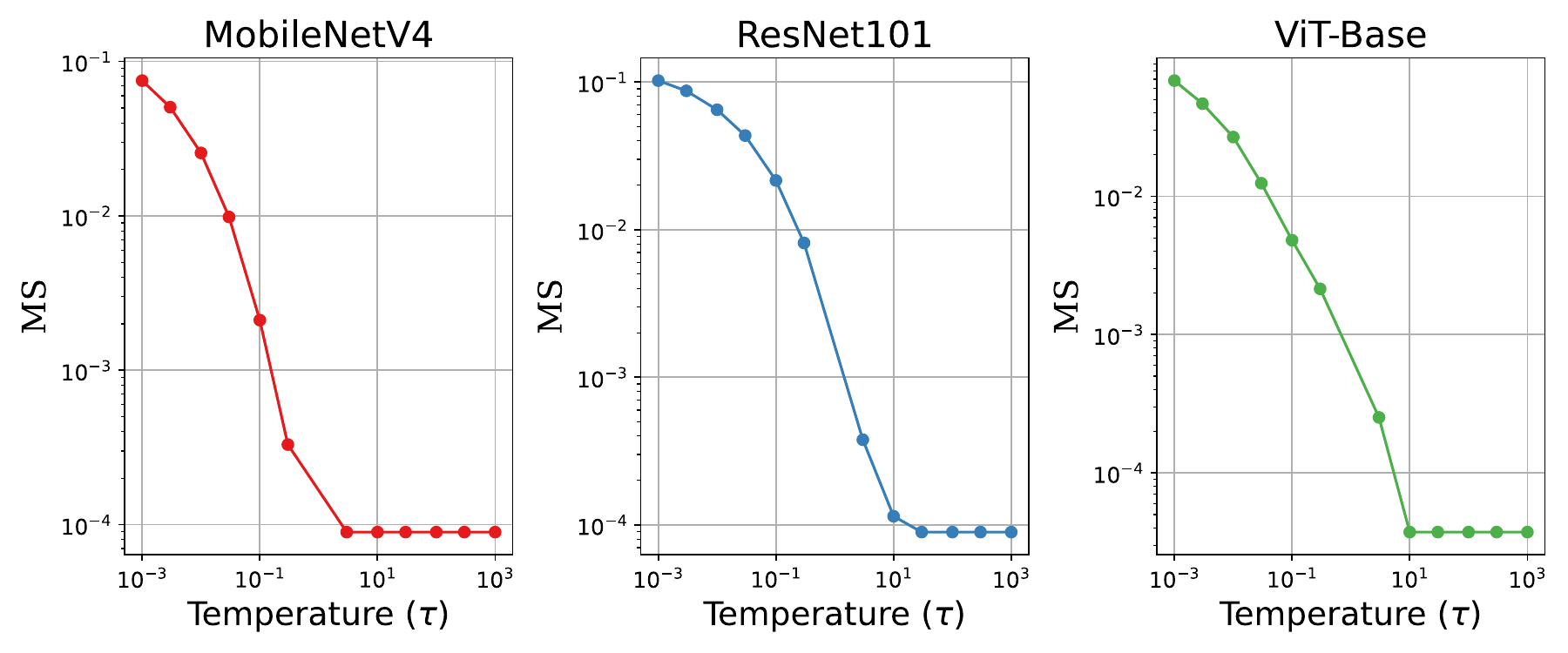}
        \caption{Stability (MS) vs.\ $\tau$}
        \label{fig:sg-sub3}
    \end{subfigure}
    \hfill
    \begin{subfigure}[b]{0.48\linewidth}
        \centering
        \includegraphics[width=\linewidth]{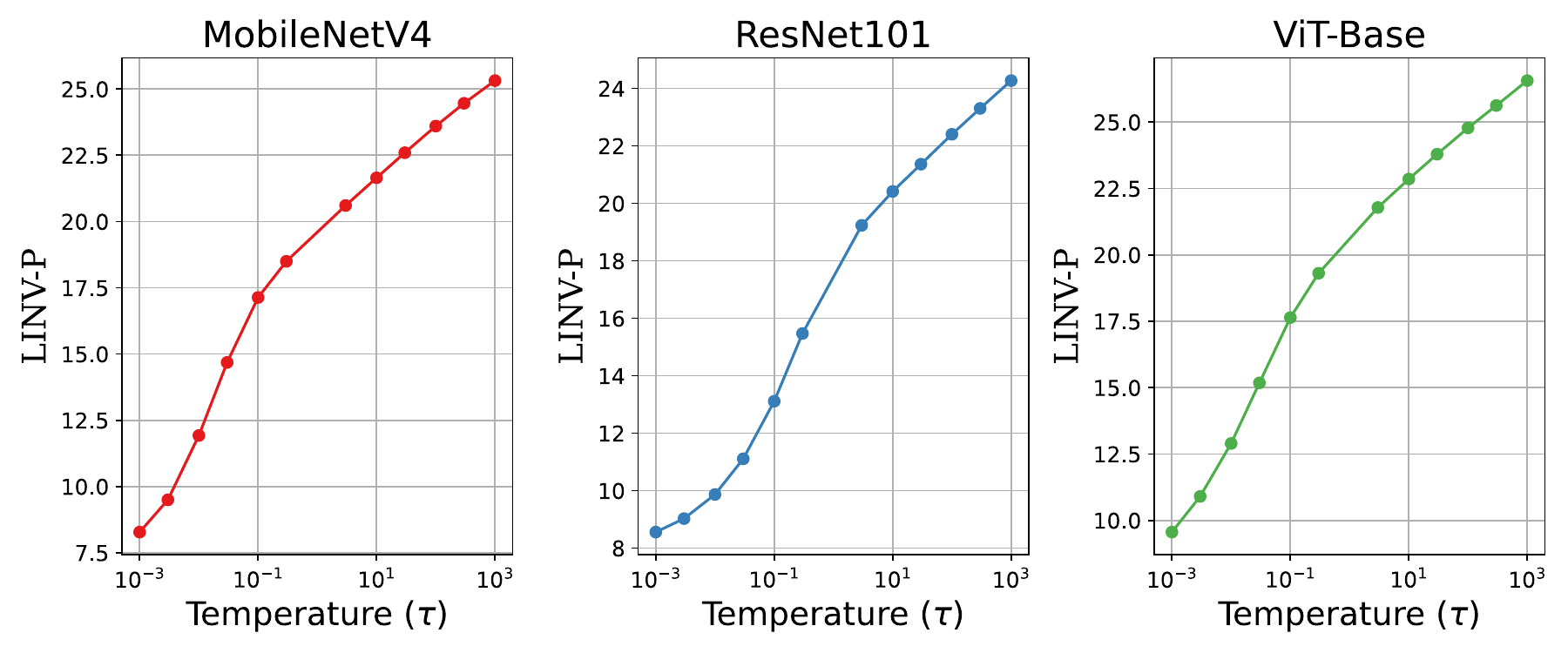}
        \caption{Noise (LINV-P) vs.\ $\tau$}
        \label{fig:sg-sub4}
    \end{subfigure}
    \caption{Ablation of the softmax temperature $\tau$ in SmoothGrad+ERIG ensembles across MobileNetV4, ResNet101, and ViT-Base models. All plots use a log-scale x-axis and a linear-scale y-axis, except stability, which also uses a log-scale y-axis.}
    \label{fig: sg-temperature-ablation}
\end{figure*}

We are also interested in how the optimal $\tau$ is affected when SmoothGrad's external smoothing interacts with ERIG's intrinsic smoothing. To this end, we sweep the ensemble temperature over $\tau \in [10^{-3}, 10^{3}]$, using $50$ noise samples and $20$ path steps per sample, on MobileNetV4, ResNet101, and ViT-Base. \Cref{fig: sg-temperature-ablation} reports the same four metrics as \cref{fig: temperature-ablation}, and the resulting pattern closely mirrors that observed in the latter. Fidelity remains on a broad plateau around the default value, localization peaks at an intermediate $\tau \in [0.1, 3]$, stability likewise reaches its minimum around these values, and noise reduction improves monotonically with $\tau$. The default $\tau = 3$ therefore remains a meaningful choice under ensembling as well.

\subsection{\texorpdfstring{Generalization Beyond ImageNet}{Generalization Beyond ImageNet}} \label{appx: beyond-imagenet}

To verify that our conclusions are not specific to the ImageNet family, we repeat the full evaluation protocol in two additional settings: the Oxford-IIIT Pet dataset~\cite{parkhi2012cats} and the FunnyBirds benchmark~\cite{hesse2023funnybirds}. This section provides the complete results behind the summary in \Cref{subsec: beyond-imagenet} of the main text.

\subsubsection{\texorpdfstring{Oxford-IIIT Pet}{Oxford-IIIT Pet}} \label{appx: pets}
Oxford-IIIT Pet~\cite{parkhi2012cats} is a 37-class fine-grained classification dataset that ships a pixel-aligned segmentation trimap with every image. Hence fidelity, stability, noise, \emph{and} localization can all be evaluated within one non-ImageNet domain; for localization, the trimap's boundary band is counted as background. For meaningful results, we evaluate the methods on models with good accuracy on the dataset. We therefore use five domain-fine-tuned public checkpoints spanning four architecture families, namely ResNet-50, MobileNetV2, ConvNeXt-B, ViT-B, and DINOv2-S. \Cref{tab: pets-fidelity,tab: pets-localization,tab: pets-stability,tab: pets-noise} report the results. ERIG remains competitive on fidelity, trailing the strongest baseline IDGI, on average, by only $-0.40\%$, and outperforms the second-best method on localization by $20.2\%$ on average across model--metric combinations. It additionally takes 24 of the 25 stability and noise cells, lowering MS and LLC by up to two orders of magnitude and posting LINV scores $6$ to $8.7$ units higher than those of the next-best method.

\begin{table}[htb]
    \centering
    \begin{minipage}[t]{0.48\textwidth}
        \centering
        \caption{\textbf{Oxford-IIIT Pet Fidelity:} ERIG is competitive and only trails the strongest baseline by $0.40\%$ on average.}
        \footnotesize
        \begin{tabular}{|c|c|c|c|c|}
            \hline
            \textbf{Model} & \textbf{Method} & \textbf{ISC $(\uparrow)$} & \textbf{AIC $(\uparrow)$} & \textbf{DEL $(\downarrow)$}\\
            \hline
            \multirow{6}{*}{ResNet50} & IG & 0.1131 & 0.3863 & 0.3668 \\
             & IDGI & \textbf{0.5649} & \textbf{0.8211} & \textbf{0.2497} \\
             & BIG & 0.1102 & 0.3265 & 0.3648 \\
             & GIG & 0.1579 & 0.4053 & 0.5246 \\
             & SmoothGrad & 0.1075 & 0.2868 & 0.6394 \\
             & ERIG (ours) & \textbf{\underline{0.5704}} & \textbf{\underline{0.8228}} & \textbf{\underline{0.2491}} \\
            \hline
            \multirow{6}{*}{MobileNetV2} & IG & 0.1204 & 0.2520 & 0.3128 \\
             & IDGI & \textbf{0.5873} & \textbf{\underline{0.6464}} & \textbf{0.2208} \\
             & BIG & 0.1027 & 0.1981 & 0.2836 \\
             & GIG & 0.1620 & 0.2807 & 0.3103 \\
             & SmoothGrad & 0.0918 & 0.1592 & 0.5581 \\
             & ERIG (ours) & \textbf{\underline{0.5898}} & \textbf{0.6320} & \textbf{\underline{0.2194}} \\
            \hline
            \multirow{6}{*}{ConvNeXt-B} & IG & 0.8434 & 0.8805 & 0.6499 \\
             & IDGI & \textbf{\underline{0.8588}} & \textbf{0.9002} & \textbf{0.5568} \\
             & BIG & 0.7725 & 0.8521 & 0.6429 \\
             & GIG & 0.8211 & 0.8744 & 0.6724 \\
             & SmoothGrad & 0.4073 & 0.4982 & 0.8228 \\
             & ERIG (ours) & \textbf{0.8577} & \textbf{\underline{0.9029}} & \textbf{\underline{0.5562}} \\
            \hline
            \multirow{6}{*}{ViT-B} & IG & 0.3919 & 0.6223 & 0.4886 \\
             & IDGI & \textbf{\underline{0.6349}} & \textbf{\underline{0.7463}} & \textbf{\underline{0.3680}} \\
             & BIG & 0.2949 & 0.5332 & 0.5286 \\
             & GIG & 0.4876 & 0.6409 & 0.4345 \\
             & SmoothGrad & 0.1624 & 0.2920 & 0.6944 \\
             & ERIG (ours) & \textbf{0.6222} & \textbf{0.7327} & \textbf{0.3808} \\
            \hline
            \multirow{6}{*}{DINOv2-S} & IG & 0.6193 & 0.7217 & 0.4638 \\
             & IDGI & \textbf{0.7104} & \textbf{0.8078} & \textbf{0.3783} \\
             & BIG & 0.4322 & 0.5879 & 0.4863 \\
             & GIG & 0.5914 & 0.7492 & 0.3766 \\
             & SmoothGrad & 0.2010 & 0.2518 & 0.6781 \\
             & ERIG (ours) & \textbf{\underline{0.7150}} & \textbf{\underline{0.8083}} & \textbf{\underline{0.3766}} \\
            \hline
        \end{tabular}
        \label{tab: pets-fidelity}
    \end{minipage}
    \hfill
    \begin{minipage}[t]{0.48\textwidth}
        \centering
        \caption{\textbf{Oxford-IIIT Pet Localization:} ERIG improves localization by $20.2\%$ on average over the second-best method.}
        \footnotesize
        \begin{tabular}{|c|c|c|c|c|}
            \hline
            \textbf{Model} & \textbf{Method} & \textbf{IoU $(\uparrow)$} & \textbf{PA $(\uparrow)$} & \textbf{PF1 $(\uparrow)$}\\
            \hline
            \multirow{6}{*}{ResNet50} & IG & 0.0829 & 0.6131 & 0.1348 \\
             & IDGI & 0.0200 & 0.5938 & 0.0361 \\
             & BIG & 0.0531 & 0.6030 & 0.0901 \\
             & GIG & 0.0251 & 0.5949 & 0.0442 \\
             & SmoothGrad & \textbf{0.1122} & \textbf{0.6232} & \textbf{0.1760} \\
             & ERIG (ours) & \textbf{\underline{0.1534}} & \textbf{\underline{0.6340}} & \textbf{\underline{0.2509}} \\
            \hline
            \multirow{6}{*}{MobileNetV2} & IG & 0.0753 & 0.5700 & 0.1245 \\
             & IDGI & 0.0218 & 0.5536 & 0.0393 \\
             & BIG & 0.0475 & 0.5600 & 0.0819 \\
             & GIG & 0.0394 & 0.5556 & 0.0666 \\
             & SmoothGrad & \textbf{0.0993} & \textbf{0.5740} & \textbf{0.1602} \\
             & ERIG (ours) & \textbf{\underline{0.1542}} & \textbf{\underline{0.5962}} & \textbf{\underline{0.2507}} \\
            \hline
            \multirow{6}{*}{ConvNeXt-B} & IG & 0.0560 & 0.5607 & 0.0941 \\
             & IDGI & 0.0088 & 0.5476 & 0.0166 \\
             & BIG & 0.0450 & 0.5547 & 0.0776 \\
             & GIG & 0.0745 & 0.5474 & 0.1243 \\
             & SmoothGrad & \textbf{\underline{0.1057}} & \textbf{0.5662} & \textbf{0.1661} \\
             & ERIG (ours) & \textbf{0.1033} & \textbf{\underline{0.5713}} & \textbf{\underline{0.1773}} \\
            \hline
            \multirow{6}{*}{ViT-B} & IG & 0.0828 & 0.6986 & 0.1370 \\
             & IDGI & 0.0169 & 0.6923 & 0.0309 \\
             & BIG & 0.0592 & 0.6955 & 0.1009 \\
             & GIG & 0.0502 & 0.6953 & 0.0862 \\
             & SmoothGrad & \textbf{0.1287} & \textbf{\underline{0.7141}} & \textbf{0.2013} \\
             & ERIG (ours) & \textbf{\underline{0.1405}} & \textbf{0.7025} & \textbf{\underline{0.2335}} \\
            \hline
            \multirow{6}{*}{DINOv2-S} & IG & 0.0477 & 0.5518 & 0.0820 \\
             & IDGI & 0.0072 & 0.5412 & 0.0138 \\
             & BIG & 0.0273 & 0.5448 & 0.0493 \\
             & GIG & 0.0266 & 0.5453 & 0.0477 \\
             & SmoothGrad & \textbf{0.0716} & \textbf{0.5589} & \textbf{0.1193} \\
             & ERIG (ours) & \textbf{\underline{0.0973}} & \textbf{\underline{0.5606}} & \textbf{\underline{0.1690}} \\
            \hline
        \end{tabular}
        \label{tab: pets-localization}
    \end{minipage}
\end{table}

\begin{table}
    \centering
    \begin{minipage}[t]{0.48\textwidth}
        \centering
        \caption{\textbf{Oxford-IIIT Pet Stability:} ERIG improves stability by ${\sim}38\times$ on average over the next-best method.}
        \footnotesize
        \begin{tabular}{|c|c|c|c|c|}
            \hline
            \textbf{Model} & \textbf{Method} & \textbf{MS $(\downarrow)$} & \textbf{LLC $(\downarrow)$} & \textbf{RIS $(\downarrow)$}\\
            \hline
            \multirow{6}{*}{ResNet50} & IG & 0.6025 & 0.0269 & 23.753 \\
             & IDGI & \textbf{0.0115} & $\mathbf{5.1e\text{\textbf{--}}04}$ & 24.485 \\
             & BIG & 0.1094 & 0.0049 & \textbf{13.453} \\
             & GIG & 0.1933 & 0.0086 & 37.823 \\
             & SmoothGrad & 0.0863 & 0.0039 & 21.475 \\
             & ERIG (ours) & $\mathbf{\underline{1.3e\text{\textbf{--}}04}}$ & $\mathbf{\underline{5.8e\text{\textbf{--}}06}}$ & \textbf{\underline{1.597}} \\
            \hline
            \multirow{6}{*}{MobileNetV2} & IG & 1.5538 & 0.0694 & 16.533 \\
             & IDGI & \textbf{0.0121} & $\mathbf{5.4e\text{\textbf{--}}04}$ & \textbf{10.858} \\
             & BIG & 0.1796 & 0.0080 & 14.231 \\
             & GIG & 0.3133 & 0.0140 & 22.746 \\
             & SmoothGrad & 0.3353 & 0.0150 & 11.542 \\
             & ERIG (ours) & $\mathbf{\underline{4.3e\text{\textbf{--}}04}}$ & $\mathbf{\underline{1.9e\text{\textbf{--}}05}}$ & \textbf{\underline{1.838}} \\
            \hline
            \multirow{6}{*}{ConvNeXt-B} & IG & 0.2638 & 0.0118 & 11.672 \\
             & IDGI & \textbf{0.0049} & $\mathbf{2.2e\text{\textbf{--}}04}$ & 6.387 \\
             & BIG & 0.0412 & 0.0018 & \textbf{4.154} \\
             & GIG & 0.1334 & 0.0060 & 45.412 \\
             & SmoothGrad & 0.0303 & 0.0014 & 13.726 \\
             & ERIG (ours) & $\mathbf{\underline{6.0e\text{\textbf{--}}05}}$ & $\mathbf{\underline{2.7e\text{\textbf{--}}06}}$ & \textbf{\underline{0.546}} \\
            \hline
            \multirow{6}{*}{ViT-B} & IG & 0.6060 & 0.0270 & \textbf{10.238} \\
             & IDGI & \textbf{0.0105} & \textbf{0.0005} & 11.206 \\
             & BIG & 0.1630 & 0.0073 & \textbf{\underline{9.854}} \\
             & GIG & 0.3275 & 0.0146 & 22.302 \\
             & SmoothGrad & 0.2084 & 0.0093 & 11.156 \\
             & ERIG (ours) & \textbf{\underline{0.0064}} & \textbf{\underline{0.0003}} & 25.986 \\
            \hline
            \multirow{6}{*}{DINOv2-S} & IG & 0.1669 & 0.0075 & 6.395 \\
             & IDGI & \textbf{0.0035} & $\mathbf{1.6e\text{\textbf{--}}04}$ & 5.402 \\
             & BIG & 0.0380 & 0.0017 & \textbf{3.749} \\
             & GIG & 0.2570 & 0.0115 & 51.372 \\
             & SmoothGrad & 0.0506 & 0.0023 & 12.857 \\
             & ERIG (ours) & $\mathbf{\underline{5.2e\text{\textbf{--}}05}}$ & $\mathbf{\underline{2.3e\text{\textbf{--}}06}}$ & \textbf{\underline{0.574}} \\
            \hline
        \end{tabular}
        \label{tab: pets-stability}
    \end{minipage}
    \hfill
    \begin{minipage}[t]{0.48\textwidth}
        \centering
        \caption{\textbf{Oxford-IIIT Pet Noise:} ERIG attains the highest LINV on every backbone, exceeding second by $6.0$ to $8.7$ units.}
        \footnotesize
        \begin{tabular}{|c|c|c|c|}
            \hline
            \textbf{Model} & \textbf{Method} & \textbf{LINV-C $(\uparrow)$} & \textbf{LINV-P $(\uparrow)$}\\
            \hline
            \multirow{6}{*}{ResNet50} & IG & 5.9780 & 5.8854 \\
             & IDGI & \textbf{9.6196} & \textbf{9.6044} \\
             & BIG & 6.8119 & 6.7880 \\
             & GIG & 7.4204 & 7.3881 \\
             & SmoothGrad & 7.8837 & 7.7788 \\
             & ERIG (ours) & \textbf{\underline{18.3365}} & \textbf{\underline{18.3209}} \\
            \hline
            \multirow{6}{*}{MobileNetV2} & IG & 5.7686 & 5.5927 \\
             & IDGI & \textbf{9.5097} & \textbf{9.4758} \\
             & BIG & 7.1561 & 7.1005 \\
             & GIG & 7.1572 & 7.1022 \\
             & SmoothGrad & 6.8095 & 6.5869 \\
             & ERIG (ours) & \textbf{\underline{17.2354}} & \textbf{\underline{17.2044}} \\
            \hline
            \multirow{6}{*}{ConvNeXt-B} & IG & 7.5820 & 7.5434 \\
             & IDGI & 9.3397 & 9.2465 \\
             & BIG & 7.3031 & 7.3255 \\
             & GIG & 7.9877 & 7.9592 \\
             & SmoothGrad & \textbf{10.9739} & \textbf{10.9233} \\
             & ERIG (ours) & \textbf{\underline{17.0081}} & \textbf{\underline{16.9278}} \\
            \hline
            \multirow{6}{*}{ViT-B} & IG & 7.1427 & 7.1941 \\
             & IDGI & \textbf{10.2746} & \textbf{10.2591} \\
             & BIG & 7.7444 & 7.7675 \\
             & GIG & 7.3616 & 7.3296 \\
             & SmoothGrad & 8.6479 & 8.7419 \\
             & ERIG (ours) & \textbf{\underline{17.9810}} & \textbf{\underline{17.9928}} \\
            \hline
            \multirow{6}{*}{DINOv2-S} & IG & 7.2419 & 7.0304 \\
             & IDGI & \textbf{9.4936} & \textbf{9.3007} \\
             & BIG & 7.3595 & 6.9305 \\
             & GIG & 7.3323 & 7.2198 \\
             & SmoothGrad & 8.8473 & 8.5305 \\
             & ERIG (ours) & \textbf{\underline{18.2027}} & \textbf{\underline{18.0135}} \\
            \hline
        \end{tabular}
        \label{tab: pets-noise}
    \end{minipage}
\end{table}

\subsubsection{\texorpdfstring{FunnyBirds}{FunnyBirds}} \label{appx: funnybirds}

\begin{table}[htb]
    \centering
    \caption{\textbf{FunnyBirds Protocol:} ERIG places first or second in all 21 cells, leading completeness by $3.4\%$ and contrastivity by $2.1\%$.}
    \small
    \begin{tabular}{|c|c|c:c:c:c|c|c|c|}
        \hline
        \multicolumn{2}{|c|}{} & \multicolumn{5}{c|}{\textbf{Completeness}} & \multicolumn{1}{c|}{\textbf{Correctness}} & \multicolumn{1}{c|}{\textbf{Contrastivity}}\\
        \hline
        \textbf{Model} & \textbf{Method} & \textbf{CSDC} & \textbf{PC} & \textbf{DC} & \textbf{Dist} & \textbf{Com.} & \textbf{SD} & \textbf{TS}\\
        \hline
        \multirow{6}{*}{ResNet-50} & IG & 0.920 & 0.940 & 0.880 & \textbf{0.810} & \textbf{0.862} & 0.590 & \textbf{0.980} \\
        & IDGI & \textbf{0.934} & \textbf{\underline{0.958}} & \textbf{\underline{0.912}} & 0.788 & 0.861 & \textbf{\underline{0.615}} & 0.958 \\
        & BIG & 0.794 & 0.700 & 0.676 & 0.699 & 0.711 & 0.440 & 0.491 \\
        & GIG & 0.765 & 0.684 & 0.646 & 0.799 & 0.749 & 0.437 & 0.780 \\
        & SmoothGrad & 0.827 & 0.836 & 0.810 & 0.795 & 0.810 & 0.545 & 0.873 \\
        & ERIG (ours) & \textbf{\underline{0.968}} & \textbf{0.951} & \textbf{0.905} & \textbf{\underline{0.872}} & \textbf{\underline{0.907}} & \textbf{0.608} & \textbf{\underline{0.985}} \\
        \hline
        \multirow{6}{*}{VGG16} & IG & 0.920 & 0.920 & 0.920 & \textbf{0.970} & 0.945 & 0.670 & 0.920 \\
        & IDGI & \textbf{0.945} & \textbf{\underline{0.972}} & \textbf{\underline{0.958}} & 0.949 & \textbf{0.954} & 0.688 & \textbf{0.931} \\
        & BIG & 0.766 & 0.668 & 0.692 & 0.723 & 0.716 & 0.612 & 0.582 \\
        & GIG & 0.770 & 0.686 & 0.694 & 0.840 & 0.778 & 0.588 & 0.751 \\
        & SmoothGrad & 0.937 & 0.820 & 0.854 & 0.824 & 0.847 & \textbf{\underline{0.709}} & 0.903 \\
        & ERIG (ours) & \textbf{\underline{0.974}} & \textbf{0.964} & \textbf{0.949} & \textbf{\underline{0.978}} & \textbf{\underline{0.970}} & \textbf{0.698} & \textbf{\underline{0.955}} \\
        \hline
        \multirow{6}{*}{ViT-B} & IG & 0.890 & 0.860 & 0.850 & \textbf{0.900} & 0.883 & 0.650 & 0.910 \\
        & IDGI & 0.928 & \textbf{\underline{0.941}} & \textbf{\underline{0.917}} & 0.871 & \textbf{0.900} & 0.664 & \textbf{0.918} \\
        & BIG & 0.761 & 0.650 & 0.708 & 0.735 & 0.721 & 0.668 & 0.533 \\
        & GIG & 0.738 & 0.628 & 0.698 & 0.841 & 0.764 & 0.663 & 0.659 \\
        & SmoothGrad & \textbf{0.937} & 0.826 & 0.880 & 0.834 & 0.857 & \textbf{\underline{0.780}} & 0.856 \\
        & ERIG (ours) & \textbf{\underline{0.961}} & \textbf{0.933} & \textbf{0.909} & \textbf{\underline{0.924}} & \textbf{\underline{0.929}} & \textbf{0.766} & \textbf{\underline{0.948}} \\
        \hline
    \end{tabular}
    \label{tab: funnybirds}
\end{table}
FunnyBirds~\cite{hesse2023funnybirds} is a synthetic vision dataset purpose-built for part-based analysis of XAI methods. Every image is assembled from annotated bird parts, so the framework can map an attribution map to per-part importances and score it through part-level interventions. We evaluate ERIG and all compared path methods under its standard protocol, which evaluates along the completeness, correctness, and contrastivity dimensions, using the three checkpoints provided by the framework, namely ResNet-50, VGG16, and ViT-B. All metrics lie in $[0,1]$ and higher is better. Following~\cite{hesse2023funnybirds}, let $x_m$ denote the $m$-th sample of $M$ test images with label $c_m$ and let $f(x_m)$ denote the model prediction. The part importance $\mathrm{PI}_p(x_m)$ of a bird part $p$ is the sum of the attribution values inside the part's segmentation mask, dilated by a $5{\times}5$ kernel to include the part edges, and the important-part set is $\mathrm{P}(x_m) = \bigl\{p \mid \mathrm{PI}_p(x_m) > t\, S(x_m)\bigr\}$, where $S(x_m)$ is the total attribution of the map and the protocol sweeps the threshold $t$ over $[0.01, 0.50]$, selecting it automatically to optimally trade off the comprehensiveness checks against distractibility. We explain the metrics below:

\begin{itemize}
    \item \emph{Controlled synthetic data check (CSDC, completeness).} With $\mathcal{P}'_{c_m,i}$ the part subsets known to suffice for class $c_m$,
    \begin{equation}
        \mathrm{CSDC} = \frac{1}{M}\sum_{m=1}^{M} \max_{i} \frac{\bigl|\mathrm{P}(x_m) \cap \mathcal{P}'_{c_m,i}\bigr|}{\bigl|\mathcal{P}'_{c_m,i}\bigr|}
    \end{equation}
    measures how completely the explanation recovers a part set that suffices for the classification.
    \item \emph{Preservation check (PC, completeness).} With $x'_m$ the image keeping only the parts in $\mathrm{P}(x_m)$ and $\mathds{1}[\cdot]$ the indicator function,
    \begin{equation}
        \mathrm{PC} = \frac{1}{M}\sum_{m=1}^{M} \mathds{1}\bigl[f(x'_m) = f(x_m)\bigr]
    \end{equation}
    checks that the parts deemed important preserve the prediction.
    \item \emph{Deletion check (DC, completeness).} With $x''_m$ the image with the parts in $\mathrm{P}(x_m)$ removed,
    \begin{equation}
        \mathrm{DC} = \frac{1}{M}\sum_{m=1}^{M} \mathds{1}\bigl[f(x''_m) \neq f(x_m)\bigr]
    \end{equation}
    checks that removing those parts changes the prediction.
    \item \emph{Distractibility (Dist, completeness).} With $\mathcal{P}''_{m}$ the parts whose individual removal changes the target logit by less than $5\%$,
    \begin{equation}
        \mathrm{Dist} = 1 - \frac{1}{M}\sum_{m=1}^{M} \frac{\bigl|\mathrm{P}(x_m) \cap \mathcal{P}''_{m}\bigr|}{\bigl|\mathcal{P}''_{m}\bigr|}
    \end{equation}
    penalizes attribution mass placed on parts the model demonstrably ignores.
    \item \emph{Single deletion (SD, correctness).} With $\rho_m$ the Spearman correlation between the explanation's part importances and the target-logit drops caused by deleting each part individually,
    \begin{equation}
        \mathrm{SD} = \frac{1}{2} + \frac{1}{2M}\sum_{m=1}^{M} \rho_m
    \end{equation}
    scores whether the explanation ranks parts the way the model actually reacts to their removal.
    \item \emph{Target sensitivity (TS, contrastivity).} For two alternative target classes $\hat{c}_1$ and $\hat{c}_2$ that each share two distinct parts with the image, with $\mathrm{PI}'$ and $\mathrm{PI}''$ the summed importances on the parts specific to $\hat{c}_1$ and $\hat{c}_2$ respectively,
    \begin{equation}
        \mathrm{TS} = \frac{1}{2M}\sum_{m=1}^{M} \Bigl( \mathds{1}\bigl[\mathrm{PI}'(x_m, \hat{c}_1) > \mathrm{PI}'(x_m, \hat{c}_2)\bigr] + \mathds{1}\bigl[\mathrm{PI}''(x_m, \hat{c}_1) < \mathrm{PI}''(x_m, \hat{c}_2)\bigr] \Bigr)
    \end{equation}
    checks that the explanation shifts its mass to the class-specific parts when the explained target changes.
\end{itemize}
\vspace{0.3cm}
In \Cref{tab: funnybirds}, Com.\ is the protocol's completeness aggregate, $\bigl((\mathrm{CSDC}+\mathrm{PC}+\mathrm{DC})/3+\mathrm{Dist}\bigr)/2$. ERIG leads the completeness aggregate by $3.4\%$ and contrastivity by $2.1\%$ on average across the three backbones while staying within $1.5\%$ of the strongest baseline on correctness. It places first or second in all 21 protocol cells, and the cells it does not win lie within $1.8\%$ of the best method. Under these definitions, ERIG's wins concentrate where the metrics score the \emph{placement} of attribution mass, covering the sufficient parts (CSDC) while avoiding the ignored ones (Dist). On the intervention and ranking checks (PC, DC, SD) it stays within $2\%$ of the strongest baseline, with IDGI again the strongest on PC and DC, mirroring our ImageNet results.

\subsection{\texorpdfstring{Fixed vs.\ Path-Dependent Temperature}{Fixed vs. Path-Dependent Temperature}} \label{appx: aerig}
Proposition~\ref{theorem: maxent-weights} permits a per-point temperature $\tau(t)$, while \cref{sec: max-ent} adopts a constant $\tau$ as a simplifying choice. Here, we further validate this choice empirically using a simple adaptive $\tau$ schedule, demonstrating that designing an effective adaptive $\tau$ is itself a nontrivial problem. We compare the constant $\tau=3$ against ERIG with an adaptive temperature (AERIG), in which $\tau(t)$ decreases log-linearly from $10^{10}$ to $10^{-10}$ along the path. This schedule is a deliberately simple monotone choice that directly implements the premise that gradients near the baseline are less informative. The temperature remains high near the baseline, suppressing the influence of these gradients toward uniform weights, and decreases toward the input, concentrating the weights on the more informative gradients there. \Cref{tab: aerig-fidelity,tab: aerig-localization,tab: aerig-stability,tab: aerig-noise} report the ImageNet evaluation suite from the main paper across all six backbones. The schedule's gains are scattered and small, whereas its losses are large and systematic. DEL worsens by 0.20 to 0.35 on every backbone. MS and LLC degrade by two to three orders of magnitude. LINV drops by $8.7$ to $10.6$ on five of the six backbones, and localization drops sharply for all backbones except EfficientNet. No backbone improves overall. Moreover, a monotone schedule is blind to changes in the model output, reaching its smallest $\tau$ in the saturation plateau, where concentrating the weights amplifies noise rather than signal (\cref{fig: weight-regions}). The constant $\tau$ is therefore also justified empirically. \emph{We acknowledge that this single schedule does not constitute an exhaustive exploration of adaptive temperatures.} A schedule that tracks changes in the model output would itself constitute an academic contribution, much as refinements of the Riemann approximation~\cite{swain2024riemann}, the integration path~\cite{xu2020attribution, kapishnikov2021guided}, and the baseline~\cite{erion2021improving, zhuo2024ig2, simpson2025tangentially, tachikawa2018compensated, liu2023new} have each constituted contributions to the IG family.

\begin{table}[htb]
    \centering
    \begin{minipage}[t]{0.48\textwidth}
        \centering
        \caption{\textbf{AERIG Ablation Fidelity:} The adaptive schedule worsens DEL everywhere, winning only three of eighteen cells.}
        \footnotesize
        \begin{tabular}{|c|c|c|c|c|}
            \hline
            \textbf{Model} & \textbf{Method} & \textbf{ISC $(\uparrow)$} & \textbf{AIC $(\uparrow)$} & \textbf{DEL $(\downarrow)$}\\
            \hline
            \multirow{2}{*}{ViT-Base} & ERIG & \textbf{\underline{0.6930}} & \textbf{\underline{0.9013}} & \textbf{\underline{0.4902}} \\
             & AERIG & 0.4525 & 0.7806 & 0.7405 \\
            \hline
            \multirow{2}{*}{EVA02} & ERIG & \textbf{\underline{0.5913}} & \textbf{\underline{0.8907}} & \textbf{\underline{0.5373}} \\
             & AERIG & 0.4888 & 0.8588 & 0.7786 \\
            \hline
            \multirow{2}{*}{MobileNetV4} & ERIG & \textbf{\underline{0.4118}} & \textbf{\underline{0.7234}} & \textbf{\underline{0.3626}} \\
             & AERIG & 0.3859 & 0.6501 & 0.6959 \\
            \hline
            \multirow{2}{*}{EfficientNet} & ERIG & 0.4521 & \textbf{\underline{0.8323}} & \textbf{\underline{0.4562}} \\
             & AERIG & \textbf{\underline{0.4767}} & 0.7792 & 0.6889 \\
            \hline
            \multirow{2}{*}{ResNet101} & ERIG & 0.4893 & 0.6610 & \textbf{\underline{0.3270}} \\
             & AERIG & \textbf{\underline{0.5353}} & \textbf{\underline{0.6793}} & 0.6730 \\
            \hline
            \multirow{2}{*}{DeiT3} & ERIG & \textbf{\underline{0.7320}} & \textbf{\underline{0.9010}} & \textbf{\underline{0.5561}} \\
             & AERIG & 0.6094 & 0.8641 & 0.7512 \\
            \hline
        \end{tabular}
        \label{tab: aerig-fidelity}
    \end{minipage}
    \hfill
    \begin{minipage}[t]{0.48\textwidth}
        \centering
        \caption{\textbf{AERIG Ablation Localization:} The constant temperature is better in every cell except EfficientNet IoU.}
        \footnotesize
        \begin{tabular}{|c|c|c|c|c|}
            \hline
            \textbf{Model} & \textbf{Method} & \textbf{IoU $(\uparrow)$} & \textbf{PA $(\uparrow)$} & \textbf{PF1 $(\uparrow)$}\\
            \hline
            \multirow{2}{*}{ViT-Base} & ERIG & \textbf{\underline{0.1203}} & \textbf{\underline{0.7054}} & \textbf{\underline{0.1893}} \\
             & AERIG & 0.0599 & 0.6382 & 0.0837 \\
            \hline
            \multirow{2}{*}{EVA02} & ERIG & \textbf{\underline{0.1317}} & \textbf{\underline{0.7054}} & \textbf{\underline{0.2031}} \\
             & AERIG & 0.0388 & 0.6505 & 0.0546 \\
            \hline
            \multirow{2}{*}{MobileNetV4} & ERIG & \textbf{\underline{0.1615}} & \textbf{\underline{0.6521}} & \textbf{\underline{0.2424}} \\
             & AERIG & 0.0988 & 0.5694 & 0.1340 \\
            \hline
            \multirow{2}{*}{EfficientNet} & ERIG & 0.1085 & \textbf{\underline{0.5982}} & \textbf{\underline{0.1714}} \\
             & AERIG & \textbf{\underline{0.1224}} & 0.5475 & 0.1640 \\
            \hline
            \multirow{2}{*}{ResNet101} & ERIG & \textbf{\underline{0.1617}} & \textbf{\underline{0.6580}} & \textbf{\underline{0.2419}} \\
             & AERIG & 0.0423 & 0.6128 & 0.0588 \\
            \hline
            \multirow{2}{*}{DeiT3} & ERIG & \textbf{\underline{0.1179}} & \textbf{\underline{0.6637}} & \textbf{\underline{0.1819}} \\
             & AERIG & 0.0133 & 0.6368 & 0.0180 \\
            \hline
        \end{tabular}
        \label{tab: aerig-localization}
    \end{minipage}

    \vspace{1.5em}

    \begin{minipage}[t]{0.48\textwidth}
        \centering
        \caption{\textbf{AERIG Ablation Stability:} The adaptive schedule degrades MS and LLC by two to three orders of magnitude.}
        \footnotesize
        \begin{tabular}{|c|c|c|c|c|}
            \hline
            \textbf{Model} & \textbf{Method} & \textbf{MS $(\downarrow)$} & \textbf{LLC $(\downarrow)$} & \textbf{RIS $(\downarrow)$}\\
            \hline
            \multirow{2}{*}{ViT-Base} & ERIG & $\mathbf{\underline{3.0e\text{--}04}}$ & $\mathbf{\underline{9.0e\text{--}06}}$ & \textbf{\underline{4.9460}} \\
             & AERIG & 0.0255 & $6.6e\text{--}04$ & 50.070 \\
            \hline
            \multirow{2}{*}{EVA02} & ERIG & $\mathbf{\underline{6.7e\text{--}05}}$ & $\mathbf{\underline{1.5e\text{--}06}}$ & \textbf{\underline{1.6010}} \\
             & AERIG & 0.0187 & $4.2e\text{--}04$ & 66.966 \\
            \hline
            \multirow{2}{*}{MobileNetV4} & ERIG & $\mathbf{\underline{1.0e\text{--}04}}$ & $\mathbf{\underline{5.4e\text{--}06}}$ & \textbf{\underline{2.1960}} \\
             & AERIG & 0.0473 & 0.0018 & 72.558 \\
            \hline
            \multirow{2}{*}{EfficientNet} & ERIG & $\mathbf{\underline{6.7e\text{--}05}}$ & $\mathbf{\underline{2.1e\text{--}06}}$ & \textbf{\underline{1.2710}} \\
             & AERIG & 0.0265 & $8.3e\text{--}04$ & 62.522 \\
            \hline
            \multirow{2}{*}{ResNet101} & ERIG & $\mathbf{\underline{8.6e\text{--}05}}$ & $\mathbf{\underline{3.8e\text{--}06}}$ & \textbf{\underline{1.0660}} \\
             & AERIG & 0.0621 & 0.0028 & 104.7 \\
            \hline
            \multirow{2}{*}{DeiT3} & ERIG & $\mathbf{\underline{3.9e\text{--}05}}$ & $\mathbf{\underline{1.0e\text{--}06}}$ & \textbf{\underline{0.7560}} \\
             & AERIG & 0.0157 & $4.1e\text{--}04$ & 83.403 \\
            \hline
        \end{tabular}
        \label{tab: aerig-stability}
    \end{minipage}
    \hfill
    \begin{minipage}[t]{0.48\textwidth}
        \centering
        \caption{\textbf{AERIG Ablation Noise:} The adaptive schedule loses $8.7$ to $10.6$ LINV on five of six backbones.}
        \footnotesize
        \begin{tabular}{|c|c|c|c|}
            \hline
            \textbf{Model} & \textbf{Method} & \textbf{LINV-C $(\uparrow)$} & \textbf{LINV-P $(\uparrow)$}\\
            \hline
            \multirow{2}{*}{ViT-Base} & ERIG & \textbf{\underline{19.343}} & \textbf{\underline{19.201}} \\
             & AERIG & 10.390 & 10.382 \\
            \hline
            \multirow{2}{*}{EVA02} & ERIG & \textbf{\underline{20.128}} & \textbf{\underline{20.115}} \\
             & AERIG & 10.552 & 10.572 \\
            \hline
            \multirow{2}{*}{MobileNetV4} & ERIG & \textbf{\underline{18.130}} & \textbf{\underline{18.113}} \\
             & AERIG & 9.3252 & 9.3624 \\
            \hline
            \multirow{2}{*}{EfficientNet} & ERIG & \textbf{\underline{20.705}} & \textbf{\underline{20.712}} \\
             & AERIG & 10.061 & 10.272 \\
            \hline
            \multirow{2}{*}{ResNet101} & ERIG & 16.807 & 16.778 \\
             & AERIG & \textbf{\underline{19.772}} & \textbf{\underline{19.746}} \\
            \hline
            \multirow{2}{*}{DeiT3} & ERIG & \textbf{\underline{19.578}} & \textbf{\underline{19.584}} \\
             & AERIG & 10.795 & 10.932 \\
            \hline
        \end{tabular}
        \label{tab: aerig-noise}
    \end{minipage}
\end{table}

\clearpage
\subsection{\texorpdfstring{Baseline Sensitivity}{Baseline Sensitivity}} \label{appx: baselines}
\begin{table}
    \centering
    \begin{minipage}[t]{0.48\textwidth}
        \centering
        \caption{\textbf{Baseline Ablation Fidelity:} The constant baselines track the zero default within $0.7$--$4.0\%$.}
        \footnotesize
        \begin{tabular}{|c|c|c|c|c|}
            \hline
            \textbf{Model} & \textbf{Baseline} & \textbf{ISC $(\uparrow)$} & \textbf{AIC $(\uparrow)$} & \textbf{DEL $(\downarrow)$}\\
            \hline
            \multirow{12}{*}{EVA-02} & zero & 0.5913 & 0.8907 & 0.5373 \\
            \cline{2-5}
             & ones & 0.5872 & 0.8923 & \textbf{\underline{0.5368}} \\
             & random color & 0.5893 & 0.8972 & 0.5374 \\
             & channel mean & 0.5906 & 0.8964 & \textbf{0.5369} \\
             & channel median & 0.5901 & 0.8961 & 0.5381 \\
            \cline{2-5}
             & uniform & \textbf{\underline{0.6178}} & 0.9020 & 0.5491 \\
             & Gaussian & \textbf{0.6118} & 0.9042 & 0.5577 \\
            \cline{2-5}
             & blur & 0.5798 & \textbf{0.9110} & 0.5638 \\
            \cline{2-5}
             & pixel perm. & 0.5984 & 0.8962 & 0.5487 \\
             & window perm. & 0.5490 & \textbf{\underline{0.9128}} & 0.6587 \\
             & magnitude scr. & 0.6004 & 0.9013 & 0.5482 \\
             & phase scr. & 0.6031 & 0.9085 & 0.5568 \\
            \hline
            \multirow{12}{*}{ResNet101} & zero & 0.4893 & 0.6610 & \textbf{0.3270} \\
            \cline{2-5}
             & ones & \textbf{\underline{0.4917}} & \textbf{\underline{0.6881}} & \textbf{\underline{0.3254}} \\
             & random color & 0.4817 & 0.6778 & 0.3276 \\
             & channel mean & \textbf{0.4899} & \textbf{0.6823} & 0.3275 \\
             & channel median & 0.4886 & 0.6819 & 0.3272 \\
            \cline{2-5}
             & uniform & 0.4552 & 0.6634 & 0.3293 \\
             & Gaussian & 0.4565 & 0.6550 & 0.3304 \\
            \cline{2-5}
             & blur & 0.4775 & 0.6715 & 0.3373 \\
            \cline{2-5}
             & pixel perm. & 0.4470 & 0.6527 & 0.3273 \\
             & window perm. & 0.4245 & 0.6605 & 0.3759 \\
             & magnitude scr. & 0.4446 & 0.6586 & 0.3285 \\
             & phase scr. & 0.4412 & 0.6449 & 0.3412 \\
            \hline
        \end{tabular}
        \label{tab: baseline-fidelity}
    \end{minipage}
    \hfill
    \begin{minipage}[t]{0.48\textwidth}
        \centering
        \caption{\textbf{Baseline Ablation Localization:} Constant baselines stay within $8.4\%$ on IoU and PF1; the rest are backbone-specific.}
        \footnotesize
        \begin{tabular}{|c|c|c|c|c|}
            \hline
            \textbf{Model} & \textbf{Baseline} & \textbf{IoU $(\uparrow)$} & \textbf{PA $(\uparrow)$} & \textbf{PF1 $(\uparrow)$}\\
            \hline
            \multirow{12}{*}{EVA-02} & zero & \textbf{\underline{0.1317}} & \textbf{0.7054} & \textbf{\underline{0.2031}} \\
            \cline{2-5}
             & ones & 0.1212 & 0.7046 & 0.1879 \\
             & random color & \textbf{0.1245} & \textbf{\underline{0.7060}} & \textbf{0.1925} \\
             & channel mean & 0.1228 & 0.7045 & 0.1897 \\
             & channel median & 0.1230 & 0.7053 & 0.1902 \\
            \cline{2-5}
             & uniform & 0.0906 & 0.7032 & 0.1395 \\
             & Gaussian & 0.0319 & 0.6965 & 0.0507 \\
            \cline{2-5}
             & blur & 0.0973 & 0.7000 & 0.1516 \\
            \cline{2-5}
             & pixel perm. & 0.0523 & 0.6987 & 0.0814 \\
             & window perm. & 0.0270 & 0.6949 & 0.0428 \\
             & magnitude scr. & 0.0416 & 0.6978 & 0.0660 \\
             & phase scr. & 0.0576 & 0.6998 & 0.0905 \\
            \hline
            \multirow{12}{*}{ResNet101} & zero & 0.1617 & \textbf{\underline{0.6580}} & 0.2419 \\
            \cline{2-5}
             & ones & 0.1568 & 0.6561 & 0.2361 \\
             & random color & 0.1559 & 0.6560 & 0.2347 \\
             & channel mean & 0.1556 & 0.6563 & 0.2346 \\
             & channel median & 0.1557 & 0.6564 & 0.2347 \\
            \cline{2-5}
             & uniform & \textbf{\underline{0.1660}} & \textbf{0.6577} & \textbf{\underline{0.2471}} \\
             & Gaussian & 0.1581 & 0.6550 & 0.2363 \\
            \cline{2-5}
             & blur & 0.1539 & 0.6562 & 0.2325 \\
            \cline{2-5}
             & pixel perm. & \textbf{0.1642} & 0.6573 & \textbf{0.2436} \\
             & window perm. & 0.1369 & 0.6435 & 0.2081 \\
             & magnitude scr. & 0.1628 & 0.6557 & 0.2422 \\
             & phase scr. & 0.1611 & 0.6556 & 0.2410 \\
            \hline
        \end{tabular}
        \label{tab: baseline-localization}
    \end{minipage}

    \vspace{1.5em}

    \begin{minipage}[t]{0.48\textwidth}
        \centering
        \caption{\textbf{Baseline Ablation Stability:} Effects are backbone-specific rather than systematic across the twelve baselines.}
        \footnotesize
        \setlength{\tabcolsep}{4pt}
        \begin{tabular}{|c|c|c|c|c|}
            \hline
            \textbf{Model} & \textbf{Baseline} & \textbf{MS $(\downarrow)$} & \textbf{LLC $(\downarrow)$} & \textbf{RIS $(\downarrow)$}\\
            \hline
            \multirow{12}{*}{EVA-02} & zero & $6.7e\text{--}05$ & $1.5e\text{--}06$ & 1.601 \\
            \cline{2-5}
             & ones & $\mathbf{6.2e\text{\textbf{--}}05}$ & $\mathbf{1.4e\text{\textbf{--}}06}$ & \textbf{\underline{1.410}} \\
             & random color & $6.4e\text{--}05$ & $1.4e\text{--}06$ & 1.465 \\
             & channel mean & $6.2e\text{--}05$ & $1.4e\text{--}06$ & 1.413 \\
             & channel median & $\mathbf{\underline{6.2e\text{\textbf{--}}05}}$ & $\mathbf{\underline{1.4e\text{\textbf{--}}06}}$ & 1.411 \\
            \cline{2-5}
             & uniform & $6.4e\text{--}05$ & $1.4e\text{--}06$ & 1.445 \\
             & Gaussian & $6.2e\text{--}05$ & $1.4e\text{--}06$ & 1.410 \\
            \cline{2-5}
             & blur & $6.2e\text{--}05$ & $1.4e\text{--}06$ & 1.439 \\
            \cline{2-5}
             & pixel perm. & $6.2e\text{--}05$ & $1.4e\text{--}06$ & \textbf{1.410} \\
             & window perm. & $2.0e\text{--}04$ & $5.2e\text{--}06$ & 47.530 \\
             & magnitude scr. & $1.0e\text{--}04$ & $2.2e\text{--}06$ & 29.804 \\
             & phase scr. & $6.2e\text{--}05$ & $1.4e\text{--}06$ & 1.414 \\
            \hline
            \multirow{12}{*}{ResNet101} & zero & $\mathbf{\underline{8.6e\text{--}05}}$ & $\mathbf{\underline{3.8e\text{--}06}}$ & \textbf{\underline{1.066}} \\
            \cline{2-5}
             & ones & $2.0e\text{--}04$ & $1.0e\text{--}05$ & 2.599 \\
             & random color & $1.1e\text{--}02$ & $5.0e\text{--}04$ & 89.584 \\
             & channel mean & $2.0e\text{--}04$ & $9.7e\text{--}06$ & 2.741 \\
             & channel median & $2.0e\text{--}04$ & $8.0e\text{--}06$ & 2.340 \\
            \cline{2-5}
             & uniform & $\mathbf{1.0e\text{\textbf{--}}04}$ & $6.4e\text{--}06$ & 2.046 \\
             & Gaussian & $1.0e\text{--}04$ & $\mathbf{6.3e\text{\textbf{--}}06}$ & \textbf{2.036} \\
            \cline{2-5}
             & blur & $2.0e\text{--}04$ & $9.4e\text{--}06$ & 2.641 \\
            \cline{2-5}
             & pixel perm. & $1.0e\text{--}04$ & $6.3e\text{--}06$ & 2.040 \\
             & window perm. & $3.0e\text{--}04$ & $1.5e\text{--}05$ & 49.918 \\
             & magnitude scr. & $2.0e\text{--}04$ & $8.3e\text{--}06$ & 2.670 \\
             & phase scr. & $1.0e\text{--}04$ & $6.4e\text{--}06$ & 2.048 \\
            \hline
        \end{tabular}
        \label{tab: baseline-stability}
    \end{minipage}
    \hfill
    \begin{minipage}[t]{0.48\textwidth}
        \centering
        \caption{\textbf{Baseline Ablation Noise:} LINV stays within $4.5$ units of the zero default across all twelve baselines.}
        \footnotesize
        \begin{tabular}{|c|c|c|c|}
            \hline
            \textbf{Model} & \textbf{Baseline} & \textbf{LINV-C $(\uparrow)$} & \textbf{LINV-P $(\uparrow)$}\\
            \hline
            \multirow{12}{*}{EVA-02} & zero & 20.1285 & 20.1150 \\
            \cline{2-4}
             & ones & 20.0902 & 20.0854 \\
             & random color & 20.3127 & 20.2931 \\
             & channel mean & 20.2000 & 20.1827 \\
             & channel median & 20.2892 & 20.2729 \\
            \cline{2-4}
             & uniform & 22.3942 & 22.2114 \\
             & Gaussian & \textbf{23.8521} & 23.6896 \\
            \cline{2-4}
             & blur & 21.4313 & 21.3799 \\
            \cline{2-4}
             & pixel perm. & 23.6319 & 23.6009 \\
             & window perm. & \textbf{\underline{24.6156}} & \textbf{\underline{24.4948}} \\
             & magnitude scr. & 23.7934 & \textbf{23.7436} \\
             & phase scr. & 22.9985 & 22.9185 \\
            \hline
            \multirow{12}{*}{ResNet101} & zero & 16.8069 & 16.7781 \\
            \cline{2-4}
             & ones & 16.2361 & 16.2020 \\
             & random color & 16.3634 & 16.3341 \\
             & channel mean & 16.2264 & 16.1936 \\
             & channel median & 16.2324 & 16.2009 \\
            \cline{2-4}
             & uniform & 18.5193 & 18.4302 \\
             & Gaussian & \textbf{19.5956} & \textbf{19.5321} \\
            \cline{2-4}
             & blur & 16.7069 & 16.6935 \\
            \cline{2-4}
             & pixel perm. & 19.3372 & 19.2665 \\
             & window perm. & 18.5602 & 18.5158 \\
             & magnitude scr. & \textbf{\underline{19.6389}} & \textbf{\underline{19.5709}} \\
             & phase scr. & 18.3164 & 18.3061 \\
            \hline
        \end{tabular}
        \label{tab: baseline-noise}
    \end{minipage}
\end{table}

As can be seen from \Cref{tab: baseline-fidelity} to \Cref{tab: baseline-noise}, only one pattern is consistent across both backbones: the constant references closely track the zero default. ISC, AIC, DEL, and PA remain within $0.7\%$ of the family mean on EVA-02 and within $4.0\%$ on ResNet101, while IoU and PF1 remain within $8.4\%$ on both backbones, indicating that ERIG's fidelity and localization do not depend critically on the black baseline. Beyond this, the effects of the baseline are backbone-specific rather than systematic, and the same family can exhibit opposite trends across backbones. Uniform noise increases ISC by $4.5\%$ while reducing IoU by $31\%$ on EVA-02, yet decreases ISC by $7\%$ and achieves the best IoU on ResNet101. Standard normal noise leaves localization on ResNet101 largely unchanged but reduces EVA-02's IoU by three quarters. The per-image random color baseline is indistinguishable from the other constant baselines on EVA-02, yet degrades stability on ResNet101 by approximately two orders of magnitude. The structure-destroying replacements substantially impair localization on EVA-02, with IoU decreasing by $56$ to $80\%$, whereas ResNet101 experiences a decrease of at most $15\%$. Even within the same family, the window permutation increases RIS thirtyfold on EVA-02, while the phase scramble produces results comparable to the zero default. Baseline choice therefore has model-specific effects that are not captured by family-level summaries. The consistent finding is that constant references closely reproduce the behavior of the default baseline.

\subsection{\texorpdfstring{Validation of the Noise Metrics}{Validation of the Noise Metrics}} \label{appx: linv-validation}
To relate LINV to existing complexity, compressibility, and sparsity measures, we evaluate five such measures alongside LINV on the ImageNet evaluation of the main paper (six methods on six backbones). Each of the new metrics quantifies how strongly attribution mass concentrates, so we refer to them collectively as \emph{concentration measures}. Since these measures originate in digital-image analysis, they are computed on the 8-bit grayscale rendering $\bm{u} \in \{0,\dots,255\}^{N}$ of the channel-summed attribution map, except the Gini coefficient, which uses the attribution visualization $\bm{v}$.
\begin{itemize}
    \item \emph{Histogram entropy.} $H_{\mathrm{hist}} = -\sum_{\ell=0}^{255} q_\ell \log_2 q_\ell \in [0,8]$ bits, with $q_\ell$ the fraction of pixels of $\bm{u}$ at gray level $\ell$.
    \item \emph{JPEG compressibility.} JPEG$_{\mathrm{ratio}}$, the encoded-to-raw byte ratio of $\bm{u}$ at quality 90 (a practical stand-in for Kolmogorov complexity), and JPEG$_{\mathrm{RMSE}}$, the root-mean-square error between $\bm{u}$ and its JPEG reconstruction.
    \item \emph{Gini sparsity}~\cite{chalasani2020concise}\emph{.} $\mathrm{GS} = \sum_{i=1}^{N} (2i - N - 1)\, v_{(i)} \big/ \bigl(N \sum_{i} v_{(i)}\bigr) \in [0,1)$ over the sorted entries $v_{(1)} \le \dots \le v_{(N)}$ of $\bm{v}$; higher means attribution mass concentrated on fewer features.
    \item \emph{Entropy complexity}~\cite{bhatt2020evaluating}\emph{.} $H = -\sum_{i=1}^{N} p_i \ln p_i \big/ \ln N \in [0,1]$ with $p_i = u_i / \sum_j u_j$, the normalized Shannon entropy of the fractional-contribution distribution; a value of $1$ means maximally spread.
\end{itemize}

\begin{table}
    \centering
    \caption{\textbf{LINV vs.\ Concentration:} Compared measures disagree not only with LINV but also among themselves on the overall order.}
    \footnotesize
    \begin{tabular}{|c|c|c|c|c|c|c|c|c|}
        \hline
        \multicolumn{2}{|c|}{} & \multicolumn{2}{c|}{\textbf{Noise}} & \multicolumn{4}{c|}{\textbf{Complexity / Compressibility}} & \multicolumn{1}{c|}{\textbf{Sparsity}}\\
        \hline
        \textbf{Model} & \textbf{Method} & \textbf{LINV-C} & \textbf{LINV-P} & $\bm{H}$ & $\bm{H_{\mathrm{hist}}}$ & \textbf{JPEG$_{\mathrm{ratio}}$} & \textbf{JPEG$_{\mathrm{RMSE}}$} & $\bm{\mathrm{GS}}$\\
        \hline
        \multirow{6}{*}{ResNet101} & IG & 6.587 & 6.489 & 0.9976 & 5.937 & 0.3957 & 3.345 & 0.6989 \\
         & IDGI & 9.431 & 9.402 & \textbf{0.8910} & \textbf{4.528} & \textbf{0.2790} & \textbf{2.673} & \textbf{\underline{0.8346}} \\
         & BIG & 6.649 & 6.605 & 0.9978 & 5.674 & 0.3868 & 3.325 & 0.7304 \\
         & GIG & 7.965 & 7.909 & 0.9980 & 4.600 & 0.3272 & 2.998 & \textbf{0.8036} \\
         & SmoothGrad & \textbf{10.135} & \textbf{9.918} & 0.9968 & 6.675 & 0.4925 & 3.705 & 0.6000 \\
         & ERIG (ours) & \textbf{\underline{16.807}} & \textbf{\underline{16.778}} & \textbf{\underline{0.8748}} & \textbf{\underline{4.229}} & \textbf{\underline{0.2673}} & \textbf{\underline{2.573}} & 0.7467 \\
        \hline
        \multirow{6}{*}{EfficientNet} & IG & 6.877 & 6.857 & 0.9975 & 6.318 & 0.4333 & 3.532 & 0.6552 \\
         & IDGI & \textbf{10.243} & \textbf{10.246} & \textbf{0.9208} & 4.803 & 0.3222 & \textbf{2.968} & \textbf{\underline{0.8122}} \\
         & BIG & 7.917 & 8.054 & 0.9977 & 6.008 & 0.3807 & 3.289 & 0.6974 \\
         & GIG & 7.364 & 7.232 & 0.9981 & \textbf{4.643} & \textbf{\underline{0.3197}} & 2.983 & \textbf{0.7984} \\
         & SmoothGrad & 8.729 & 8.710 & 0.9972 & 6.605 & 0.4679 & 3.705 & 0.6151 \\
         & ERIG (ours) & \textbf{\underline{20.705}} & \textbf{\underline{20.712}} & \textbf{\underline{0.8959}} & \textbf{\underline{4.458}} & \textbf{0.3212} & \textbf{\underline{2.867}} & 0.7315 \\
        \hline
        \multirow{6}{*}{MobileNetV4} & IG & 5.971 & 5.793 & 0.9972 & 6.406 & 0.4620 & 3.648 & 0.6490 \\
         & IDGI & \textbf{10.086} & \textbf{10.073} & \textbf{0.9389} & 5.071 & \textbf{0.3306} & \textbf{3.015} & \textbf{0.7940} \\
         & BIG & 6.506 & 6.476 & 0.9977 & 5.958 & 0.4035 & 3.432 & 0.7078 \\
         & GIG & 6.598 & 6.447 & 0.9979 & \textbf{4.683} & 0.3339 & 3.052 & \textbf{\underline{0.7980}} \\
         & SmoothGrad & 7.848 & 7.651 & 0.9965 & 6.837 & 0.5243 & 3.802 & 0.5814 \\
         & ERIG (ours) & \textbf{\underline{18.130}} & \textbf{\underline{18.113}} & \textbf{\underline{0.8959}} & \textbf{\underline{4.632}} & \textbf{\underline{0.3088}} & \textbf{\underline{2.795}} & 0.7132 \\
        \hline
        \multirow{6}{*}{ViT-Base} & IG & 7.222 & 7.061 & 0.9979 & 6.014 & 0.3900 & 3.333 & 0.7004 \\
         & IDGI & \textbf{10.162} & \textbf{10.008} & \textbf{0.9207} & \textbf{3.087} & \textbf{0.1503} & \textbf{1.859} & \textbf{\underline{0.9296}} \\
         & BIG & 8.036 & 7.945 & 0.9979 & 5.945 & 0.3830 & 3.252 & 0.6968 \\
         & GIG & 8.005 & 7.918 & 0.9982 & 4.546 & 0.3023 & 2.889 & 0.8188 \\
         & SmoothGrad & 9.260 & 8.876 & 0.9976 & 6.352 & 0.4185 & 3.407 & 0.6483 \\
         & ERIG (ours) & \textbf{\underline{19.343}} & \textbf{\underline{19.201}} & \textbf{\underline{0.8234}} & \textbf{\underline{2.491}} & \textbf{\underline{0.1287}} & \textbf{\underline{1.648}} & \textbf{0.8412} \\
        \hline
        \multirow{6}{*}{EVA-02} & IG & 7.568 & 7.523 & 0.9978 & 6.291 & 0.4164 & 3.397 & 0.6499 \\
         & IDGI & \textbf{10.803} & \textbf{10.778} & \textbf{0.9404} & \textbf{3.967} & \textbf{0.2159} & \textbf{2.270} & \textbf{\underline{0.8855}} \\
         & BIG & 8.467 & 8.422 & 0.9981 & 5.916 & 0.3695 & 3.213 & 0.7029 \\
         & GIG & 8.505 & 8.332 & 0.9982 & 4.699 & 0.3209 & 2.956 & 0.7917 \\
         & SmoothGrad & 9.595 & 9.127 & 0.9972 & 6.737 & 0.4866 & 3.654 & 0.5784 \\
         & ERIG (ours) & \textbf{\underline{20.128}} & \textbf{\underline{20.115}} & \textbf{\underline{0.8642}} & \textbf{\underline{3.346}} & \textbf{\underline{0.1910}} & \textbf{\underline{2.012}} & \textbf{0.7977} \\
        \hline
        \multirow{6}{*}{DeiT3} & IG & 7.158 & 7.570 & 0.9982 & 5.673 & 0.3272 & 2.997 & 0.7478 \\
         & IDGI & \textbf{10.247} & \textbf{10.254} & \textbf{0.9177} & \textbf{3.439} & \textbf{0.1907} & \textbf{2.053} & \textbf{\underline{0.9131}} \\
         & BIG & 7.747 & 8.102 & 0.9981 & 5.621 & 0.3124 & 2.772 & 0.7460 \\
         & GIG & 8.658 & 8.721 & 0.9984 & 4.458 & 0.2891 & 2.862 & \textbf{0.8273} \\
         & SmoothGrad & 9.592 & 9.361 & 0.9982 & 5.971 & 0.3643 & 3.300 & 0.6971 \\
         & ERIG (ours) & \textbf{\underline{19.578}} & \textbf{\underline{19.584}} & \textbf{\underline{0.8523}} & \textbf{\underline{3.011}} & \textbf{\underline{0.1812}} & \textbf{\underline{1.916}} & 0.8177 \\
        \hline
    \end{tabular}
    \label{tab: linv-proxies}
\end{table}

\begin{table}
    \centering
    \caption{\textbf{LINV Correlations:} LINV correlates at $+0.91$ to $+0.93$ with the stability metrics but at most $+0.66$ with any other compared measure. Even with ERIG excluded, the stability correlations stay at $+0.86$ to $+0.89$ while the rest collapse further.}
    \footnotesize
    \begin{tabular}{|c|c|c|c|c|c|c|c|c|c|}
        \hline
        \textbf{Cells} & \textbf{Metric} & $\bm{H}$ & $\bm{H_{\mathrm{hist}}}$ & \textbf{JPEG$_{\mathrm{ratio}}$} & \textbf{JPEG$_{\mathrm{RMSE}}$} & $\bm{\mathrm{GS}}$ & \textbf{MS} & \textbf{LLC} & \textbf{RIS}\\
        \hline
        \multirow{2}{*}{All 36 cells} & LINV-C & $+0.66$ & $+0.55$ & $+0.57$ & $+0.60$ & $+0.36$ & $+0.91$ & $+0.92$ & $+0.52$ \\
         & LINV-P & $+0.65$ & $+0.56$ & $+0.60$ & $+0.63$ & $+0.38$ & $+0.92$ & $+0.93$ & $+0.55$ \\
        \hline
        \multirow{2}{*}{Without ERIG} & LINV-C & $+0.43$ & $+0.28$ & $+0.36$ & $+0.38$ & $+0.27$ & $+0.84$ & $+0.86$ & $+0.17$ \\
         & LINV-P & $+0.42$ & $+0.31$ & $+0.41$ & $+0.42$ & $+0.30$ & $+0.87$ & $+0.89$ & $+0.23$ \\
        \hline
    \end{tabular}
    \label{tab: linv-corr}
\end{table}

Two observations follow from \cref{tab: linv-proxies}. First, the concentration measures disagree with each other. By the four measures computed on $\bm{u}$, ERIG produces the most concentrated and most compressible maps, with the lowest $H$, $H_{\mathrm{hist}}$, and JPEG$_{\mathrm{RMSE}}$ on all six backbones, while the Gini of the visualization puts IDGI first on five of six backbones with ERIG in mid-field. IDGI holds its mass in a few isolated, extremely large values, which the Gini coefficient rewards, yet around those spikes its maps still flicker with fine-grained variation, so as images they remain busy; ERIG's maps have no towering pixels but a genuinely quiet background, which the entropy and compressibility measures reward. Second, no concentration measure orders the methods the way noise does. SmoothGrad is the most diffuse baseline according to histogram entropy and both JPEG measures, hence the least concentrated and, by those measures' logic, the noisiest, yet cleaner than IG, BIG, and GIG by LINV on every backbone; GIG is more concentrated than SmoothGrad by those proxies yet noisier by LINV on every backbone; and IDGI and ERIG nearly coincide on the entropy and JPEG measures while LINV nearly doubles between them.

Concentration can nevertheless serve as a footprint of noise removal. Small-amplitude components of gradient irregularities can propagate into the attribution with broad spatial support, assigning nonzero relevance to pixels that provide little or no evidence for the prediction. This broad support spreads the attribution values across gray levels, increasing both histogram and contribution entropies, introducing high-frequency content that is less amenable to compression, and shifting attribution mass away from relevant features. Suppressing these components has the opposite effect, reducing variation in the background and concentrating the attribution mass around the evidence. ERIG's entropy-regularized weighting leaves exactly this footprint, the lowest entropies and the most compressible maps in \cref{tab: linv-proxies}, and its localization confirms the reading; with the best IoU and PF1 on all six backbones (\cref{tab: localization-results}), the mass that survives its weighting sits on the ground-truth object regions, so its concentration comes from removing off-object, noise-like relevance rather than from discarding evidence.

Concentration can, however, arise without noise removal. Because concentration measures are functionals of the value distribution alone, they provide no information about the spatial location of the attribution mass. An attribution map concentrated on the object and one concentrated on a small number of noise pixels can therefore exhibit similar concentration statistics, with their distinction captured instead by localization metrics. IDGI illustrates this limitation. It ranks second to ERIG according to $H$ on every backbone and first according to $\mathrm{GS}$ on five of six backbones, yet exhibits the weakest localization performance throughout. Its attribution mass is concentrated in a small number of high-amplitude gradient irregularities, which its direct squared-gradient coupling passes through without attenuation. Moreover, concentration measures can conflate noise with the spatial distribution of the underlying evidence by mistaking structure for unwanted variation, making their absolute values insufficient for assessing noise removal. Concentration is therefore best interpreted as a potential byproduct of denoising rather than as a direct measure of it.

The correlations quantify this lack of correspondence. \Cref{tab: linv-corr} reports Spearman rank correlations between LINV, the five concentration measures, and the stability metrics from \cref{tab: stability-results}. We compute correlations across all 36 model-method cells and separately across the 30 cells excluding ERIG, thereby controlling for the possibility that a single method drives the observed correlations. Each measure is evaluated using its nominal orientation, with lower values receiving better ranks for $H$, $H_{\mathrm{hist}}$, JPEG$_{\mathrm{ratio}}$, JPEG$_{\mathrm{RMSE}}$, MS, LLC, and RIS, and higher values receiving better ranks for $\mathrm{GS}$. Thus, a positive coefficient indicates agreement with LINV's ordering in which higher values correspond to cleaner attributions. For the concentration measures, however, this orientation is a ranking convention rather than a claim about their relationship to attribution quality. LINV exhibits strong agreement with the stability metrics associated with gradient noise and irregularities. Its Spearman correlation with MS and LLC ranges from $+0.91$ to $+0.93$ and remains between $+0.84$ and $+0.89$ when ERIG is excluded, while its agreement with RIS is weaker, consistent with RIS being the only scale-normalized metric among the three which LINV is not. In contrast, LINV exhibits at most moderate agreement with any concentration measure ($\rho \le 0.66$), which decreases to $\rho \le 0.43$ when ERIG is excluded. Empirically, LINV therefore behaves more consistently with the stability family than with the concentration family.

None of the concentration measures directly quantifies noise. $H$, $H_{\mathrm{hist}}$, and $\mathrm{GS}$ are functionals of the value distribution alone. Spatially scrambling an attribution map, thereby transforming its structure into salt-and-pepper noise, leaves these measures exactly unchanged, whereas LINV, which is spatial by construction, identifies the scrambled map as highly noisy through its Laplacian-like convolution ($A * S$) and patch statistics. The JPEG measures quantify overall fine-scale content and cannot distinguish unstructured fluctuations from legitimate fine-grained evidence. For example, an attribution map with sharp object boundaries may compress less effectively than a blurred map without necessarily being noisier. In contrast, LINV-C and LINV-P target the quantity of interest directly, namely the variance of the spatially unstructured component, using the Laplacian-like convolution of~\cite{immerkaer1996fast} and the patch-PCA eigenvalue tail of~\cite{chen2015efficient}, respectively. The concentration measures therefore serve as secondary descriptive statistics, whereas LINV directly quantifies the noise component of attribution evaluation. The two LINV variants also exhibit close agreement across all backbones (\cref{tab: noise-results}), indicating that either variant can serve as a standalone noise metric.

\subsection{\texorpdfstring{Adversarial Stability}{Adversarial Stability}} \label{appx: astability}
The uniform stability protocol of \cref{subsec: stability} samples isotropic perturbations, and no finite Monte Carlo sample can cover every direction of a high-dimensional input space. We therefore complement it with adversarially selected perturbation directions, computed by projected gradient ascent in the spirit of the attribution attacks of Ghorbani et al.~\cite{ghorbani2019interpretation} and Dombrowski et al.~\cite{dombrowski2019explanations}. At the same perturbation radius $r=0.1$ as the uniform protocol, the attack maximizes the dissimilarity between the original and perturbed attribution maps while keeping the prediction unchanged; the constraint is enforced by a margin penalty during the ascent and by a final feasibility check afterwards. With $\mathbf{e}_{\bm{x}}$ the attribution map at input $\bm{x}$, $f$ the model output, $y$ the predicted class, and $[z]_+ = \max(z,0)$, each restart ascends
\begin{equation}
\mathcal{J}(\bm{\delta}) = 1 - \frac{\bigl\langle \mathbf{e}_{\bm{x}+\bm{\delta}},\, \mathbf{e}_{\bm{x}} \bigr\rangle}{\lVert \mathbf{e}_{\bm{x}+\bm{\delta}} \rVert_2 \, \lVert \mathbf{e}_{\bm{x}} \rVert_2} - \lambda \Bigl[ \max_{c \neq y} f_c(\bm{x}+\bm{\delta}) - f_y(\bm{x}+\bm{\delta}) + m \Bigr]_+
\end{equation}
with sign-gradient steps under a cosine-decayed step size $\eta_t$,
\begin{equation}
\bm{\delta}_{t+1} = \Pi_{r}\Bigl( \bm{\delta}_t + \eta_t \, \mathrm{sign}\bigl( \nabla_{\bm{\delta}} \mathcal{J}(\bm{\delta}_t) \bigr) \Bigr), \qquad \eta_t = \eta_0\, r \Bigl( 0.55 + 0.45 \cos \tfrac{\pi t}{T} \Bigr),
\end{equation}
where $\Pi_r$ projects back onto the radius-$r$ perturbation set. The defaults are $\lambda = 1$ and $m = 0.0$, with $m$ the logit margin the perturbation must maintain between the predicted class's logit and the next largest. The ascent is run for $T=100$ steps from each of $R=3$ independent random restarts per image; all three are drawn with the uniform protocol's perturbation law, the first from a smaller ball of radius $0.01r$ so that the ascent also starts close to the input, the other two at the full radius $r$. Instead of relying on a single ascent, the restarts therefore probe the non-convex objective from starting points near the unperturbed input as well as across the full perturbation set. The ascent needs second-order derivative information, since the attribution is built from input gradients and differentiating it with respect to $\bm{\delta}$ passes through second derivatives of $f$. ReLU activations have a second derivative that is not only undefined at their kinks, but also zero away from them. This degrades the optimization signal and can make the ascent direction vanish. Following~\cite{dombrowski2019explanations}, ReLU backbones are therefore softplus-smoothed during the attack optimization only, restoring nonzero curvature in the spirit of the smooth surrogates of Appendix~\ref{appx: relu-scope}. Every reported quantity, however, is evaluated on the original network at full path resolution, so that the smoothing cannot affect any metric computation.

We report the adversarial counterparts of the three stability metrics of \cref{tab: stability-results}, Adversarial Sensitivity (AS), Adversarial Lipschitz Constant (ALC), and Adversarial Input Stability (AIS). They are the same functionals as MS, LLC, and RIS, with the worst case taken over adversarially chosen directions instead of uniform draws. Let $\mathcal{A}(\bm{x}) = \bigl\{ \bm{x} + \bm{\delta}^{(k)} \bigr\}_{k=1}^{R}$ collect the perturbed inputs the attack retains for image $\bm{x}$, i.e.\ the endpoint $\bm{\delta}^{(k)}$ of each of the $R$ restarts after the feasibility backtracking that shrinks any perturbation still changing the predicted class. Then
\begin{align}
\operatorname{AS}(\bm{x},r) &= \max_{\bm{x}' \in \mathcal{A}(\bm{x})}
    \lVert \mathbf{e}_{\bm{x}} - \mathbf{e}_{\bm{x}'} \rVert_2 \label{eq:as} \\
\operatorname{ALC}(\bm{x},r) &= \max_{\bm{x}' \in \mathcal{A}(\bm{x})}
    \frac{\lVert \mathbf{e}_{\bm{x}} - \mathbf{e}_{\bm{x}'} \rVert_2}
         {\lVert \bm{x} - \bm{x}' \rVert_2} \label{eq:alc} \\
\operatorname{AIS}(\bm{x},r) &= \max_{\bm{x}' \in \mathcal{A}(\bm{x})}
    \frac{\lVert \mathbf{e}_{\bm{x}} - \mathbf{e}_{\bm{x}'} \rVert_2 / \lVert \mathbf{e}_{\bm{x}} \rVert_2}
         {\lVert \bm{x} - \bm{x}' \rVert_2 / \lVert \bm{x} \rVert_2} \label{eq:ais}
\end{align}
Every per-sample score is thus a maximum over all $R$ retained adversarial perturbations, not the outcome of a single attack run, and the dataset median of these per-sample worst cases is what \cref{tab: astability} reports. The attack set consists of the methods whose attributions are differentiable functions of the input, namely IG, IDGI, and ERIG. The quantile-based feature selection, the saturation rule, and the repeat-until inner loop in Algorithm~1 of GIG~\cite{kapishnikov2021guided} are input-dependent discrete operations that render the attribution piecewise-defined and discontinuous, so gradient-based worst-case optimization is ill-posed for it. Blur IG is excluded on cost rather than principle; the attack is already second order as noted above, and Blur IG's path points are blurred copies of the input, which increases that cost considerably. We keep the uniform-perturbation results in \Cref{subsec: stability}, as they follow the established protocols of the metrics' original papers~\cite{yeh2019fidelity, alvarez2018robustness, agarwal2022rethinking} and enable comparison with a wider set of methods and with prior work.

As \cref{tab: astability} shows, the conclusion of \cref{tab: stability-results} carries over unchanged to adversarially selected directions. ERIG attains the lowest score on every backbone and every metric, improving worst-case attribution stability by ${\sim}90\times$ on average over the next-best method and reaching up to four orders of magnitude over IG. Notably, ERIG's adversarial worst case remains below every baseline's uniform MS and LLC in \cref{tab: stability-results}; even adversarially selected directions cannot push ERIG's attribution shift up to the level the baselines already reach under isotropic noise, consistent with the direction-independent $\mathcal{O}(1/(2\tau))$ weight-variation bound of \cref{cor: decomp}.

\begin{table}
    \centering
    \caption{\textbf{ImageNet Adversarial Stability:} ERIG is lowest on every backbone and metric, with a ${\sim}90\times$ improvement over the next-best method on average. Moreover, its adversarial AS and ALC stay below every baseline's uniform MS and LLC respectively.}
    \footnotesize
    \begin{tabular}{|c|c|c|c|c|}
        \hline
        \textbf{Model} & \textbf{Method} & \textbf{AS $(\downarrow)$} & \textbf{ALC $(\downarrow)$} & \textbf{AIS $(\downarrow)$}\\
        \hline
        \multirow{3}{*}{ResNet101} & IG & 2.0400 & 0.0811 & 117.301 \\
         & IDGI & \textbf{0.0334} & \textbf{0.0014} & \textbf{65.461} \\
         & ERIG (ours) & $\mathbf{\underline{1.3e\text{\textbf{--}}04}}$ & $\mathbf{\underline{5.7e\text{\textbf{--}}06}}$ & \textbf{\underline{1.033}} \\
        \hline
        \multirow{3}{*}{EfficientNet} & IG & 1.2695 & 0.0331 & \textbf{65.008} \\
         & IDGI & \textbf{0.0254} & $\mathbf{7.4e\text{\textbf{--}}04}$ & 77.782 \\
         & ERIG (ours) & $\mathbf{\underline{9.0e\text{\textbf{--}}04}}$ & $\mathbf{\underline{2.8e\text{\textbf{--}}05}}$ & \textbf{\underline{14.239}} \\
        \hline
        \multirow{3}{*}{MobileNetV4} & IG & 1.7649 & 0.0583 & 40.621 \\
         & IDGI & \textbf{0.0092} & $\mathbf{4.8e\text{\textbf{--}}04}$ & \textbf{39.747} \\
         & ERIG (ours) & $\mathbf{\underline{9.1e\text{\textbf{--}}04}}$ & $\mathbf{\underline{3.6e\text{\textbf{--}}05}}$ & \textbf{\underline{11.086}} \\
        \hline
        \multirow{3}{*}{ViT-Base} & IG & 1.2932 & 0.0288 & \textbf{52.296} \\
         & IDGI & \textbf{0.0369} & $\mathbf{9.0e\text{\textbf{--}}04}$ & 58.266 \\
         & ERIG (ours) & $\mathbf{\underline{3.2e\text{\textbf{--}}04}}$ & $\mathbf{\underline{8.0e\text{\textbf{--}}06}}$ & \textbf{\underline{4.352}} \\
        \hline
        \multirow{3}{*}{EVA-02} & IG & 1.1663 & 0.0237 & \textbf{65.522} \\
         & IDGI & \textbf{0.0295} & $\mathbf{6.2e\text{\textbf{--}}04}$ & 65.661 \\
         & ERIG (ours) & $\mathbf{\underline{2.3e\text{\textbf{--}}04}}$ & $\mathbf{\underline{4.9e\text{\textbf{--}}06}}$ & \textbf{\underline{4.074}} \\
        \hline
        \multirow{3}{*}{DeiT3} & IG & 1.3926 & 0.0323 & 77.716 \\
         & IDGI & \textbf{0.0329} & $\mathbf{8.0e\text{\textbf{--}}04}$ & \textbf{61.983} \\
         & ERIG (ours) & $\mathbf{\underline{1.4e\text{\textbf{--}}04}}$ & $\mathbf{\underline{3.4e\text{\textbf{--}}06}}$ & \textbf{\underline{1.903}} \\
        \hline
    \end{tabular}
    \label{tab: astability}
\end{table}

\subsection{\texorpdfstring{Signed ERIG}{Signed ERIG}} \label{appx: serig}
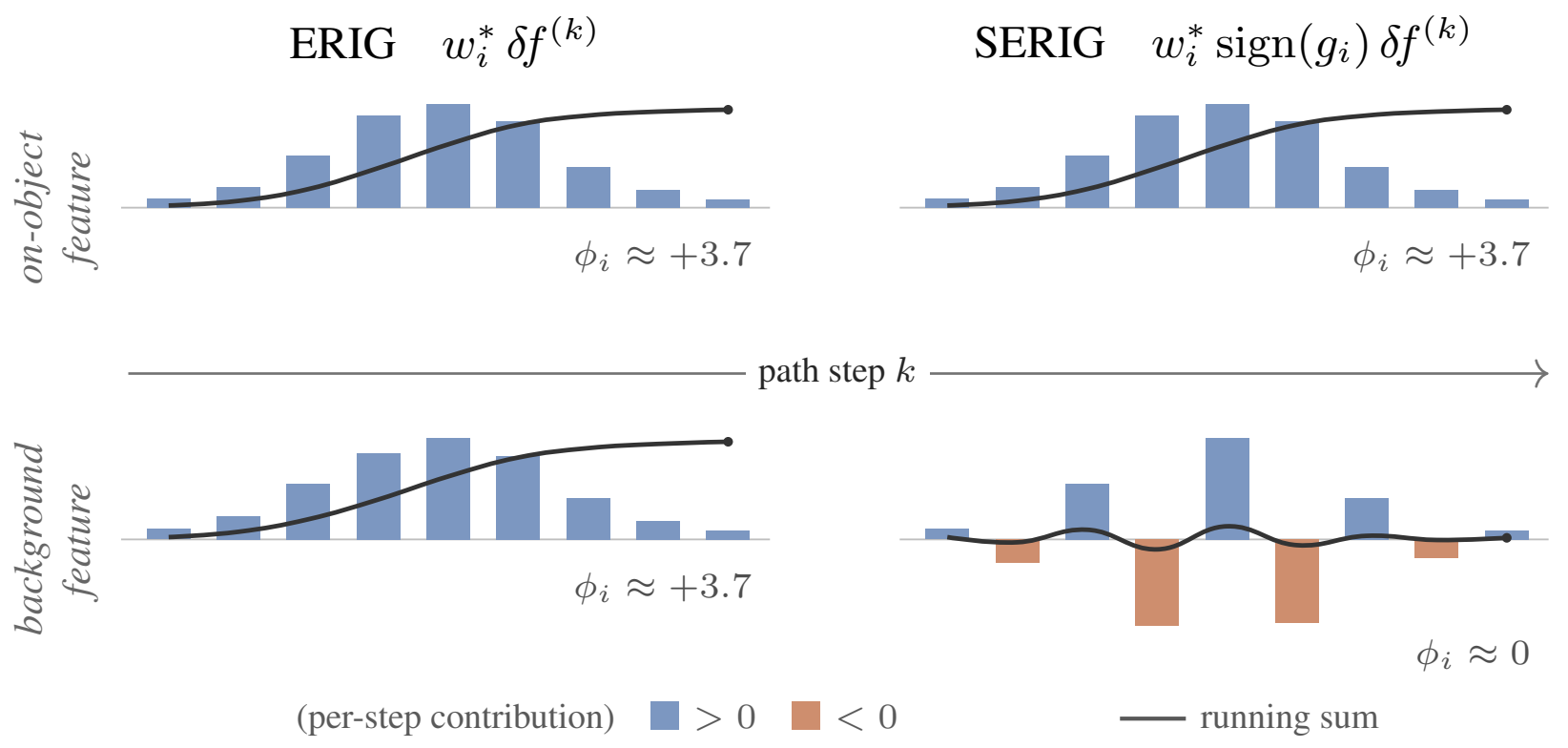
\begin{figure}[htb]
\centering
\definecolor{barpos}{RGB}{124,151,193}
\definecolor{barneg}{RGB}{206,142,110}
\begin{tikzpicture}[
    every node/.style={font=\footnotesize},
    zero/.style={gray!50, line width=0.35pt},
    bpos/.style={draw=none, fill=barpos},
    bneg/.style={draw=none, fill=barneg},
    cum/.style={line width=0.9pt, black!80, line join=round},
    dot/.style={circle, fill=black!80, inner sep=0pt, minimum size=1.9pt},
    hdr/.style={font=\small},
    rowlab/.style={font=\footnotesize\itshape, black!60, align=center},
    tot/.style={font=\scriptsize, black!70, anchor=east},
]
\def\bw{0.15}
\def\px{0.28}
\def\dx{0.485}

\node[hdr] at (2.2, 1.15) {ERIG\quad$w_i^{*}\,\delta\!f^{(k)}$};
\node[hdr] at (7.6, 1.15) {SERIG\quad$w_i^{*}\,\mathrm{sign}(g_i)\,\delta\!f^{(k)}$};

\node[rowlab, rotate=90] at (-0.5,  0.0) {on-object\\feature};
\node[rowlab, rotate=90] at (-0.5, -2.3) {background\\feature};

\foreach \ox in {0, 5.4} {
  \begin{scope}[shift={(\ox,0)}]
    \draw[zero] (-0.05,0) -- (4.45,0);
    \foreach \k/\h in {1/0.064,2/0.144,3/0.360,4/0.640,5/0.720,6/0.600,7/0.280,8/0.120,9/0.056}
      {\pgfmathsetmacro{\x}{\px+(\k-1)*\dx}
       \fill[bpos] (\x-\bw,0) rectangle (\x+\bw,\h);}
    \draw[cum] plot[smooth, tension=0.75] coordinates {(0.28,0.015)(0.765,0.047)(1.25,0.129)(1.735,0.275)(2.22,0.439)(2.705,0.575)(3.19,0.639)(3.675,0.666)(4.16,0.679)};
    \node[dot] at (4.16,0.679) {};
    \node[tot] at (4.45,-0.34) {$\phi_i \approx +3.7$};
  \end{scope}}

\draw[->, black!55, line width=0.5pt] (0,-1.15) -- (9.85,-1.15);
\node[fill=white, inner sep=2.5pt, font=\scriptsize, text=black!85] at (4.92,-1.15) {path step $k$};

\begin{scope}[shift={(0,-2.3)}]
  \draw[zero] (-0.05,0) -- (4.45,0);
  \foreach \k/\h in {1/0.072,2/0.160,3/0.384,4/0.600,5/0.704,6/0.576,7/0.288,8/0.128,9/0.064}
    {\pgfmathsetmacro{\x}{\px+(\k-1)*\dx}
     \fill[bpos] (\x-\bw,0) rectangle (\x+\bw,\h);}
  \draw[cum] plot[smooth, tension=0.75] coordinates {(0.28,0.016)(0.765,0.053)(1.25,0.140)(1.735,0.277)(2.22,0.437)(2.705,0.568)(3.19,0.633)(3.675,0.662)(4.16,0.677)};
  \node[dot] at (4.16,0.677) {};
  \node[tot] at (4.45,-0.34) {$\phi_i \approx +3.7$};
\end{scope}

\begin{scope}[shift={(5.4,-2.3)}]
  \draw[zero] (-0.05,0) -- (4.45,0);
  \foreach \k/\h in {1/0.072,3/0.384,5/0.704,7/0.288,9/0.064}
    {\pgfmathsetmacro{\x}{\px+(\k-1)*\dx}
     \fill[bpos] (\x-\bw,0) rectangle (\x+\bw,\h);}
  \foreach \k/\h in {2/-0.160,4/-0.600,6/-0.576,8/-0.128}
    {\pgfmathsetmacro{\x}{\px+(\k-1)*\dx}
     \fill[bneg] (\x-\bw,0) rectangle (\x+\bw,\h);}
  \draw[cum] plot[smooth, tension=0.75] coordinates {(0.28,0.016)(0.765,-0.020)(1.25,0.067)(1.735,-0.069)(2.22,0.091)(2.705,-0.040)(3.19,0.025)(3.675,-0.004)(4.16,0.011)};
  \node[dot] at (4.16,0.011) {};
  \node[tot] at (4.45,-0.80) {$\phi_i \approx 0$};
\end{scope}

\node[anchor=north, font=\scriptsize, black!65] at (4.92,-3.30)
  {(per-step contribution)\quad
   \textcolor{barpos}{\rule[-0.1ex]{5.5pt}{5.5pt}}\,~$>0$\quad
   \textcolor{barneg}{\rule[-0.1ex]{5.5pt}{5.5pt}}\,~$<0$\hspace{6em}
   \rule[0.45ex]{13pt}{0.8pt}\,~running sum};
\end{tikzpicture}
\caption{\textbf{Schematic -- Why the Signed Variant Cancels:} ERIG's per-step terms carry the sign of $\delta\!f^{(k)}$, which is shared by every feature, so nothing cancels if $f$ keeps increasing, and both on-object and background features accumulate alike. SERIG multiplies the weights by $\mathrm{sign}(g_i)$. This sign is usually stable on the object but alternating along the path for background, so only the background feature cancels toward zero in the aggregation (summation). Here, the hypothetical per-step weight magnitudes follow the movement of the model output along the path (\cref{fig: weight-regions}), small over the flat and saturated segments and large where the output changes rapidly.}
\label{fig: serig-cancellation}
\end{figure}
In \cref{eq:erig-formula} the weights are nonnegative, but the sign of each contribution comes from the function increments $df$, which are signed. ERIG attributions therefore can be negative; features weighted strongly on path segments where $f$ decreases receive negative mass. Squaring removes the sign of the per-feature gradient inside the weighting, not the sign of the attribution, and IDGI~\cite{yang2023idgi} shares exactly this structure. The attribution sign is nevertheless coarsened, and it is true that downstream applications such as absent-concept explanation~\cite{hesse2026missing} do exploit sign information. We therefore ablate a signed variant (SERIG) with weights $\tilde{w}_i = w_i^{*} \cdot \mathrm{sign}(g_i)$, which restores the per-step gradient sign at no extra computational cost while preserving the magnitude profile (\cref{fig: serig-cancellation}). \Cref{tab: serig-fidelity,tab: serig-localization,tab: serig-stability,tab: serig-noise} report the ImageNet evaluation suite of the main paper.

\begin{table}
    \centering
    \begin{minipage}[t]{0.48\textwidth}
        \centering
        \caption{\textbf{SERIG Ablation Fidelity:} ERIG beats the signed variant in every cell, by factors up to $1.7\times$ on ISC and $2.1\times$ on DEL.}
        \footnotesize
        \begin{tabular}{|c|c|c|c|c|}
            \hline
            \textbf{Model} & \textbf{Method} & \textbf{ISC $(\uparrow)$} & \textbf{AIC $(\uparrow)$} & \textbf{DEL $(\downarrow)$}\\
            \hline
            \multirow{2}{*}{ViT-Base} & ERIG & \textbf{\underline{0.6930}} & \textbf{\underline{0.9013}} & \textbf{\underline{0.4902}} \\
             & SERIG & 0.4124 & 0.6523 & 0.7479 \\
            \hline
            \multirow{2}{*}{EVA02} & ERIG & \textbf{\underline{0.5913}} & \textbf{\underline{0.8907}} & \textbf{\underline{0.5373}} \\
             & SERIG & 0.4210 & 0.7621 & 0.8037 \\
            \hline
            \multirow{2}{*}{MobileNetV4} & ERIG & \textbf{\underline{0.4118}} & \textbf{\underline{0.7234}} & \textbf{\underline{0.3626}} \\
             & SERIG & 0.3105 & 0.5086 & 0.7007 \\
            \hline
            \multirow{2}{*}{EfficientNet} & ERIG & \textbf{\underline{0.4521}} & \textbf{\underline{0.8323}} & \textbf{\underline{0.4562}} \\
             & SERIG & 0.4106 & 0.7030 & 0.7140 \\
            \hline
            \multirow{2}{*}{ResNet101} & ERIG & \textbf{\underline{0.4893}} & \textbf{\underline{0.6610}} & \textbf{\underline{0.3270}} \\
             & SERIG & 0.3112 & 0.4379 & 0.7003 \\
            \hline
            \multirow{2}{*}{DeiT3} & ERIG & \textbf{\underline{0.7320}} & \textbf{\underline{0.9010}} & \textbf{\underline{0.5561}} \\
             & SERIG & 0.4363 & 0.6311 & 0.7743 \\
            \hline
        \end{tabular}
        \label{tab: serig-fidelity}
    \end{minipage}
    \hfill
    \begin{minipage}[t]{0.48\textwidth}
        \centering
        \caption{\textbf{SERIG Ablation Localization:} The signed variant wins IoU and PF1 on every backbone, while ERIG keeps the best PA.}
        \footnotesize
        \begin{tabular}{|c|c|c|c|c|}
            \hline
            \textbf{Model} & \textbf{Method} & \textbf{IoU $(\uparrow)$} & \textbf{PA $(\uparrow)$} & \textbf{PF1 $(\uparrow)$}\\
            \hline
            \multirow{2}{*}{ViT-Base} & ERIG & 0.1203 & \textbf{\underline{0.7054}} & 0.1893 \\
             & SERIG & \textbf{\underline{0.1842}} & 0.5130 & \textbf{\underline{0.2647}} \\
            \hline
            \multirow{2}{*}{EVA02} & ERIG & 0.1317 & \textbf{\underline{0.7054}} & 0.2031 \\
             & SERIG & \textbf{\underline{0.1898}} & 0.5232 & \textbf{\underline{0.2728}} \\
            \hline
            \multirow{2}{*}{MobileNetV4} & ERIG & 0.1615 & \textbf{\underline{0.6521}} & 0.2424 \\
             & SERIG & \textbf{\underline{0.2174}} & 0.5275 & \textbf{\underline{0.3085}} \\
            \hline
            \multirow{2}{*}{EfficientNet} & ERIG & 0.1085 & \textbf{\underline{0.5982}} & 0.1714 \\
             & SERIG & \textbf{\underline{0.2265}} & 0.5097 & \textbf{\underline{0.3154}} \\
            \hline
            \multirow{2}{*}{ResNet101} & ERIG & 0.1617 & \textbf{\underline{0.6580}} & 0.2419 \\
             & SERIG & \textbf{\underline{0.2506}} & 0.5239 & \textbf{\underline{0.3469}} \\
            \hline
            \multirow{2}{*}{DeiT3} & ERIG & 0.1179 & \textbf{\underline{0.6637}} & 0.1819 \\
             & SERIG & \textbf{\underline{0.2482}} & 0.4925 & \textbf{\underline{0.3372}} \\
            \hline
        \end{tabular}
        \label{tab: serig-localization}
    \end{minipage}

    \vspace{1.5em}

    \begin{minipage}[t]{0.48\textwidth}
        \centering
        \caption{\textbf{SERIG Ablation Stability:} ERIG is more stable in every cell, by up to two orders of magnitude.}
        \footnotesize
        \begin{tabular}{|c|c|c|c|c|}
            \hline
            \textbf{Model} & \textbf{Method} & \textbf{MS $(\downarrow)$} & \textbf{LLC $(\downarrow)$} & \textbf{RIS $(\downarrow)$}\\
            \hline
            \multirow{2}{*}{ViT-Base} & ERIG & $\mathbf{\underline{3.0e\text{--}04}}$ & $\mathbf{\underline{9.0e\text{--}06}}$ & \textbf{\underline{4.9460}} \\
             & SERIG & $6.1e\text{--}04$ & $1.6e\text{--}05$ & 12.040 \\
            \hline
            \multirow{2}{*}{EVA02} & ERIG & $\mathbf{\underline{6.7e\text{--}05}}$ & $\mathbf{\underline{1.5e\text{--}06}}$ & \textbf{\underline{1.6010}} \\
             & SERIG & $4.5e\text{--}04$ & $1.0e\text{--}05$ & 10.157 \\
            \hline
            \multirow{2}{*}{MobileNetV4} & ERIG & $\mathbf{\underline{1.0e\text{--}04}}$ & $\mathbf{\underline{5.4e\text{--}06}}$ & \textbf{\underline{2.1960}} \\
             & SERIG & 0.0013 & $5.0e\text{--}05$ & 21.166 \\
            \hline
            \multirow{2}{*}{EfficientNet} & ERIG & $\mathbf{\underline{6.7e\text{--}05}}$ & $\mathbf{\underline{2.1e\text{--}06}}$ & \textbf{\underline{1.2710}} \\
             & SERIG & $7.2e\text{--}04$ & $2.2e\text{--}05$ & 16.032 \\
            \hline
            \multirow{2}{*}{ResNet101} & ERIG & $\mathbf{\underline{8.6e\text{--}05}}$ & $\mathbf{\underline{3.8e\text{--}06}}$ & \textbf{\underline{1.0660}} \\
             & SERIG & 0.0042 & $1.9e\text{--}04$ & 41.584 \\
            \hline
            \multirow{2}{*}{DeiT3} & ERIG & $\mathbf{\underline{3.9e\text{--}05}}$ & $\mathbf{\underline{1.0e\text{--}06}}$ & \textbf{\underline{0.7560}} \\
             & SERIG & 0.0080 & $2.1e\text{--}04$ & 10.234 \\
            \hline
        \end{tabular}
        \label{tab: serig-stability}
    \end{minipage}
    \hfill
    \begin{minipage}[t]{0.48\textwidth}
        \centering
        \caption{\textbf{SERIG Ablation Noise:} ERIG attains the higher LINV in every cell, by $5.8$ to $8.7$ LINV over the signed variant.}
        \footnotesize
        \begin{tabular}{|c|c|c|c|}
            \hline
            \textbf{Model} & \textbf{Method} & \textbf{LINV-C $(\uparrow)$} & \textbf{LINV-P $(\uparrow)$}\\
            \hline
            \multirow{2}{*}{ViT-Base} & ERIG & \textbf{\underline{19.343}} & \textbf{\underline{19.201}} \\
             & SERIG & 12.028 & 11.920 \\
            \hline
            \multirow{2}{*}{EVA02} & ERIG & \textbf{\underline{20.128}} & \textbf{\underline{20.115}} \\
             & SERIG & 12.639 & 12.636 \\
            \hline
            \multirow{2}{*}{MobileNetV4} & ERIG & \textbf{\underline{18.130}} & \textbf{\underline{18.113}} \\
             & SERIG & 11.440 & 11.331 \\
            \hline
            \multirow{2}{*}{EfficientNet} & ERIG & \textbf{\underline{20.705}} & \textbf{\underline{20.712}} \\
             & SERIG & 11.976 & 11.970 \\
            \hline
            \multirow{2}{*}{ResNet101} & ERIG & \textbf{\underline{16.807}} & \textbf{\underline{16.778}} \\
             & SERIG & 10.987 & 10.932 \\
            \hline
            \multirow{2}{*}{DeiT3} & ERIG & \textbf{\underline{19.578}} & \textbf{\underline{19.584}} \\
             & SERIG & 11.682 & 12.026 \\
            \hline
        \end{tabular}
        \label{tab: serig-noise}
    \end{minipage}
\end{table}

The ablation reveals a consistent trade-off. Restoring the gradient sign substantially improves localization across all six backbones, with IoU increasing from the $0.11$--$0.16$ range to the $0.18$--$0.25$ range and PF1 improving by $27$--$85\%$. This improvement comes at the cost of fidelity and stability. ISC decreases by $0.04$--$0.30$, while MS and LLC worsen by factors of $2$--$7$ on ViT-Base and EVA-02 and by one to two orders of magnitude on the remaining backbones, although SERIG remains substantially more stable than IG, BIG, and GIG in \cref{tab: stability-results}. The underlying mechanism is that the two maps capture different properties (\cref{fig: serig-cancellation}). ERIG accumulates the magnitude of each feature's influence on the output along the integration path, irrespective of the direction of that influence, and thus produces a sensitivity map. The signed variant accumulates attribution only when a feature's influence has a consistent direction, causing contributions with opposing signs along the path to cancel and producing a coherence map. This directional coherence is precisely the property favored by localization. Object regions tend to provide evidence that is more directionally consistent along the integration path, whereas background and texture can influence the prediction in more variable directions, as these regions vary substantially across images while the object remains comparatively consistent. Cancellation therefore suppresses precisely the off-object attribution mass. In contrast, fidelity probes favor sensitivity. Deletion and insertion measure whether removing attributed pixels changes the model output, and a directionally unstable pixel can still exert a positive influence on $f$ at the input $\bm{x}^a$. The signed map can therefore cancel pixels to which the model demonstrably responds, while the fidelity probes continue to detect output changes in regions to which SERIG assigns little or no attribution. The stability cost has the same underlying cause. Since $\mathrm{sign}(g_i)$ is discontinuous at $g_i = 0$, small perturbations can flip the sign factor even when the corresponding change in the softmax magnitude is negligible.

The ablation also clarifies the role of sign at different levels of the attribution process. Sign at the level of the final attribution is clearly meaningful, and ERIG preserves this information. The signed $\delta f$ increments indicate whether a feature's importance is associated with increases or decreases in the class score along the integration path. For example, a feature that receives a small weight when $\delta f$ is positive but a large weight when $\delta f$ is negative can receive a negative attribution. This directional information is useful for downstream applications. In contrast, the sign of the instantaneous gradient is a property of an individual evaluation point. Because of gradient irregularity, this sign can change as the evaluation point moves along the integration path or across noise samples in SmoothGrad-style averaging. An aggregation based on the instantaneous gradient sign therefore reflects the behavior of the ensemble of evaluation points rather than an intrinsic property of the feature's effect. The ablation suggests that this per-step gradient sign is not the primary determinant of explanation quality. Introducing it degrades fidelity, stability, and noise performance across all backbones, while improving object alignment. The broader comparison supports the same conclusion. Among the path-based methods evaluated, IDGI and ERIG, which discard the per-step gradient sign and retain sign information only through the increments, consistently achieve the strongest fidelity and stability. In contrast, methods that integrate the raw signed gradient, including IG, BIG, and GIG, perform worse on both criteria. Reintroducing the gradient sign into ERIG's otherwise identical weighting reproduces this gap in a controlled setting. Because the signed variant requires no additional computational cost and uses the same weights, SERIG can be preferred for localization-oriented applications, while ERIG remains the default choice.